\documentclass[conference]{IEEEtran}
\IEEEoverridecommandlockouts
\usepackage{cite}
\usepackage{amsmath,amssymb,amsfonts}
\usepackage{algorithmic}
\usepackage{graphicx}
\usepackage{textcomp}
\usepackage{xcolor}
\usepackage{multirow}
\def\BibTeX{{\rm B\kern-.05em{\sc i\kern-.025em b}\kern-.08em
    T\kern-.1667em\lower.7ex\hbox{E}\kern-.125emX}}
\begin{document}

\title{Fusion of Short-term and Long-term Attention for Video Mirror Detection\\
}

\author{
\IEEEauthorblockN{1\textsuperscript{st} Mingchen Xu}
\IEEEauthorblockA{\textit{School of Computer Science and Informatics}\\
\textit{Cardiff University}\\
Cardiff, United Kingdom \\
xum35@cardiff.ac.uk}\\
\IEEEauthorblockN{3\textsuperscript{rd} Yukun Lai}
\IEEEauthorblockA{\textit{School of Computer Science and Informatics}\\
\textit{Cardiff University}\\
Cardiff, United Kingdom \\
laiy4@cardiff.ac.uk}
\and
\IEEEauthorblockN{2\textsuperscript{nd} Jing Wu}
\IEEEauthorblockA{\textit{School of Computer Science and Informatics}\\
\textit{Cardiff University}\\
Cardiff, United Kingdom \\
wuj11@cardiff.ac.uk}\\
\IEEEauthorblockN{4\textsuperscript{th} Ze Ji}
\IEEEauthorblockA{\textit{School of Engineering}\\
\textit{Cardiff University}\\
Cardiff, United Kingdom \\
jiz1@cardiff.ac.uk}
}

\maketitle

\begin{abstract}
Techniques for detecting mirrors from static images have witnessed rapid growth in recent years. However, these methods detect mirrors from single input images. Detecting mirrors from video requires further consideration of temporal consistency between frames. We observe that humans can recognize mirror candidates, from just one or two frames, based on their appearance (e.g. shape, color). However, to ensure that the candidate is indeed a mirror (not a picture or a window), we often need to observe more frames for a global view. This observation motivates us to detect mirrors by fusing appearance features extracted from a short-term attention module and context information extracted from a long-term attention module. To evaluate the performance, we build a challenging benchmark dataset of 19,255 frames from 281 videos. Experimental results demonstrate that our method achieves state-of-the-art performance on the benchmark dataset.
\end{abstract}


\section{Introduction}
Mirrors are commonly seen in environments. More and more attention has been drawn to mirror detection in computer vision. It is because, on the one hand, detecting mirrors can benefit scene-understanding tasks. The reflection of the mirror can provide hints for locating objects \cite{b1} with 3D information~\cite{b2}. reconstructing human pose \cite{b2}, and reconstructing scenes \cite{b3}.  On the other hand, ignoring mirrors may affect the performance of some computer vision tasks. For example, a service robot may treat reflected objects as real ones. Therefore, it is important for computer vision systems to be able to detect and segment mirrors from input images.

\begin{figure}[htb]
\setlength\tabcolsep{2pt}
\centering
    \begin{tabular}{p{0.082\textwidth}p{0.082\textwidth}p{0.082\textwidth}p{0.082\textwidth}p{0.082\textwidth}}
      \includegraphics[width=15.5mm, height=10.5mm]
      {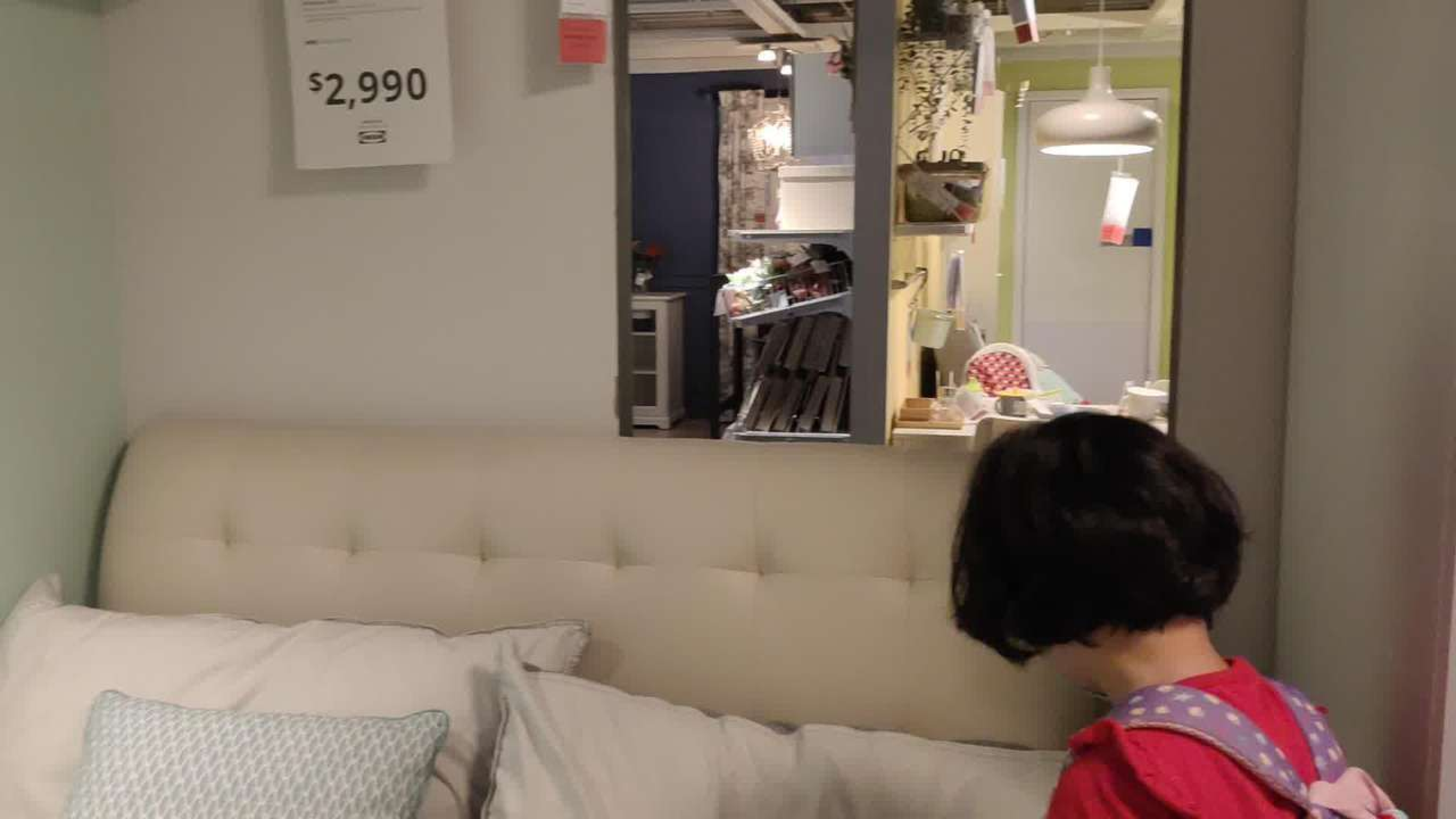}&
      \includegraphics[width=15.5mm, height=10.5mm]
      {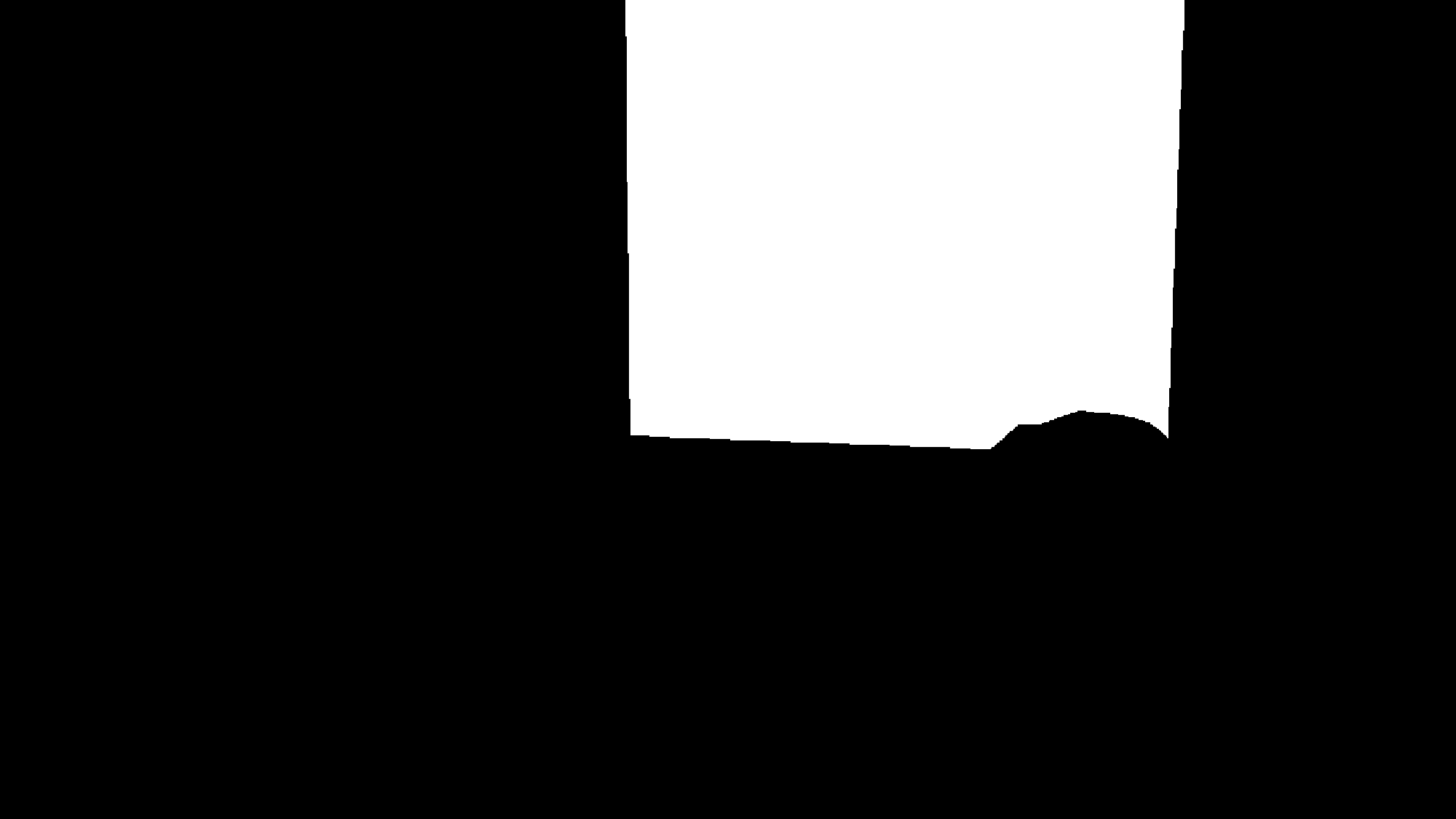}&
      \includegraphics[width=15mm, height=10mm]
      {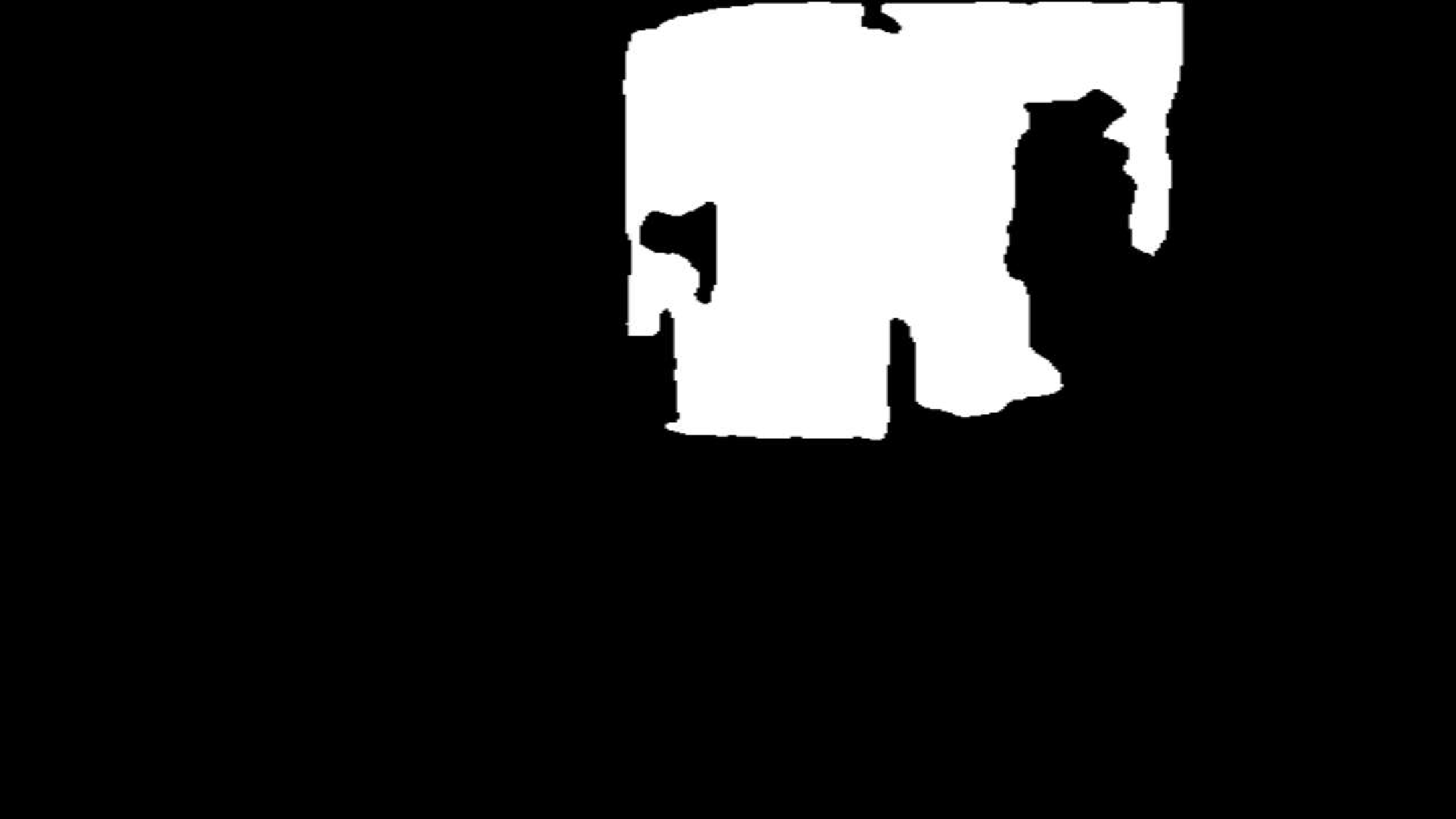}&
      \includegraphics[width=15mm, height=10mm]
      {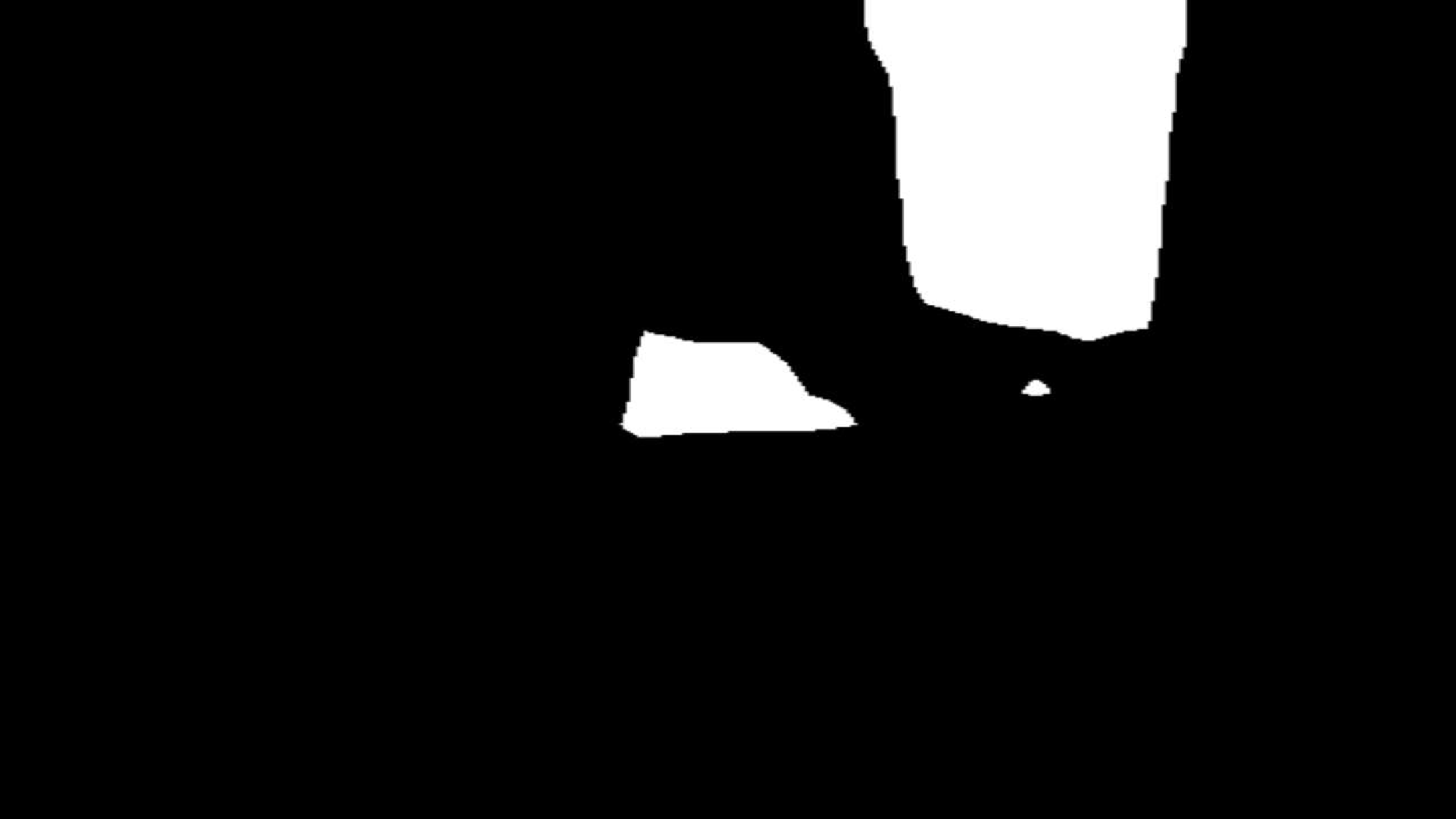}&
      \includegraphics[width=15mm, height=10mm]
      {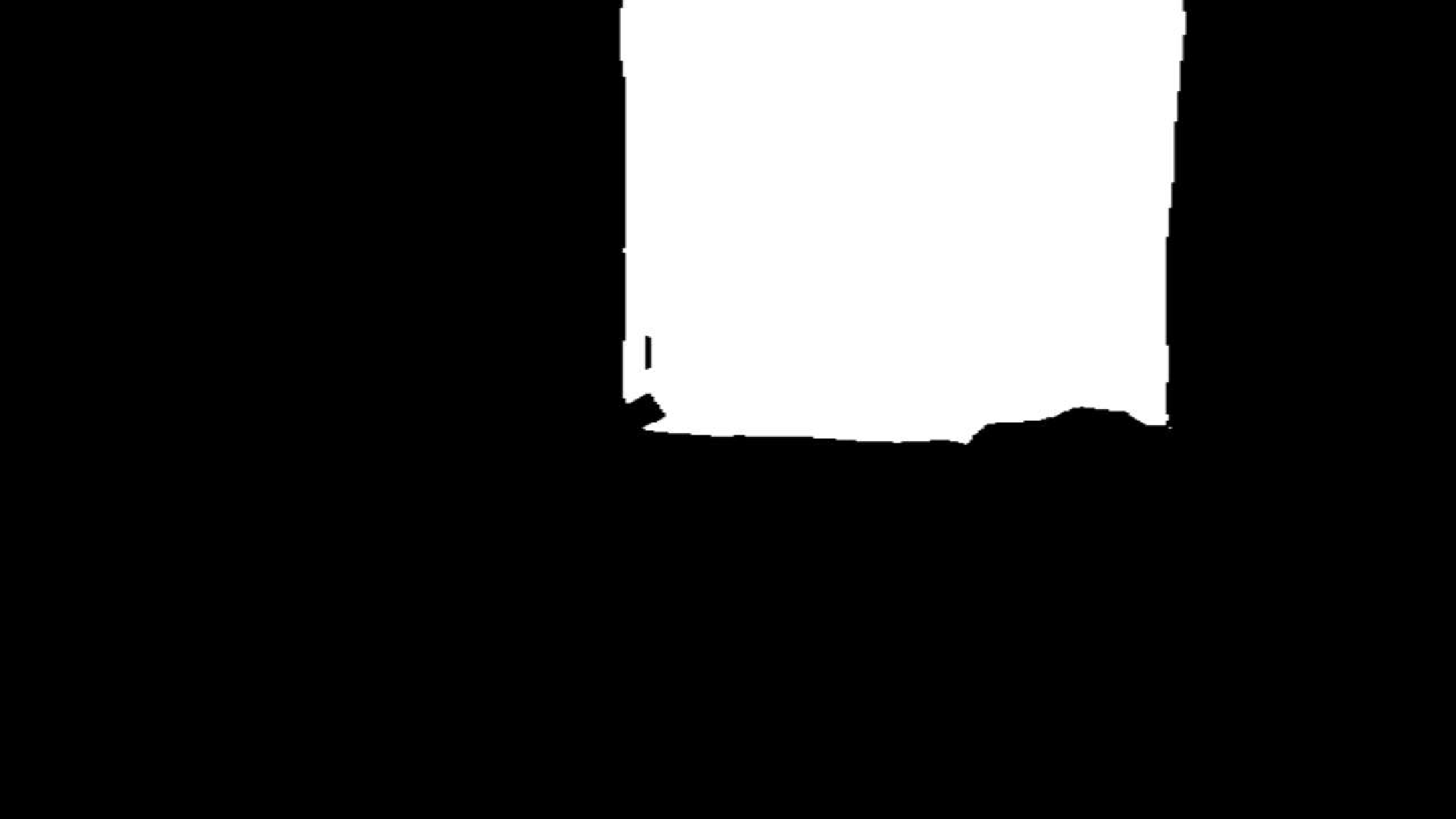}\\
      \includegraphics[width=15mm, height=10mm]
      {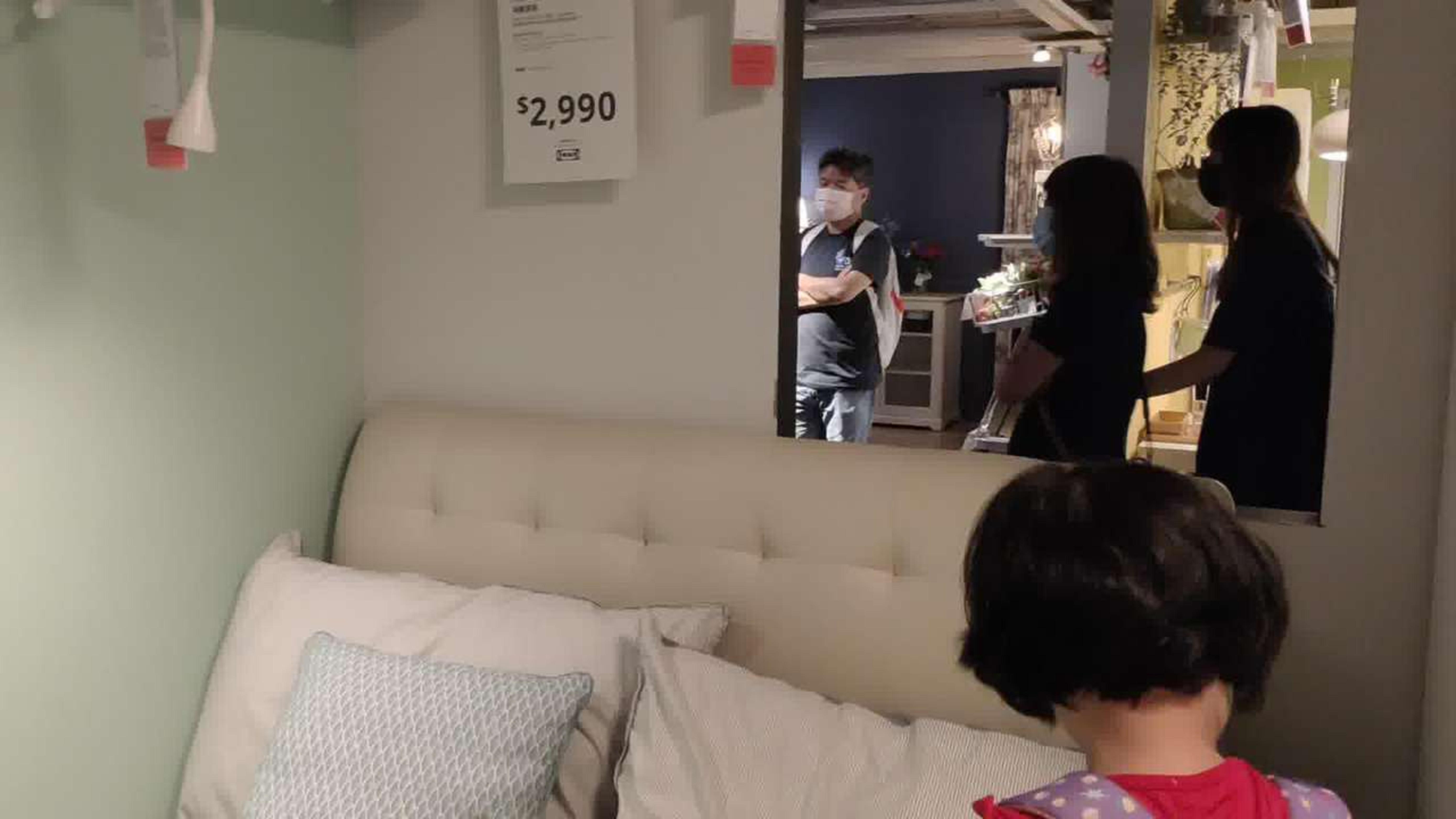}&
      \includegraphics[width=15mm, height=10mm]
      {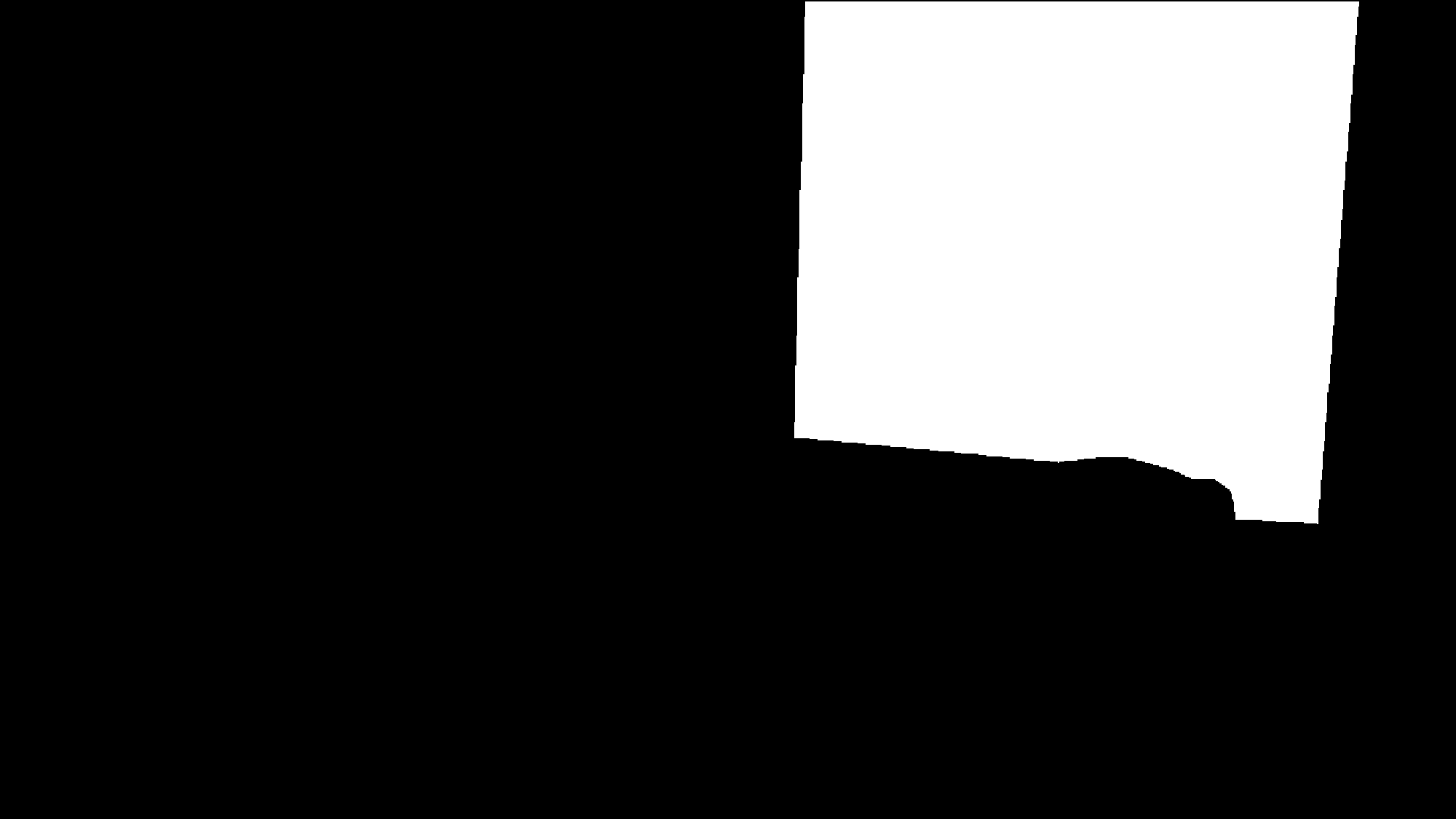}&
      \includegraphics[width=15mm, height=10mm]
      {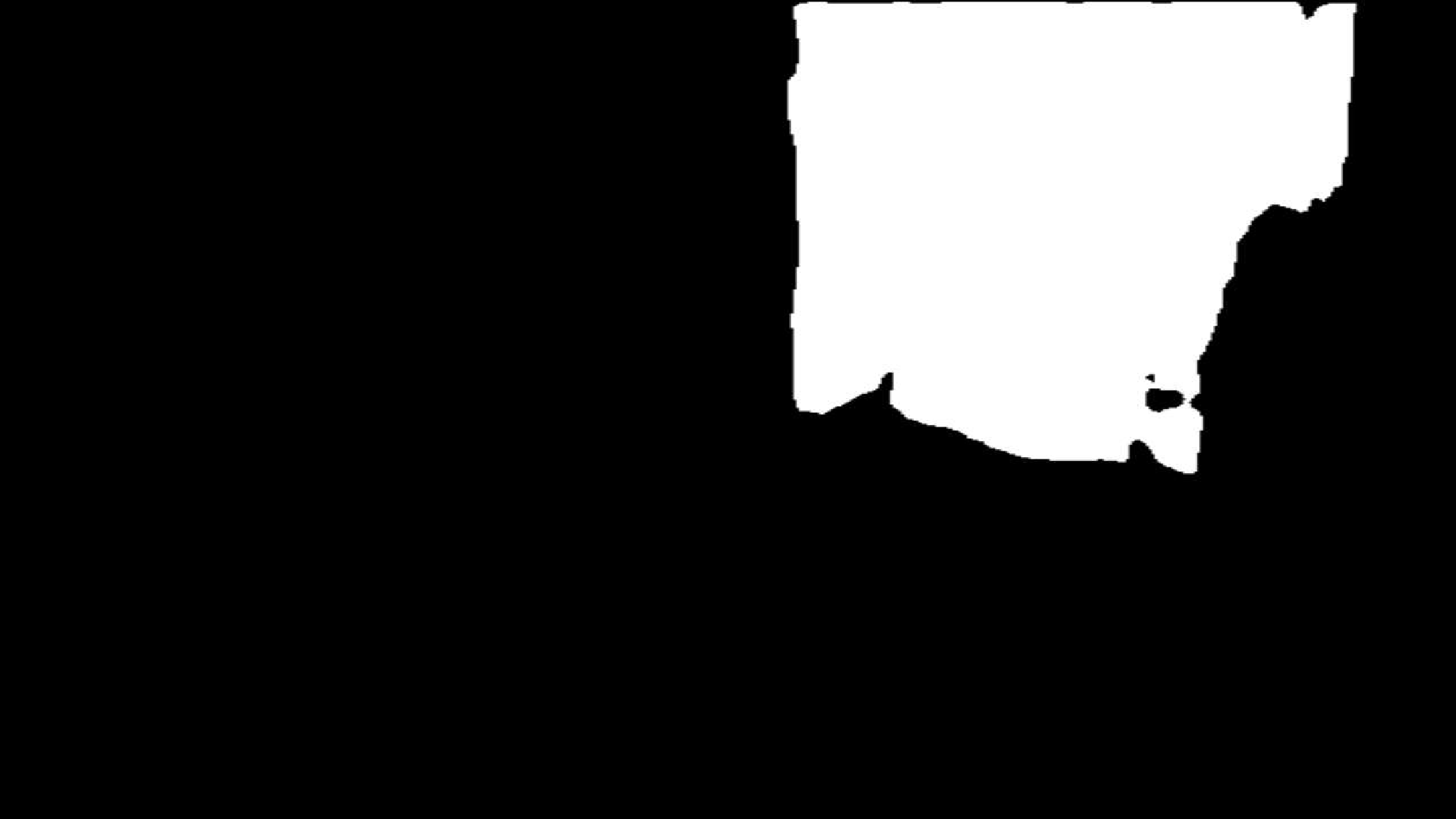}&
      \includegraphics[width=15mm, height=10mm]
      {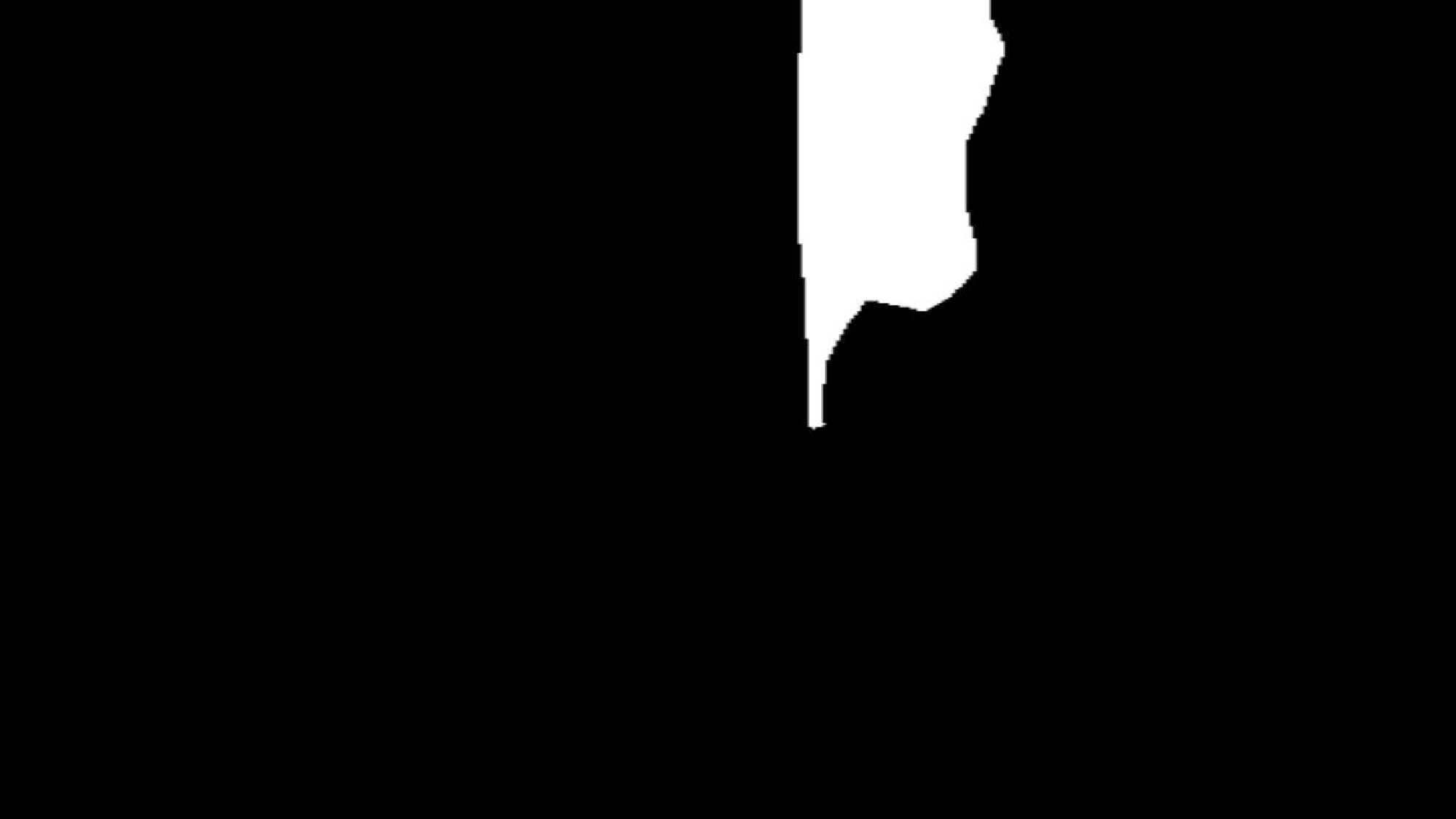}&
      \includegraphics[width=15mm, height=10mm]
      {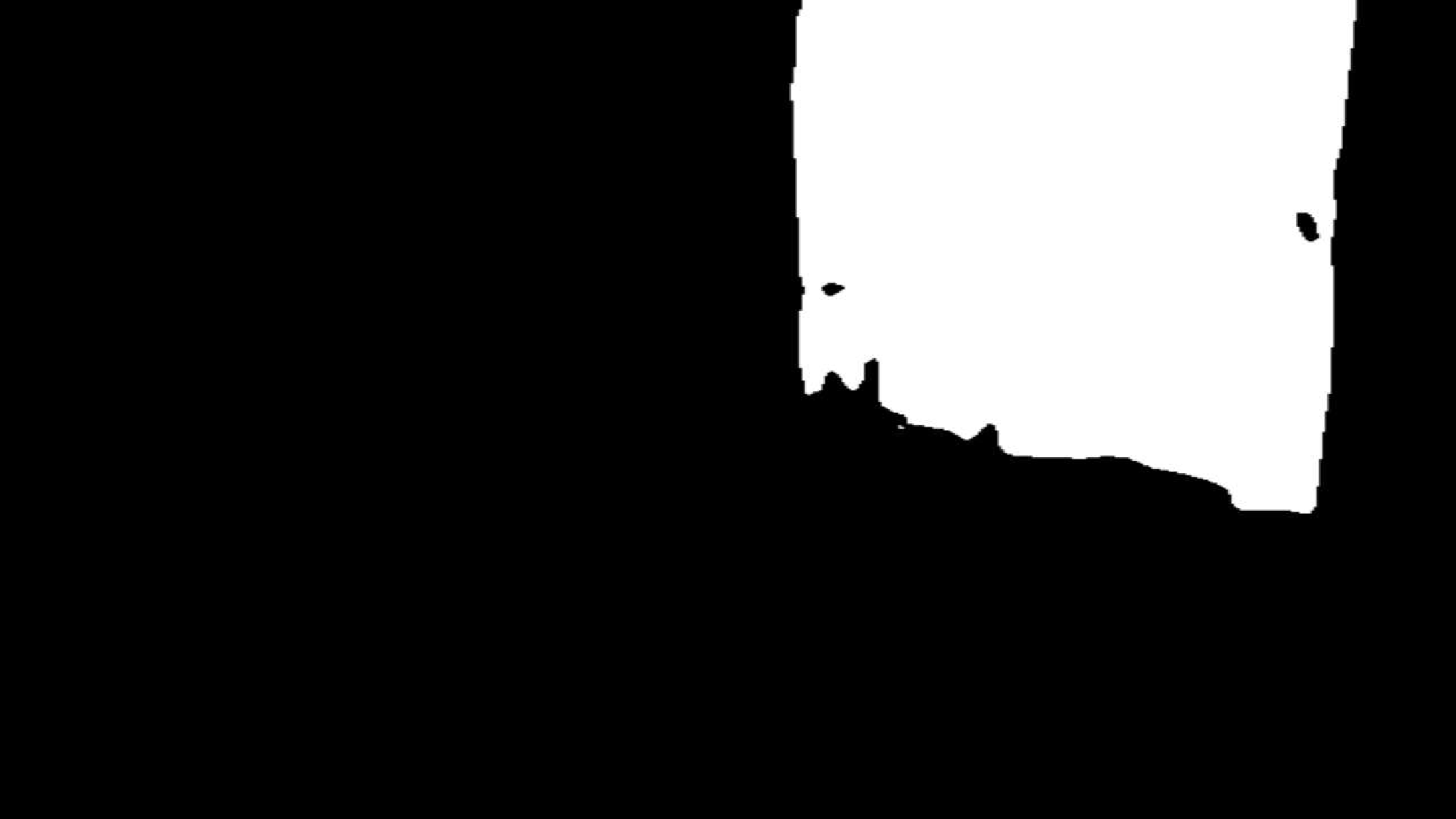}\\
      \includegraphics[width=15mm, height=10mm]
      {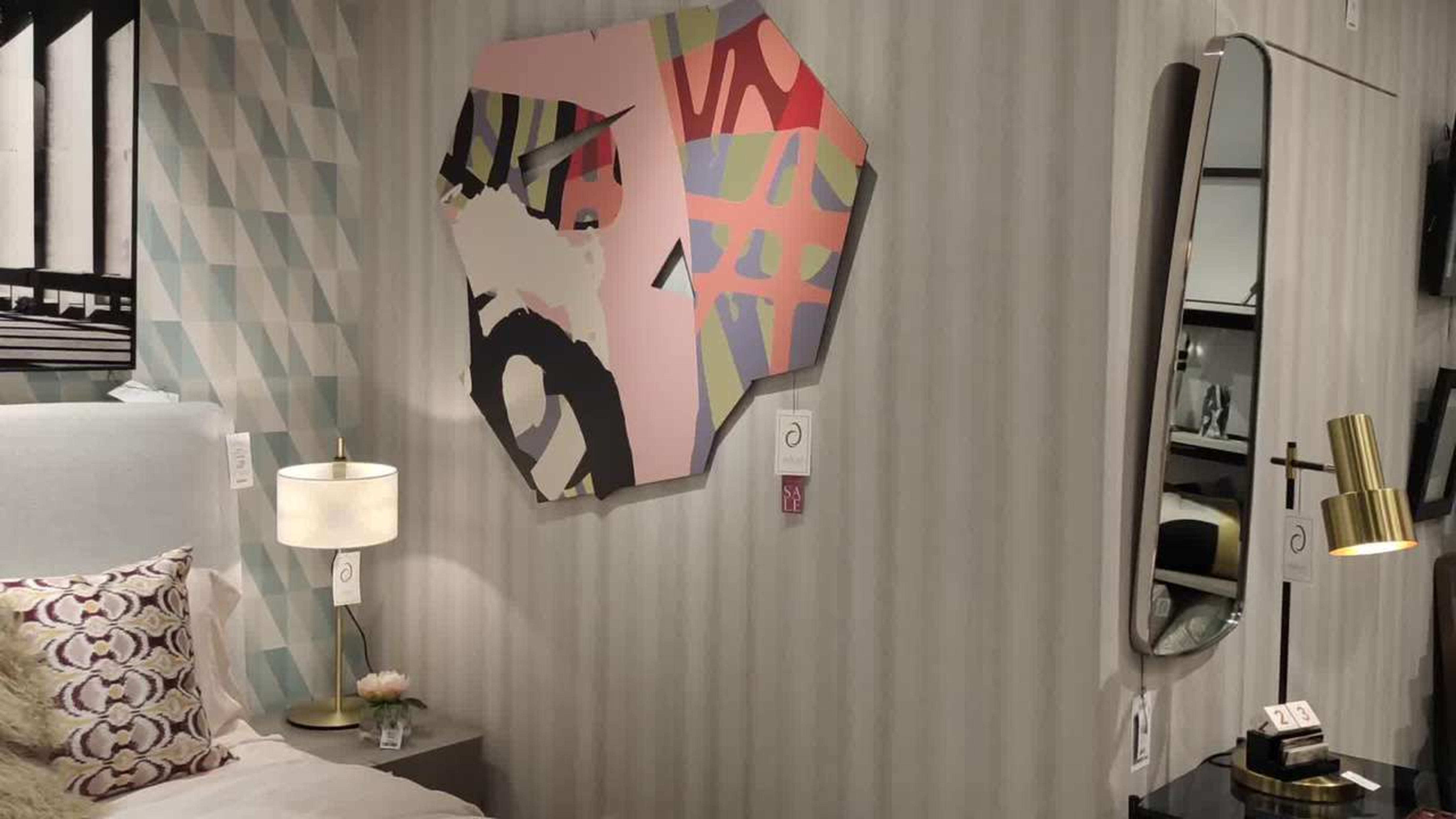}&
      \includegraphics[width=15mm, height=10mm]
      {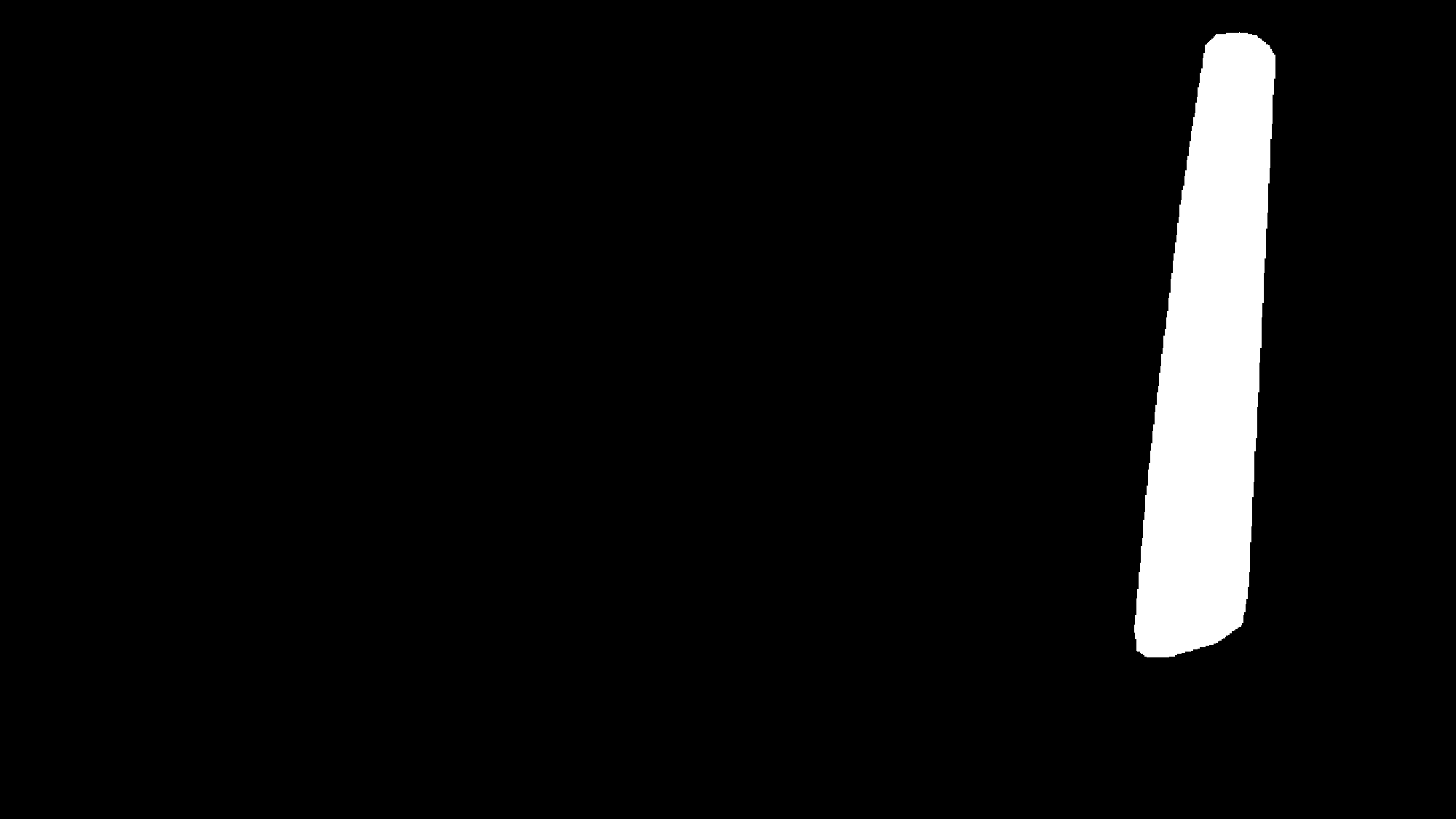}&
      \includegraphics[width=15mm, height=10mm]
      {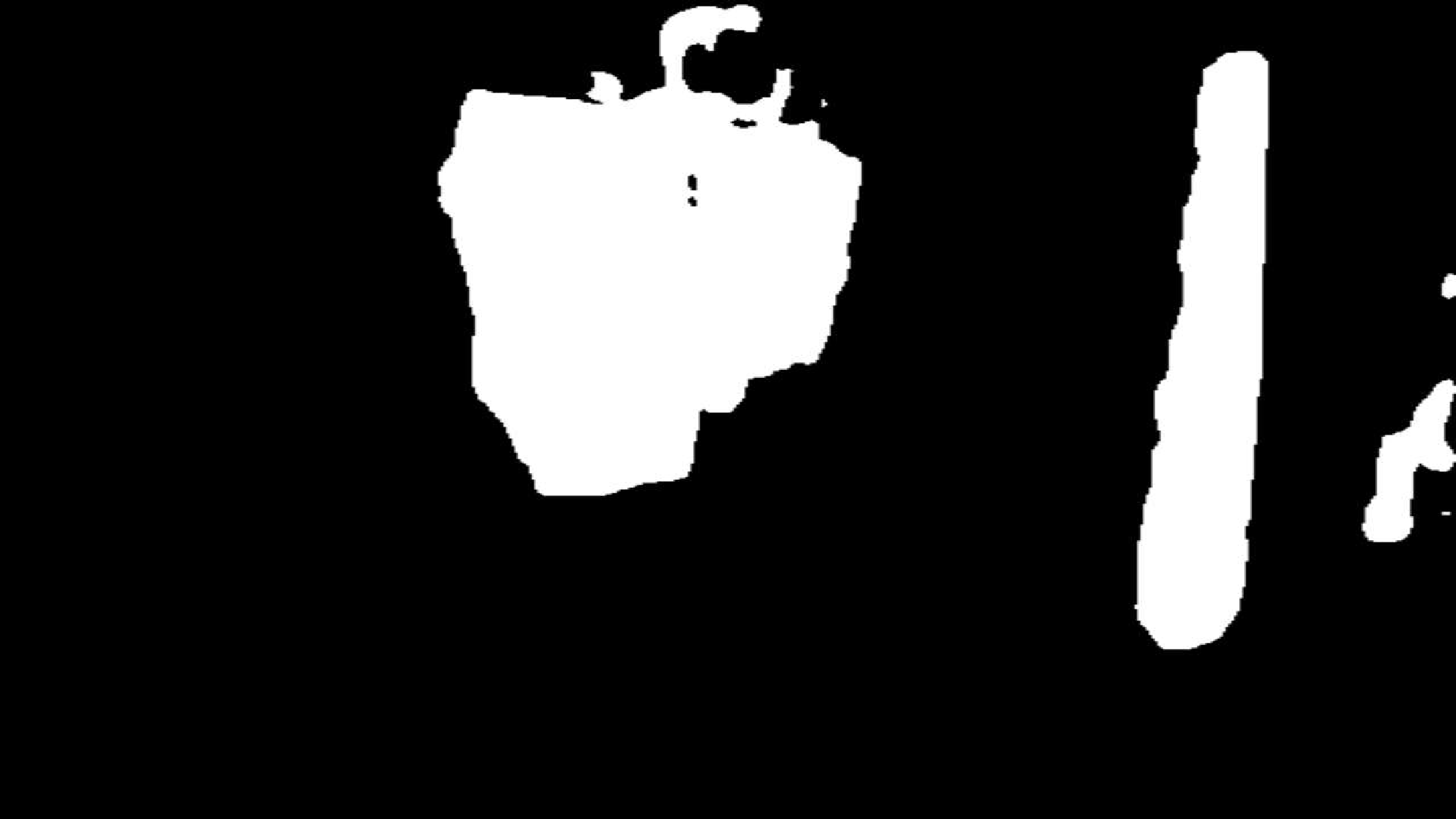}&
      \includegraphics[width=15mm, height=10mm]
      {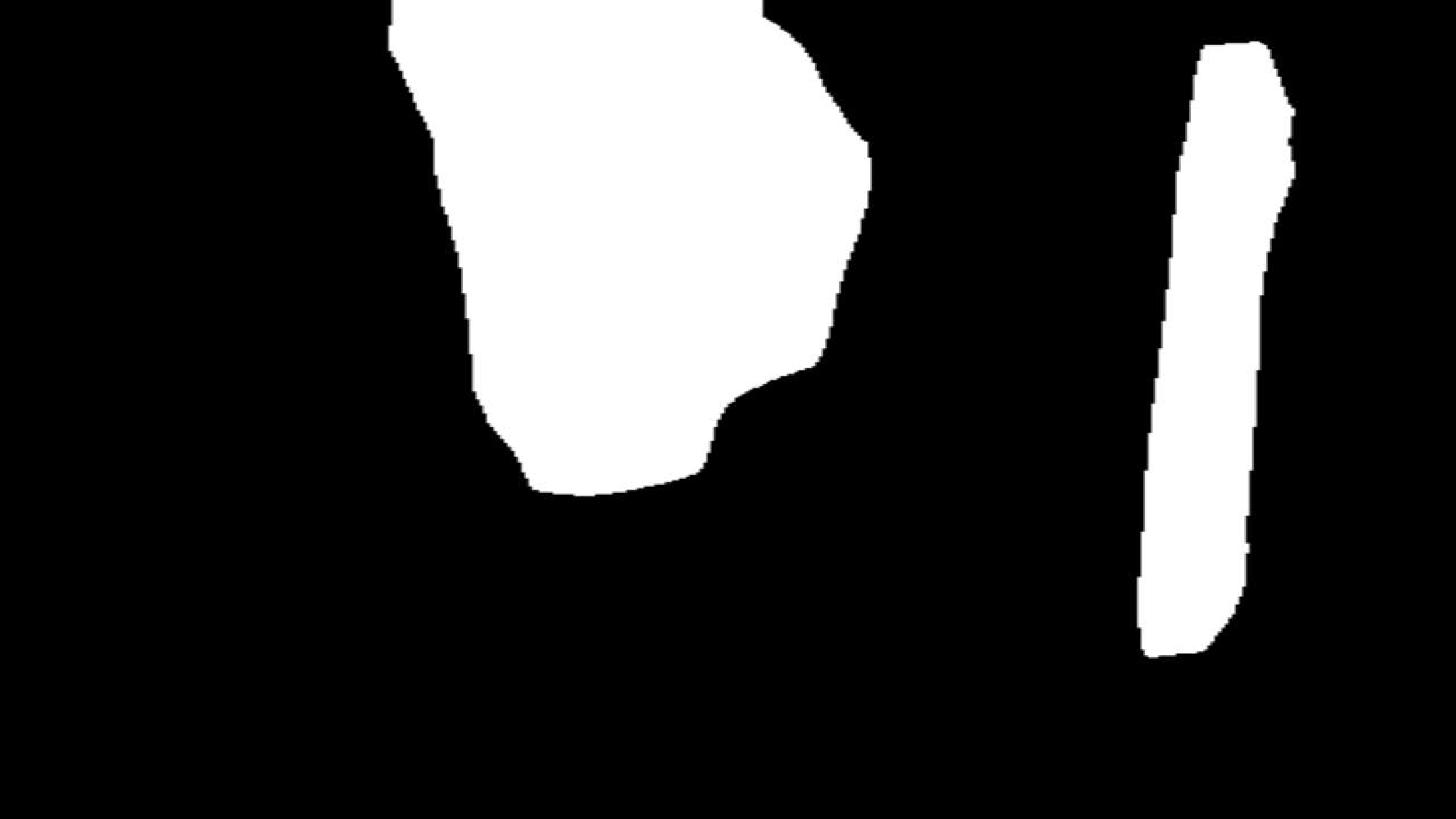}&
      \includegraphics[width=15mm, height=10mm]
      {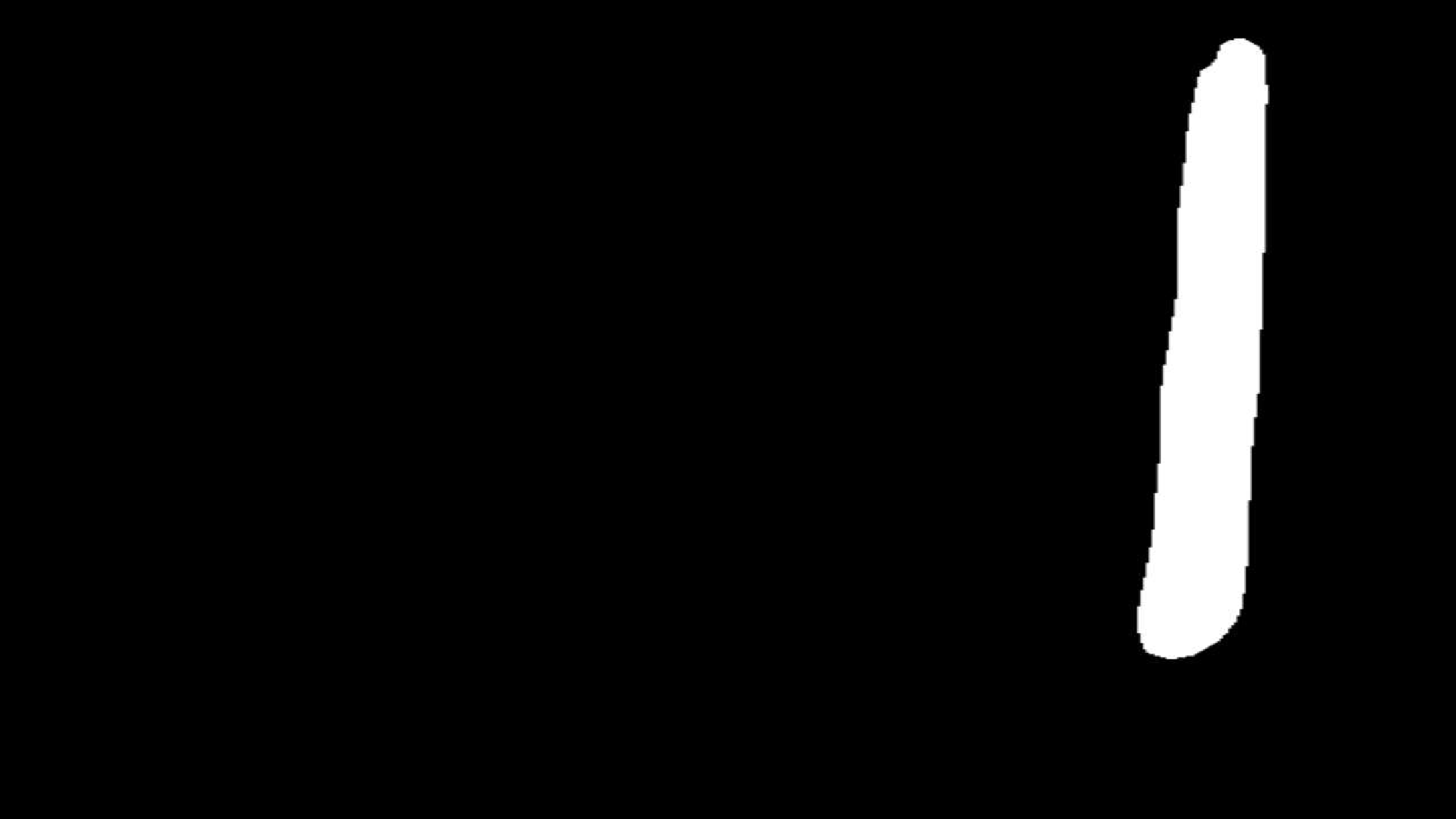}\\ 
      \includegraphics[width=15mm, height=10mm]
      {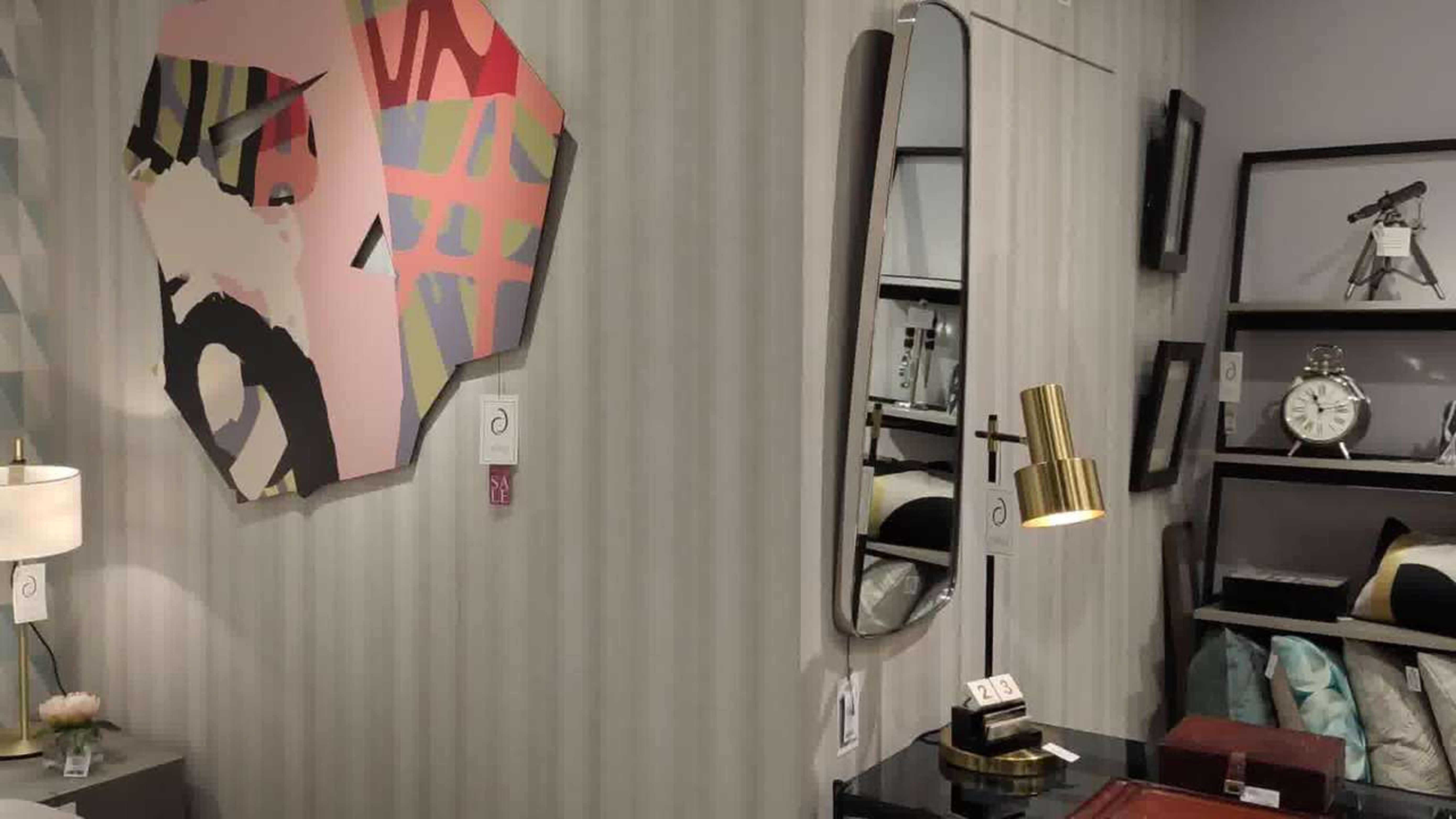}&
      \includegraphics[width=15mm, height=10mm]
      {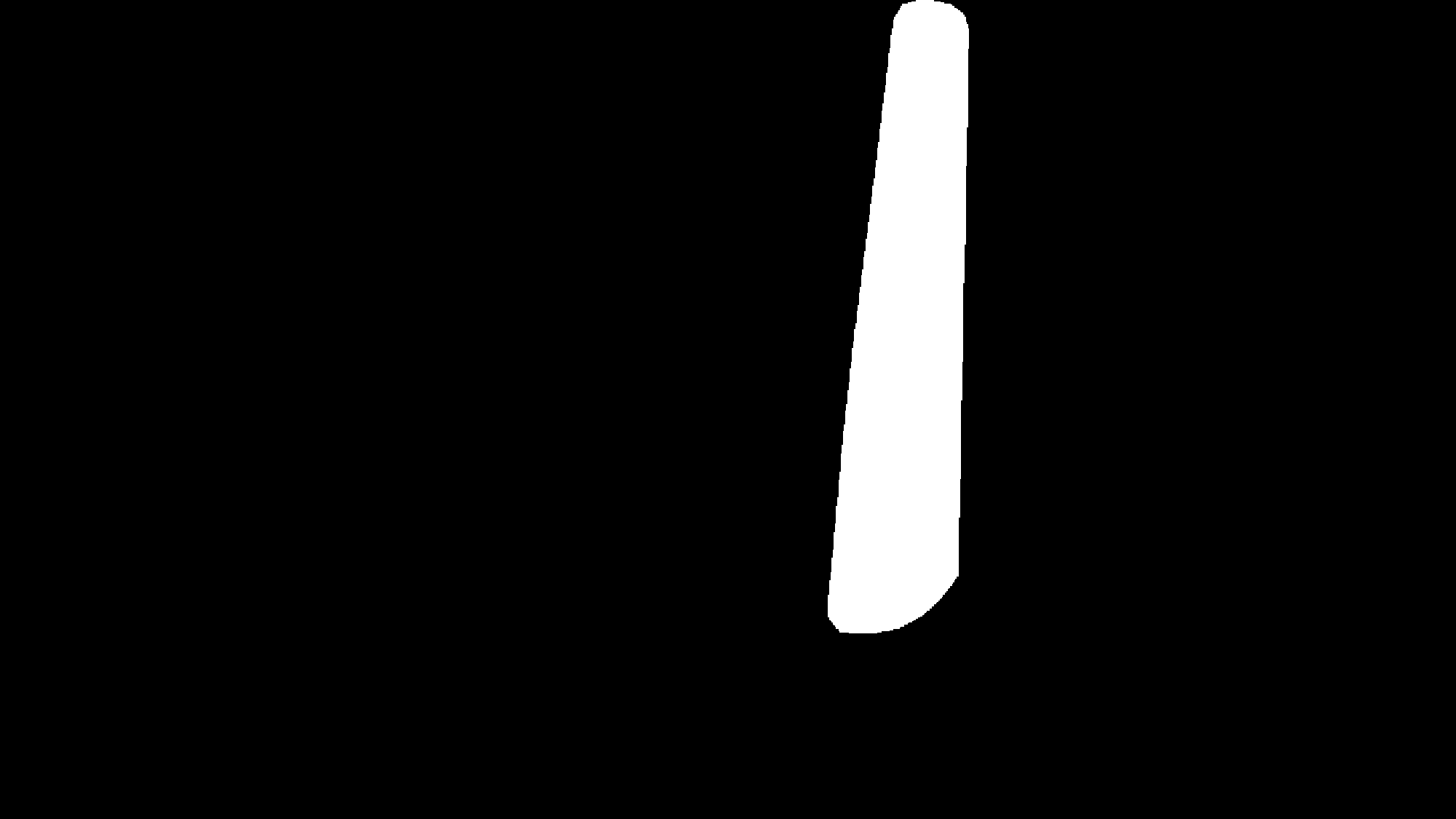}&
      \includegraphics[width=15mm, height=10mm]
      {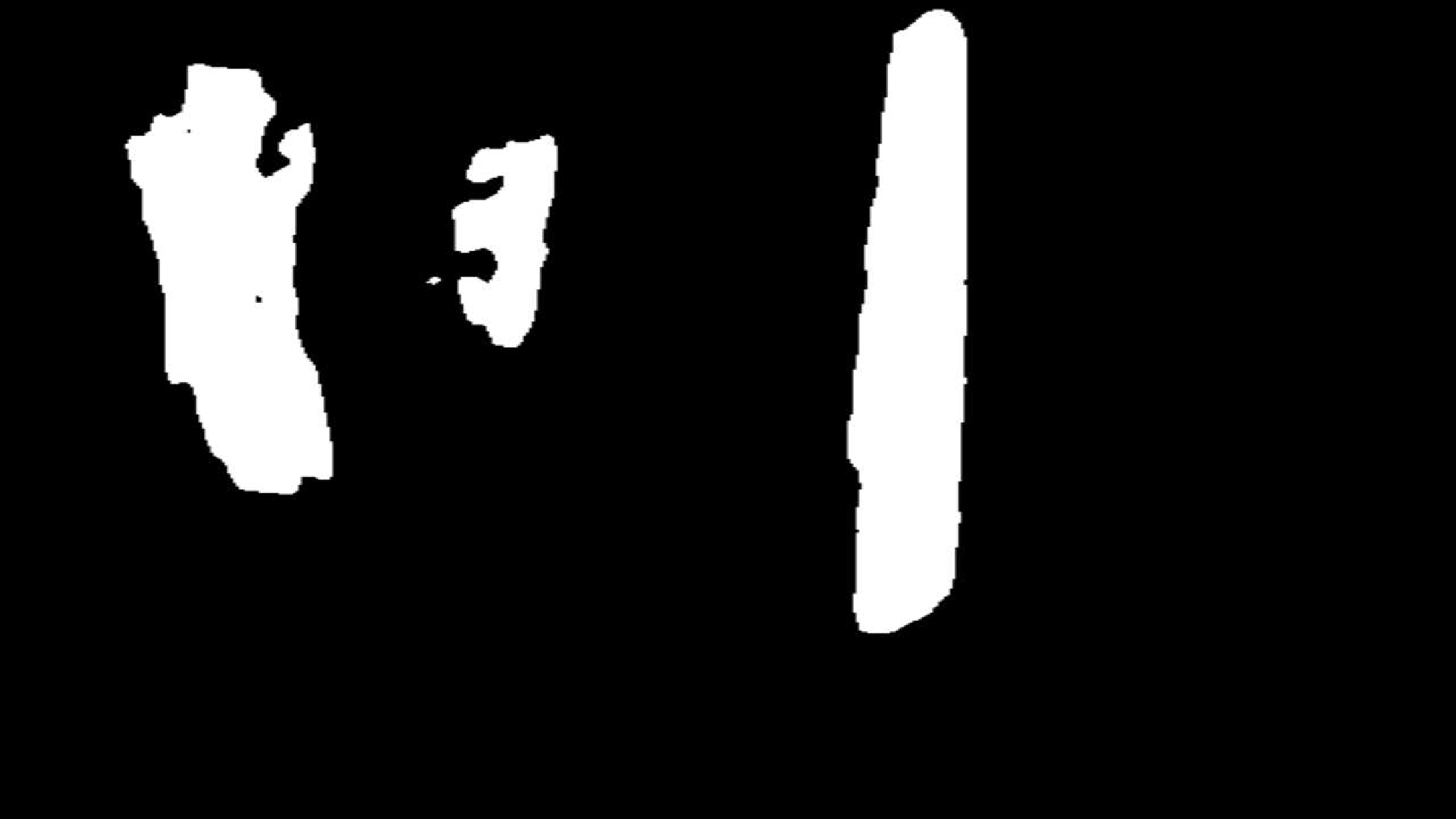}&
      \includegraphics[width=15mm, height=10mm]
      {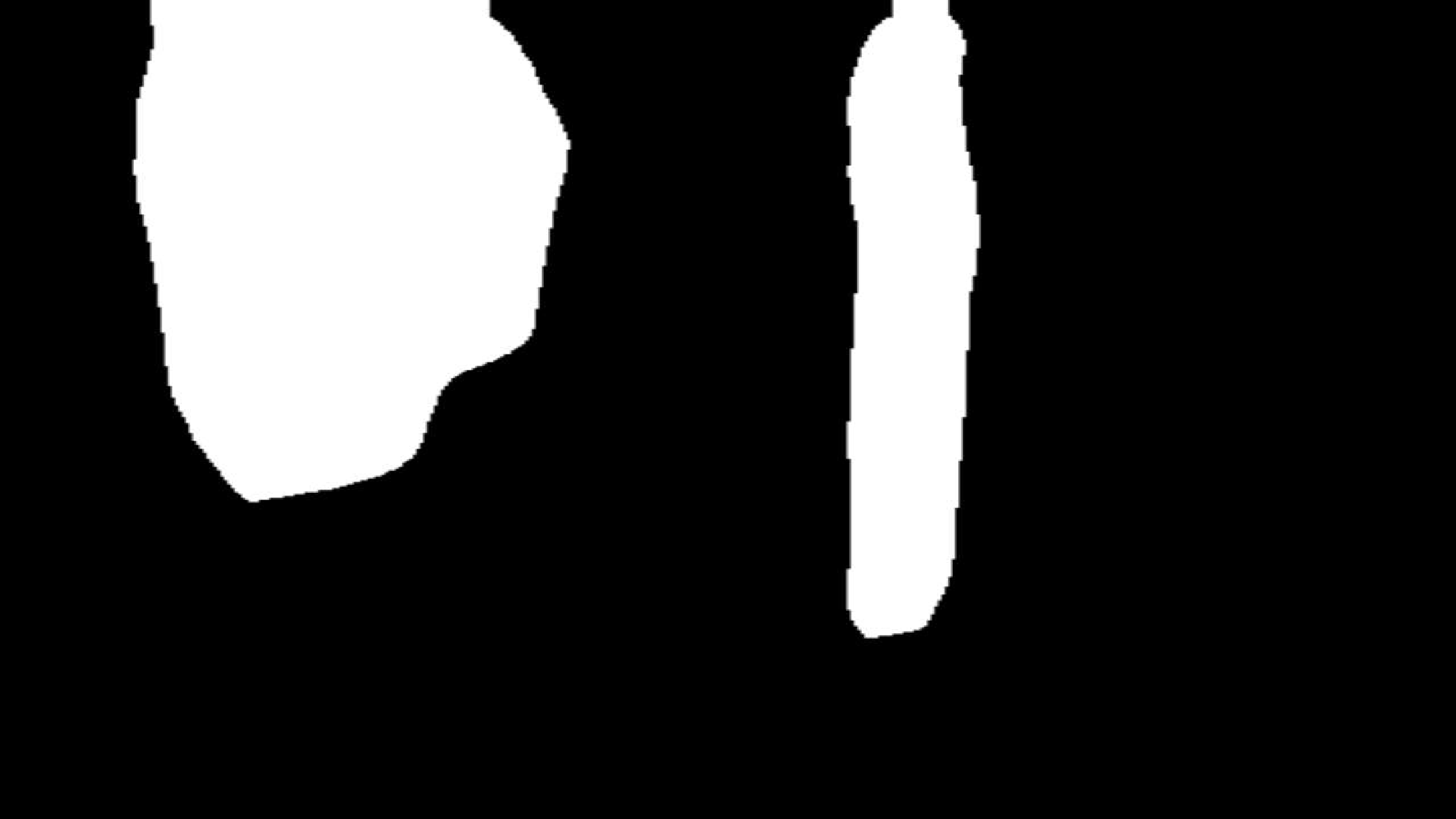}&
      \includegraphics[width=15mm, height=10mm]
      {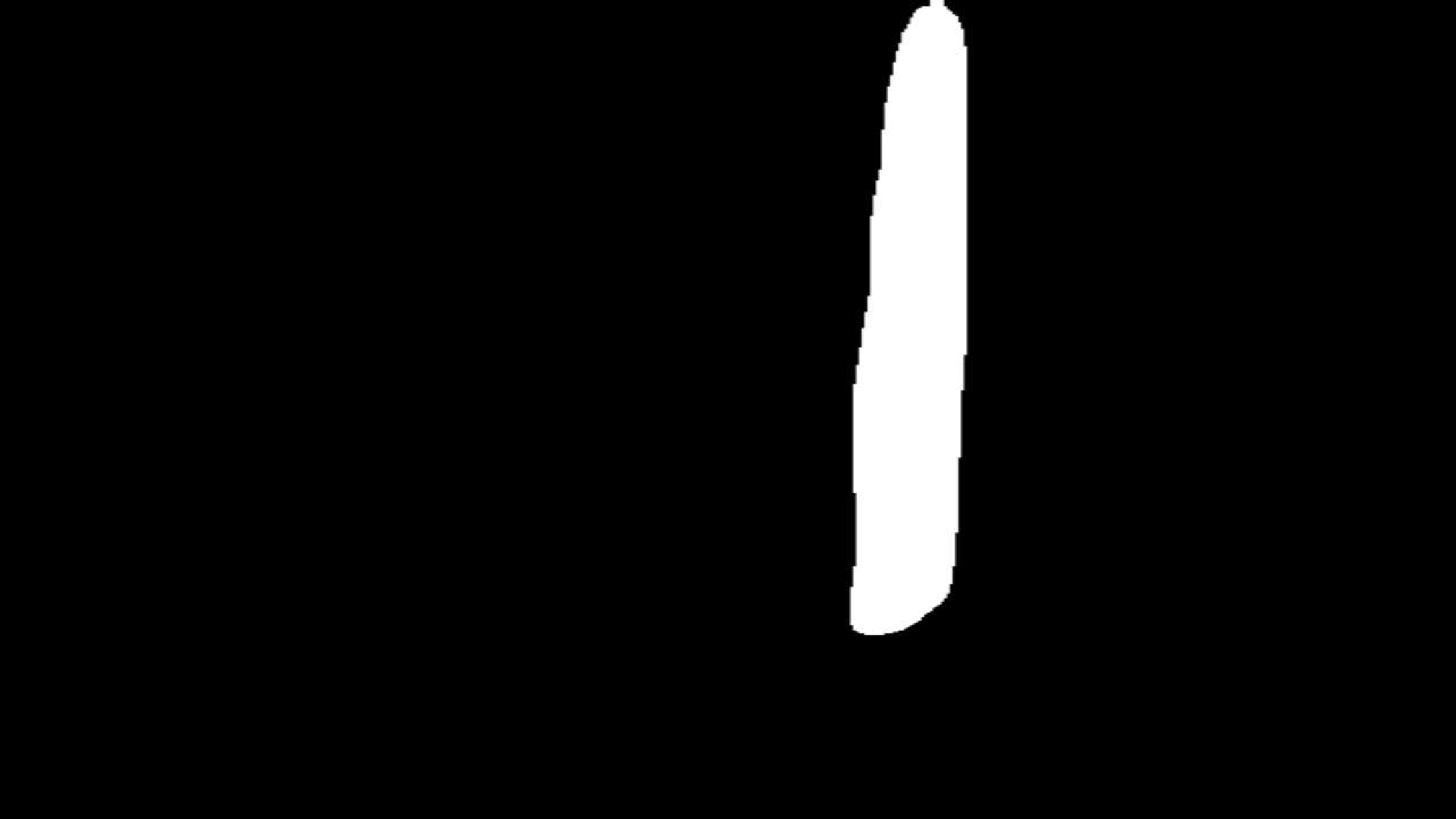}\\
    \centering Image & \centering GT & \centering HetNet&\centering VMDNet&\centering Ours
    \end{tabular}
\caption{Two normal scenarios where existing methods \cite{b4} \cite{b5} fail. HetNet\cite{b4} is a single-image mirror detection method, and VMDNet \cite{b5} is designed for video mirror detection. Compared to HetNet and VMDNet, our method can detect the mirror regions correctly by fusing short-term information and long-term information.}
\label{fig:intro_comparision}
\end{figure}

Research on detecting mirrors from static images has witnessed rapid growth in recent years. Existing methods exploit context contrast \cite{b6}, reflection relation \cite{b7}, semantic relation \cite{b8}, depth information~\cite{b9,b10,b11}, visual chirality \cite{b12} and symmetry relation \cite{b13} to detect mirrors. However, these methods detect mirrors from single input images. To detect mirrors from videos requires further consideration of temporal consistency between frames. Recently, Lin \textit{et al.} \cite{b5} proposed the first video mirror detection model, VMDNet, which extracts correspondence between the mirrors and the surroundings at both the intra-frame and the inter-frame levels. However, this method relies on the extraction of the correspondence and may fail when the correspondence cannot be established.
For example, the top two rows in Fig.~\ref{fig:intro_comparision} show the same mirror hanging on the wall. However, the VMDNet is confused by the different mirror reflections in the two frames, and cannot detect the mirror correctly. Moreover, the VMDNet will predict other objects as mirrors since it separately considers correspondences at the short-term and long-term levels. For example, the bottom two rows in Fig.\ref{fig:intro_comparision} show that the VMDNet fails to distinguish the painting and mirror, as correspondences for both of them are extracted.

\begin{figure*}[htb]
\begin{center}
  \includegraphics[width=0.85\textwidth]{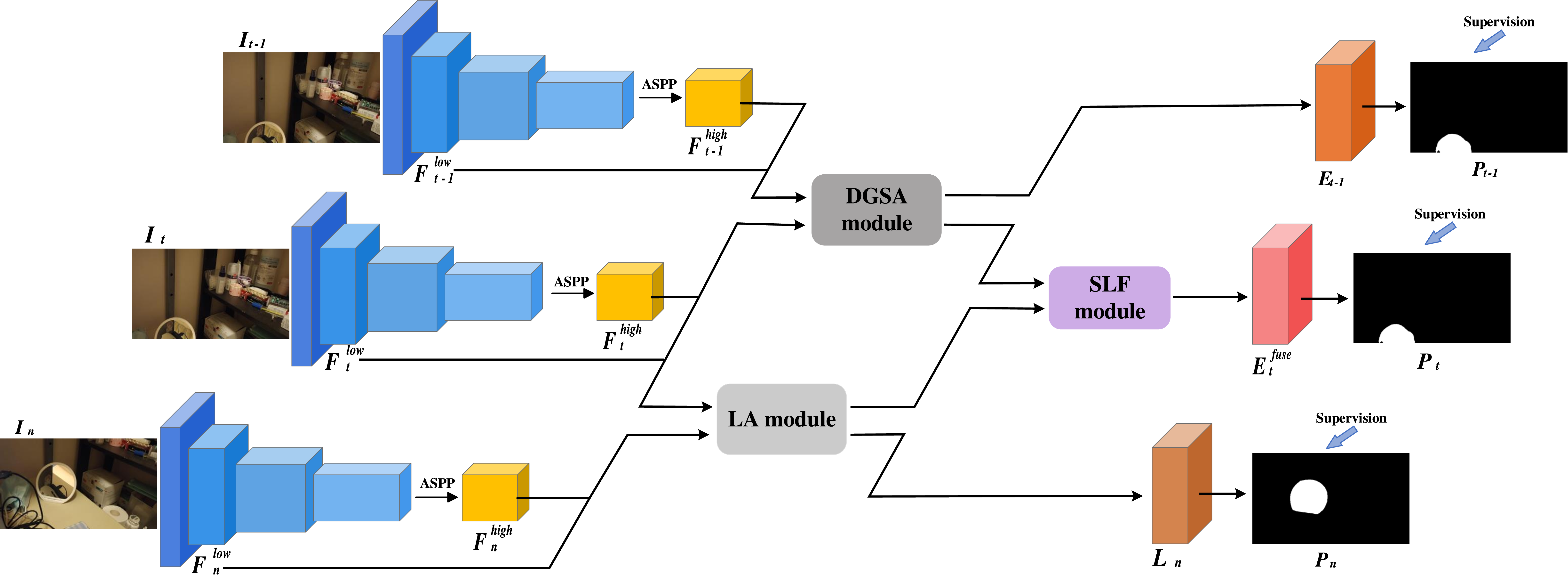}
  \caption{The architecture of our proposed model. We first feed three frames from the same video to the backbone feature extractor, then the DGSA module to extract appearance features from adjacent frames, and an LA module to extract context features from long video clips parallelly. Second, the SLF module fuses short-term attention and long-term attention to finalize the mirror region.} \label{fig:architecture}
\end{center}
\end{figure*}

To address the above problems, we propose a novel approach to detect mirrors 
in videos. We observe that humans can recognize the appearance (e.g., shape, color) of candidate mirrors from just one or two frames. However, to make sure that the candidate is indeed a mirror (not a picture or window), we often need to see more frames to have a global view. This observation motivates us to extract appearance features at a micro (short video clips) and to extract context features at a macro view (long video clips), and then combine them to predict the mirror. Our approach is different from VMDNet, which utilizes long-term information as an auxiliary task in the first stage and separately considers short-term and long-term information in the second stage. Our approach tries to combine short-term information and long-term information to better predict the mirror map.

Our method consists of three modules:
1) a Dual Gated Short-term Attention (DGSA) module to extract appearance features from adjacent frames; 
2) a Long-term Attention (LA) module to extract context features from long video clips to obtain position information of mirrors; 
and 3) a Short-Long Fusion (SLF) module to fuse appearance features and context features to finalize the mirror region.

To evaluate the performance of video mirror detection, we also construct a challenging benchmark dataset that includes a variety of scenes from real living and working environments. 
Most of our data are from two public datasets: NYUv2 \cite{b14}, ScanNet \cite{b15}, and others are captured by ourselves. Our dataset has 19,255 frames from 281 videos with pixel-wise annotation. Compared to the first video mirror dataset (VMD) proposed by~\cite{b5}, of which 95\% are collected from furniture stores, our dataset covers about 20 scene types (e.g., gym, lift, kitchen) which largely increases the diversity of data. 

Our contributions can be summarized as:
\begin{itemize}
\setlength{\itemsep}{0pt}
\setlength{\parsep}{0pt}
\setlength{\parskip}{0pt}
  \item We propose a novel transformer network for video mirror detection. It consists of three modules (DGSA module, LA module, and SLF module) to support the extraction and fusion of short-term and long-term attention to improve video mirror detection.
  \item We construct a 
  challenging benchmark dataset that contains 19,255 frames from 281 videos and pixel-wise annotations from a variety of real-world scenes.
  \item We have conducted extensive experiments on both the VMD dataset and our dataset to demonstrate that our method achieves state-of-the-art performance.
\end{itemize}

\section{Related Work}
\label{sec:related}
\subsection{Mirror Detection}
In recent years, many works~\cite{b6,b7,b8,b12,b4,b9,b10,b11,b13,b29}, are proposed to detect mirrors from single images. 
They exploit context contrast \cite{b6}, reflection relation \cite{b7}, semantic
relation \cite{b8}, depth information \cite{b9, b10, b11}, visual chirality
\cite{b12}, symmetry relation \cite{b13}, shape\cite{b29} and self-supervised pre-training \cite{b30} to detect mirrors.
Although the single-image mirror detection model achieves reliable results, their performance on video is not good because of insufficient exploitation of temporal information. Recently,  
Lin \textit{et al.} \cite{b5} propose the first video detection network, named VMDNet. It focuses on extracting mirror correspondence at intra-frame and inter-frame levels. 

\subsection{Mirror Detection Dataset}
There are two mirror datasets for image mirror detection. Yang \textit{et al.} \cite{b6} propose the first image mirror dataset, MSD. It contains 4,018 images and corresponding annotations. However, lots of images in MSD are very similar. Thus, Lin \textit{et al.} \cite{b7} propose the PMD dataset which contains 6,461 images from six public datasets. It covers a variety of scenes. These two image mirror datasets help to accelerate the development of the image mirror detection task. For the video mirror detection, Lin \textit{et al.} \cite{b5} propose the first video mirror dataset, VMD dataset. It includes 14,988 image frames from 269 videos. 
However, more than 95\% of VMD dataset are collected from the
furniture store, which limits the diversity of the dataset. Therefore, we construct a more challenging benchmark which contains 19,255 image frames from 281 videos. Our data are from two public datasets and self-capture videos. The data collected from the furniture store in our dataset is only 13\%.

\begin{figure*}[htb]
\begin{center}
  \includegraphics[width=0.7\textwidth]{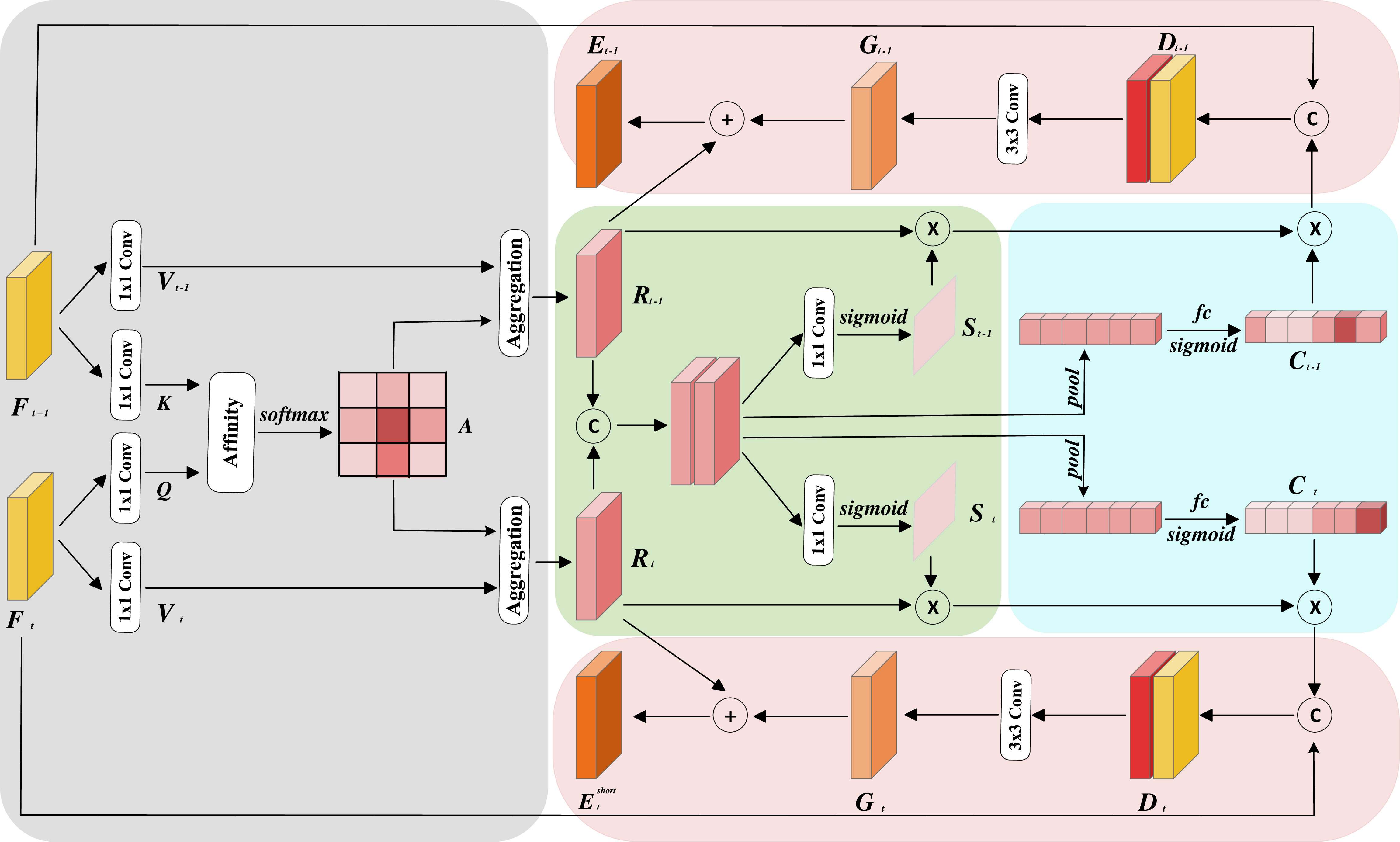}
  \caption{The schematic illustration of Dual Gated Short-term Attention (DGSA) module. The grey part represents the short-term attention (SA) block. Pink parts represent the fusion blocks. The green and blue parts represent the spatial-wise gate (SG) block and the channel-wise gate (CG) block, respectively.} \label{fig:DGRA}
\end{center}
\end{figure*}

\section{Method} 
\subsection{Overall Structure}
Fig.~\ref{fig:architecture} shows the architecture of the proposed FusionFormer. To enable extraction and fusion of short-term and long-term attention, our model first takes three frames from the same video as input. Two are adjacent frames and the third $I_{n}$ 
is a random other frame. Then, we employ the Mix Transformer \cite{b16} as the encoder, which can encode long-range dependencies. Adhere to the approach 
in \cite{b5}, we 
utilize the second scale for the low-level features ($F_i^{low}$) and the fifth scale after the atrous spatial pyramid (ASPP) for the high-level features $F_i^{high}$. 
After obtaining features from three input frames, we assign the Dual Gated Short-term Attention (DGSA) module on the low-level $F_{i\in\{t-1,t\}}^{low}$ and high-level features $F_{i\in\{{t-1,t}\}}^{high}$ to extract appearance features from adjacent frames, and the Long-term Attention (LA) module on low-level features $F_{i\in\{t,n\}}^{low}$ and high-level features $F_{{i\in\{{t,n}\}}^{high}}$ to extract context features from long video clips at the same time. Finally, the Short-Long Fusion (SLF) module combines appearance features and context features selectively to produce the final mirror prediction $P_t$.

\subsection{Dual Gated Short-term Attention Module}
The DGSA module aims to extract appearance features from the short-term information. It is inspired by the cross-attention module proposed in 
\cite{b5}, which can extract correspondences between the content inside and outside of the mirror at the intra-frame level and the inter-frame level. However, occlusions, appearance changes, etc., may affect the correspondence extraction. 
Therefore, we propose to weigh the mirror correspondence features differently. 

Fig.\ref{fig:DGRA} shows the schematic illustration of the DGSA. We use $F_{i\in\{t-1, t\}}$ to denote $F_{i\in\{t-1, t\}}^{low}$ or $F_{i\in\{t-1, t\}}^{high}$ to be visual clear. Our DGSA module consists of four blocks: a short-term attention (SA) block, a spatial-wise gate (SG) block, a channel-wise gate (CG) block, and two fusion blocks. Given the input 
features  $F_{i\in\{t-1, t\}}$, we first use the SA block to extract short-term correspondence features (denotes $R_{t-1}$, $R_{t}$):
\begin{equation}
    \begin{aligned}
        R_{t-1} = \omega_{t-1} \sum_{i}^{(2H+2W-1)\times(W\times H)} AV_{t-1},
    \end{aligned}
    \label{eq:RA_4}
\end{equation}
\begin{equation}
    \begin{aligned}
        R_t = \omega_t \sum_{i}^{(2H+2W-1)\times(W\times H)} AV_{t},
    \end{aligned}
    \label{eq:RA_5}
\end{equation}
where 
$\mathbf{A}$ is the correspondence attention map. $\omega_{t-1}$ and $\omega_{t}$ are the learnable parameters. Then, we concatenate the $R_{t-1}$, $R_{t}$ and feed them to the SG block and the CG block to produce spatial gated mask $S_{t-1}$, $S_{t}$ and channel gated mask $C_{t-1}$, $C_{t}$. 

\begin{figure*}[htb]
  \centering
    \begin{minipage}[t]{0.121\linewidth}
        \centering
        \includegraphics[width=0.83in,height=0.63in]{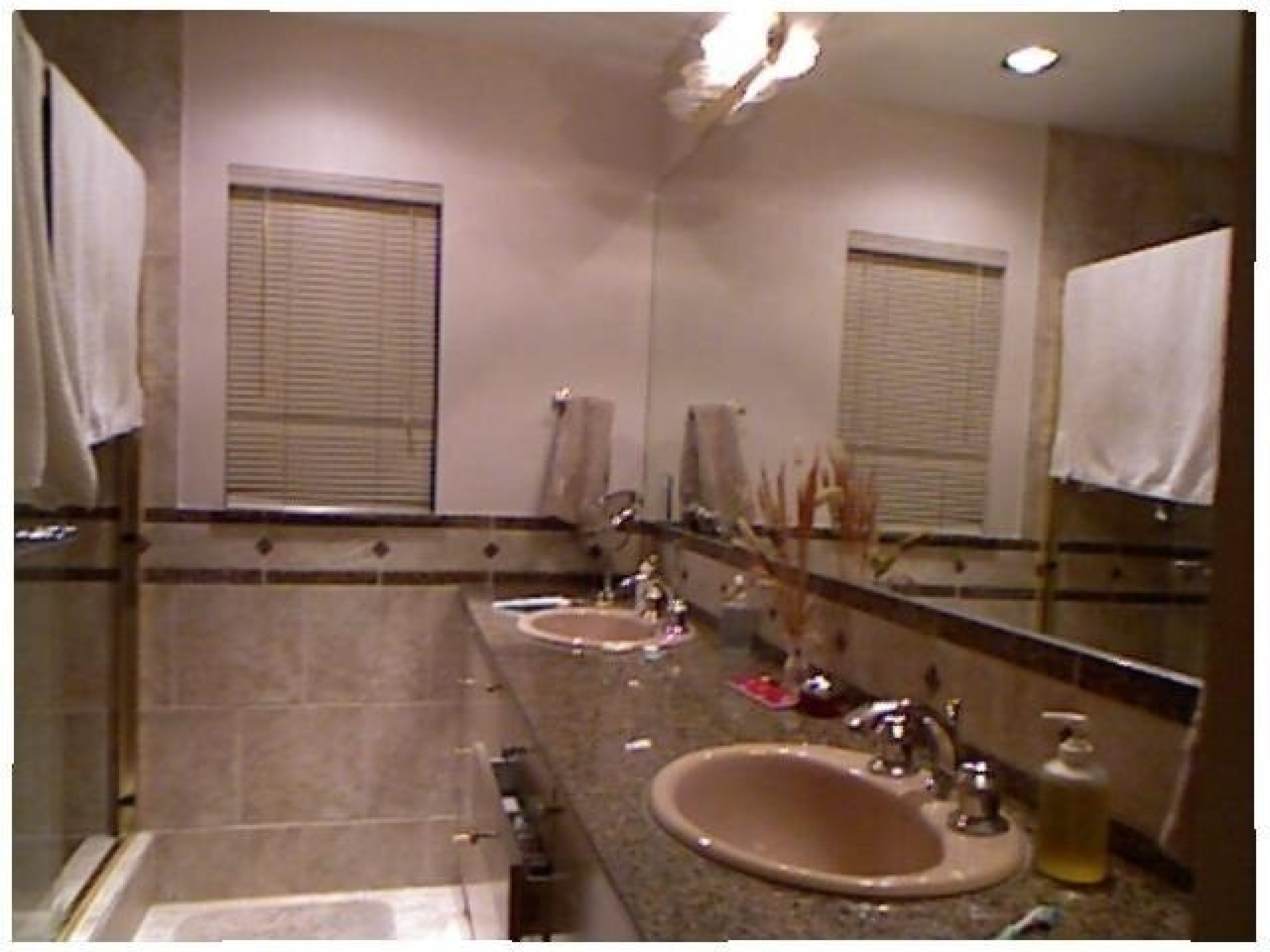}\vspace{2pt}
        \includegraphics[width=0.83in,height=0.63in]{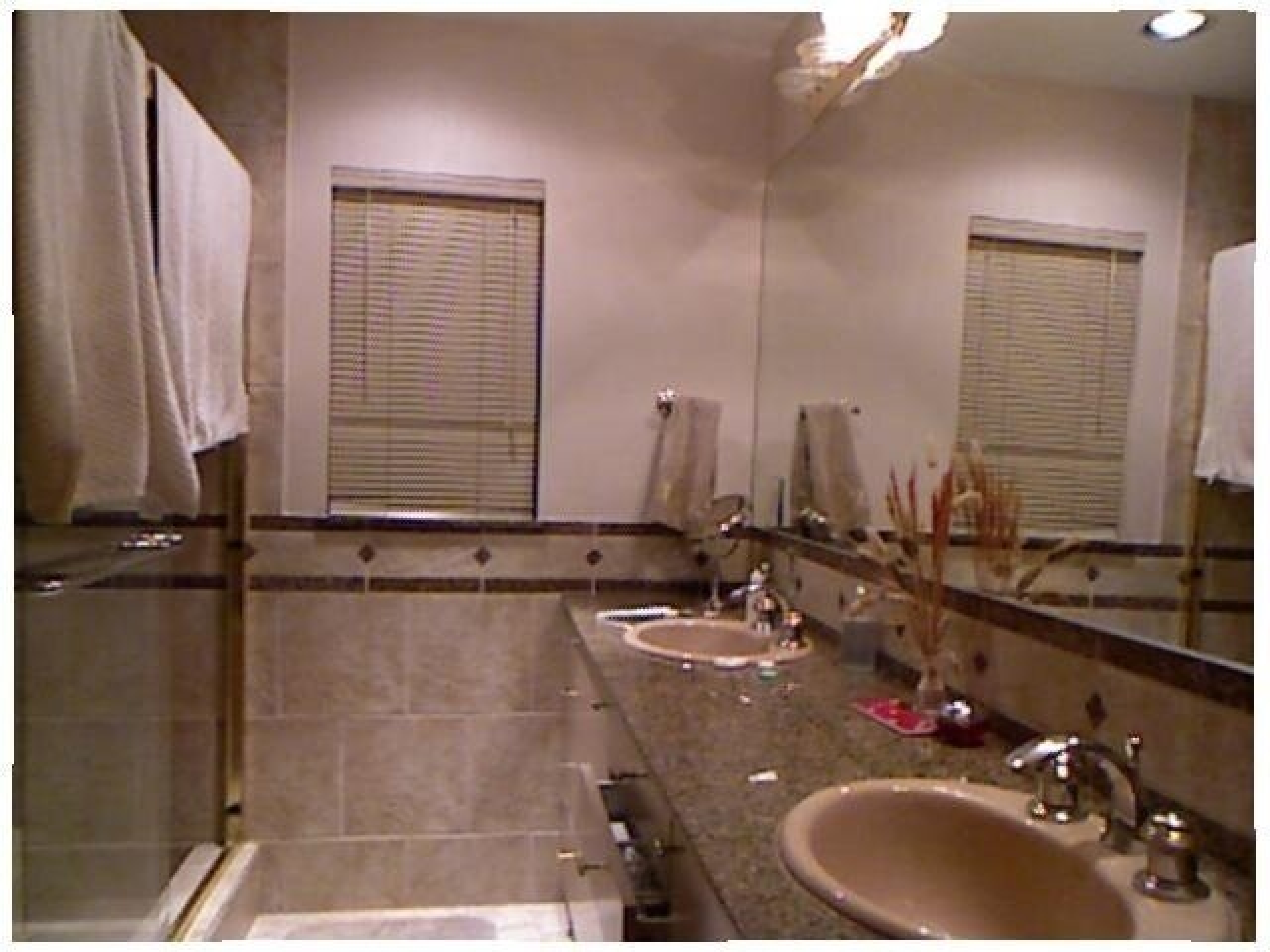}\vspace{2pt}
        \includegraphics[width=0.83in,height=0.63in]{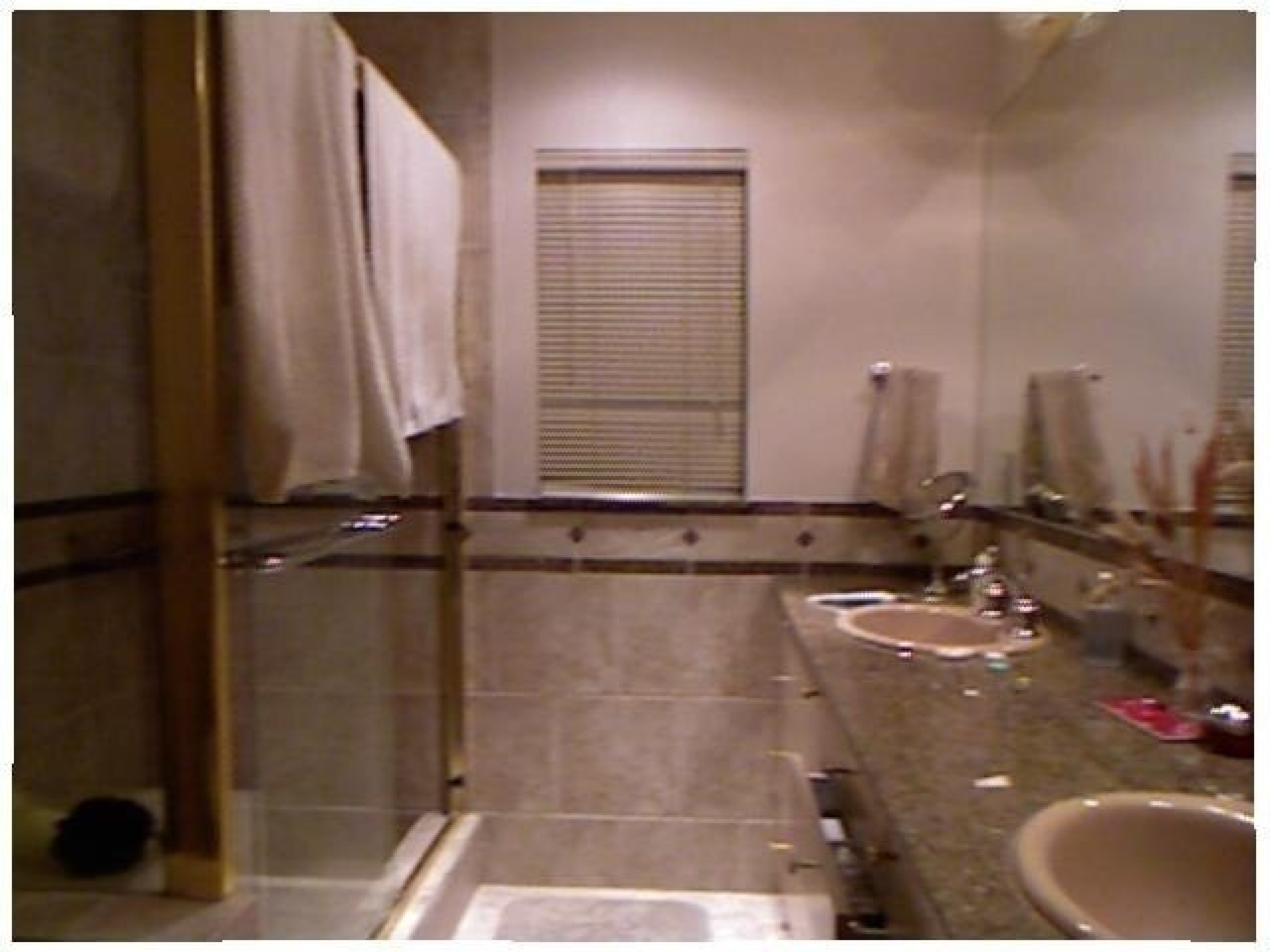}
    \end{minipage}%
    \begin{minipage}[t]{0.121\linewidth}
        \centering
        \includegraphics[width=0.83in,height=0.63in]{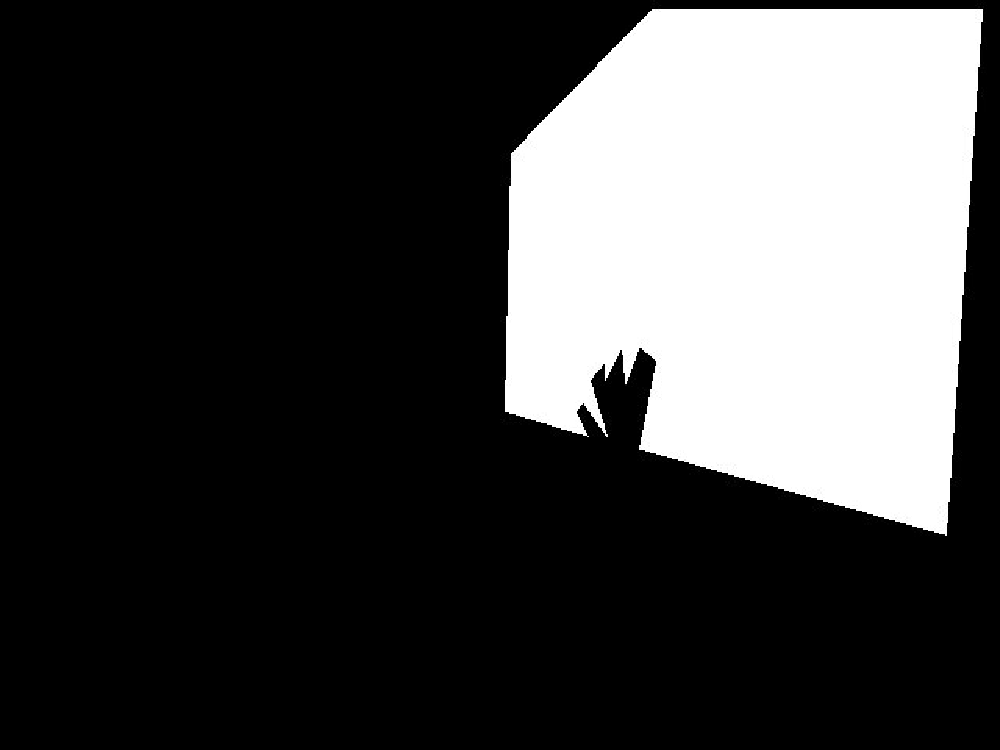}\vspace{2pt}
        \includegraphics[width=0.83in,height=0.63in]{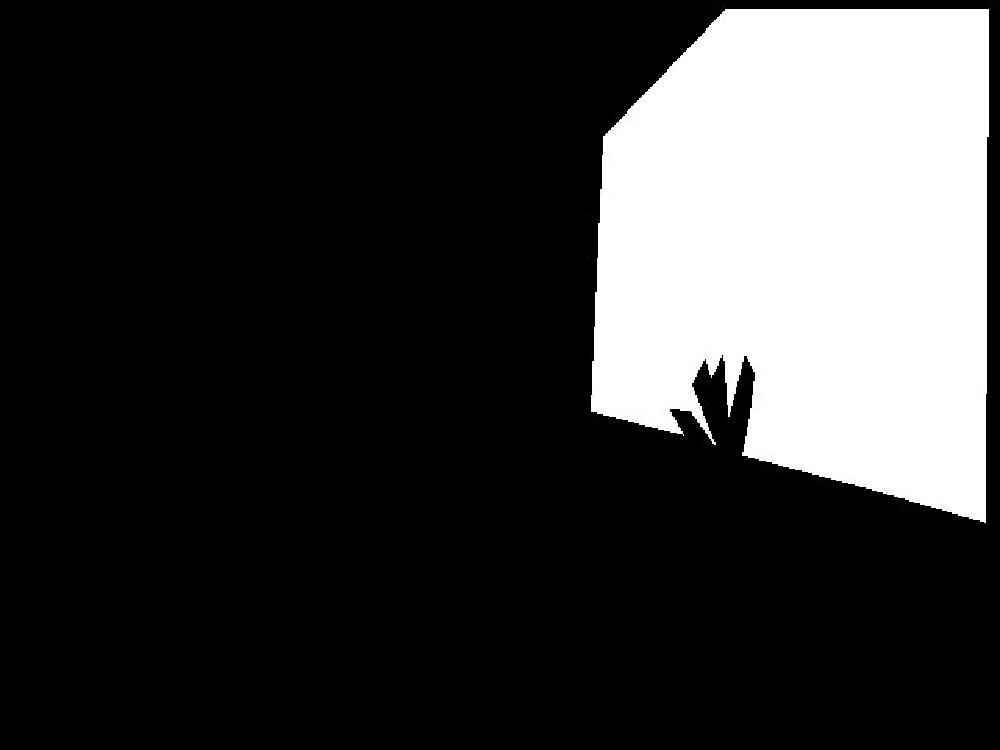}\vspace{2pt}
        \includegraphics[width=0.83in,height=0.63in]{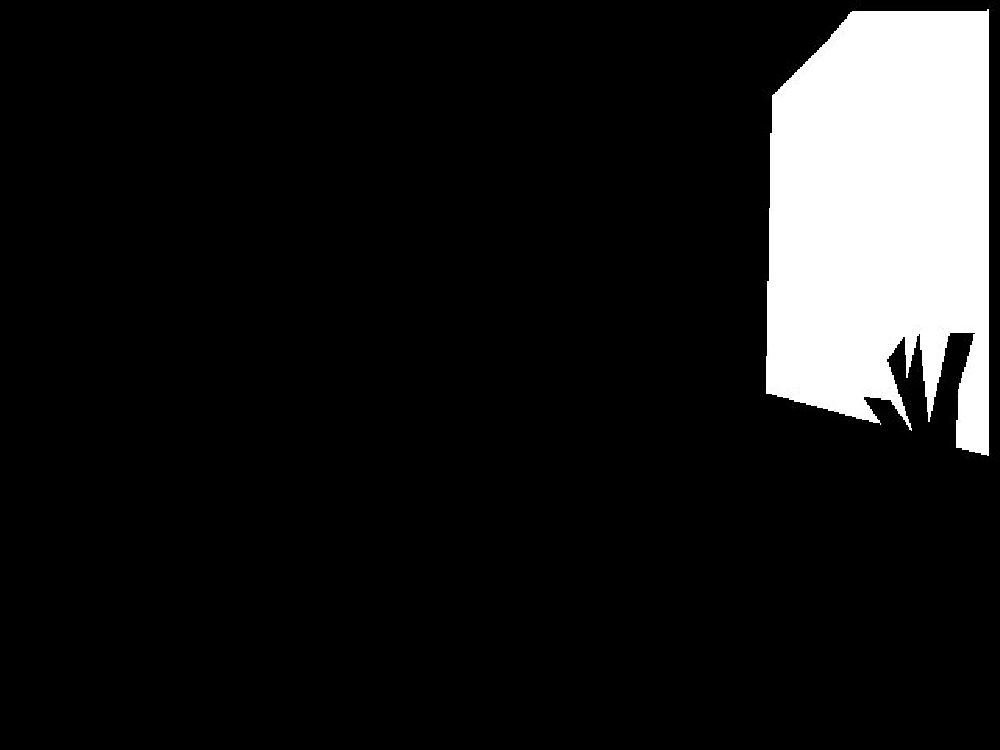}
    \end{minipage}%
    \begin{minipage}[t]{0.121\linewidth}
        \centering
        \includegraphics[width=0.83in,height=0.63in]{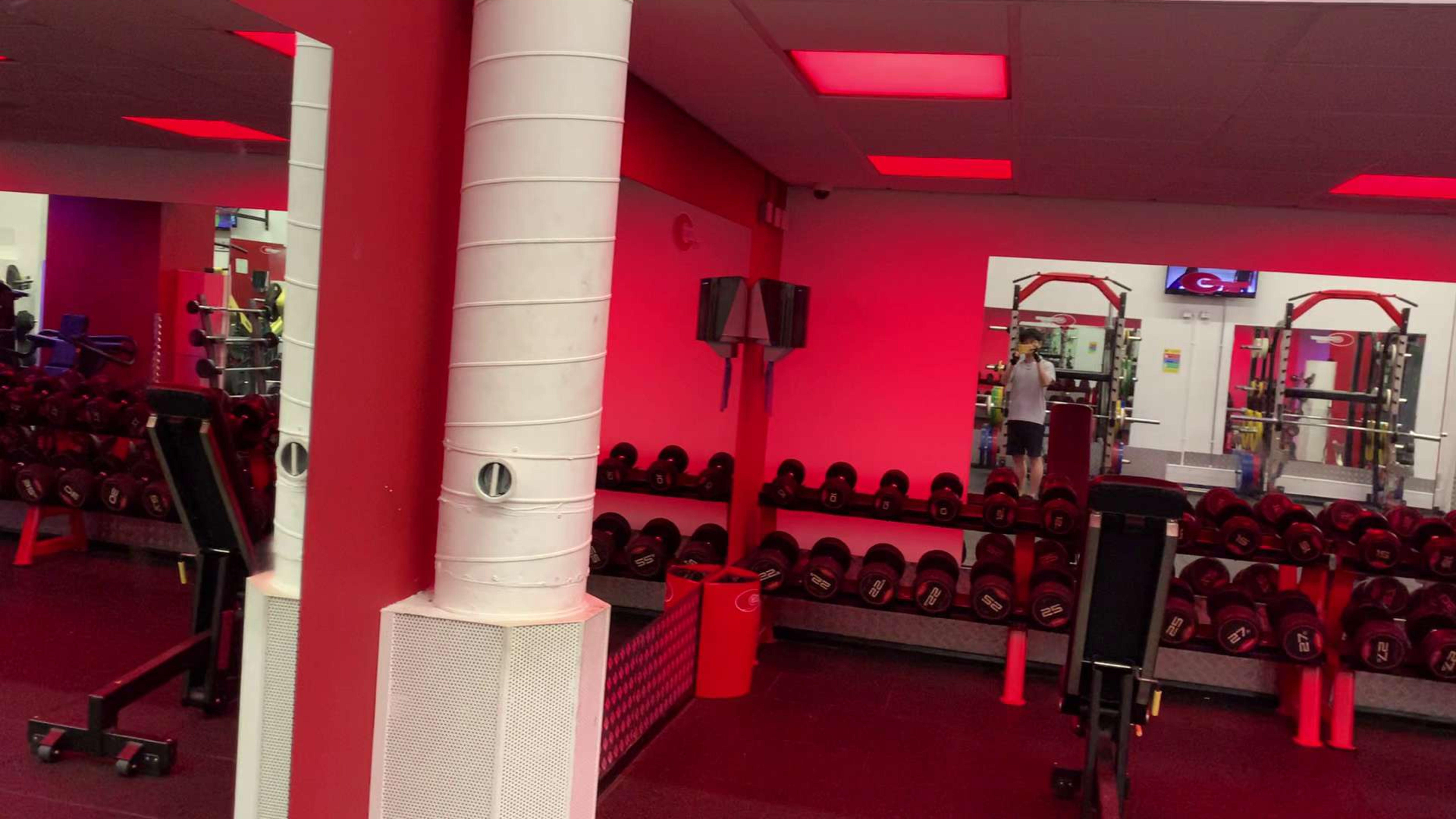}\vspace{2pt}
        \includegraphics[width=0.83in,height=0.63in]{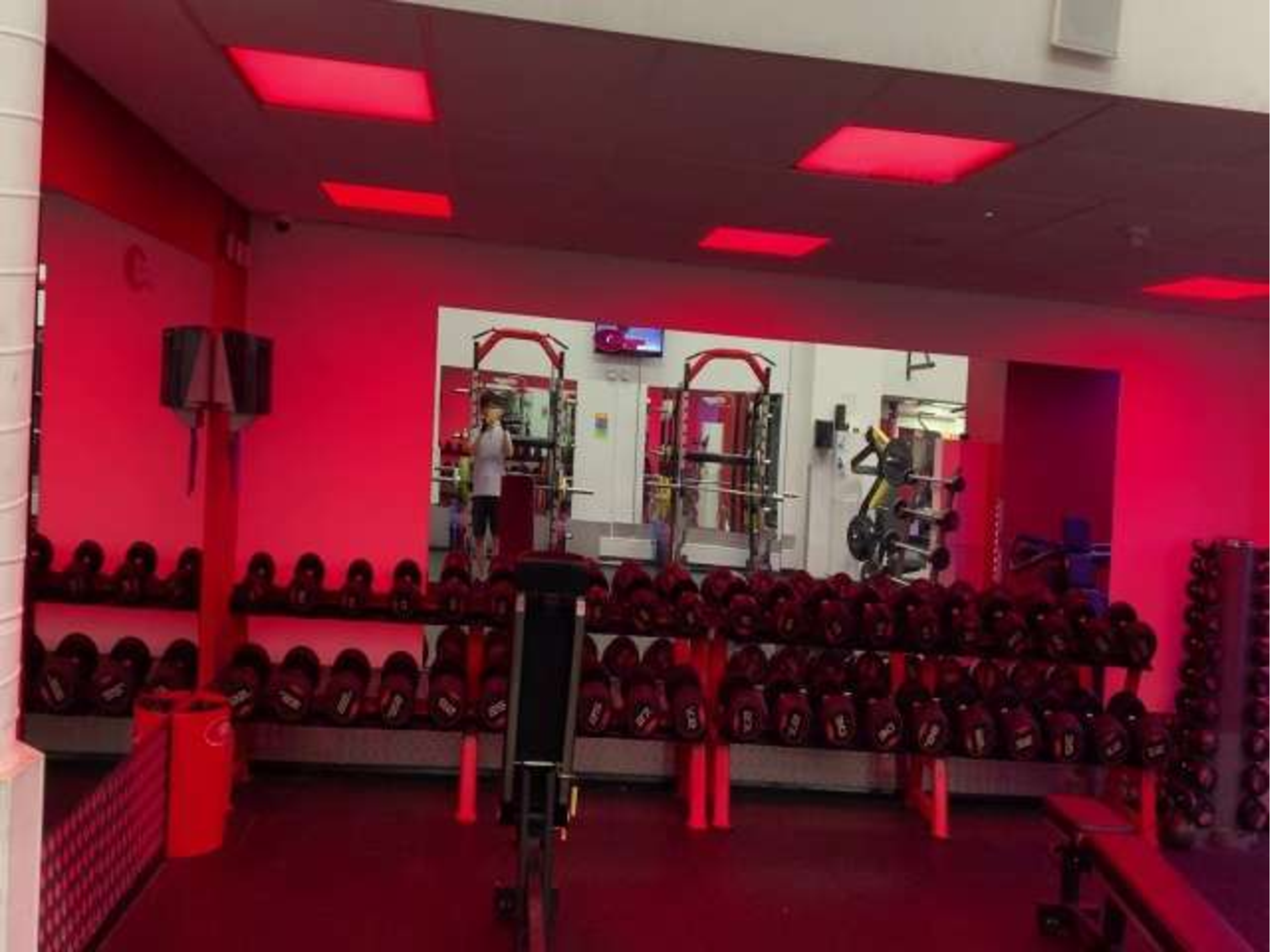}\vspace{2pt}
        \includegraphics[width=0.83in,height=0.63in]{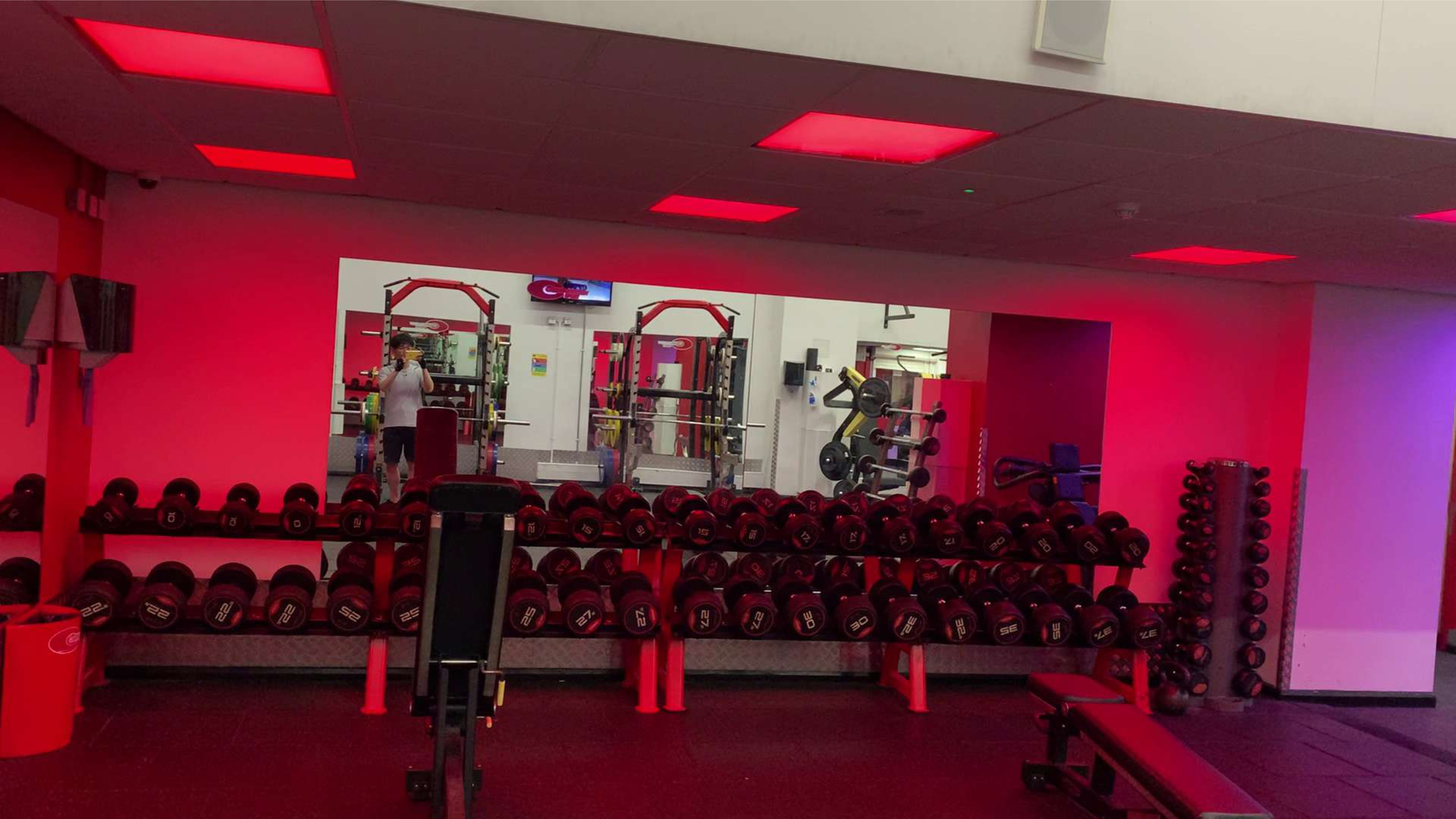}
    \end{minipage}%
    \begin{minipage}[t]{0.121\linewidth}
        \centering
        \includegraphics[width=0.83in,height=0.63in]{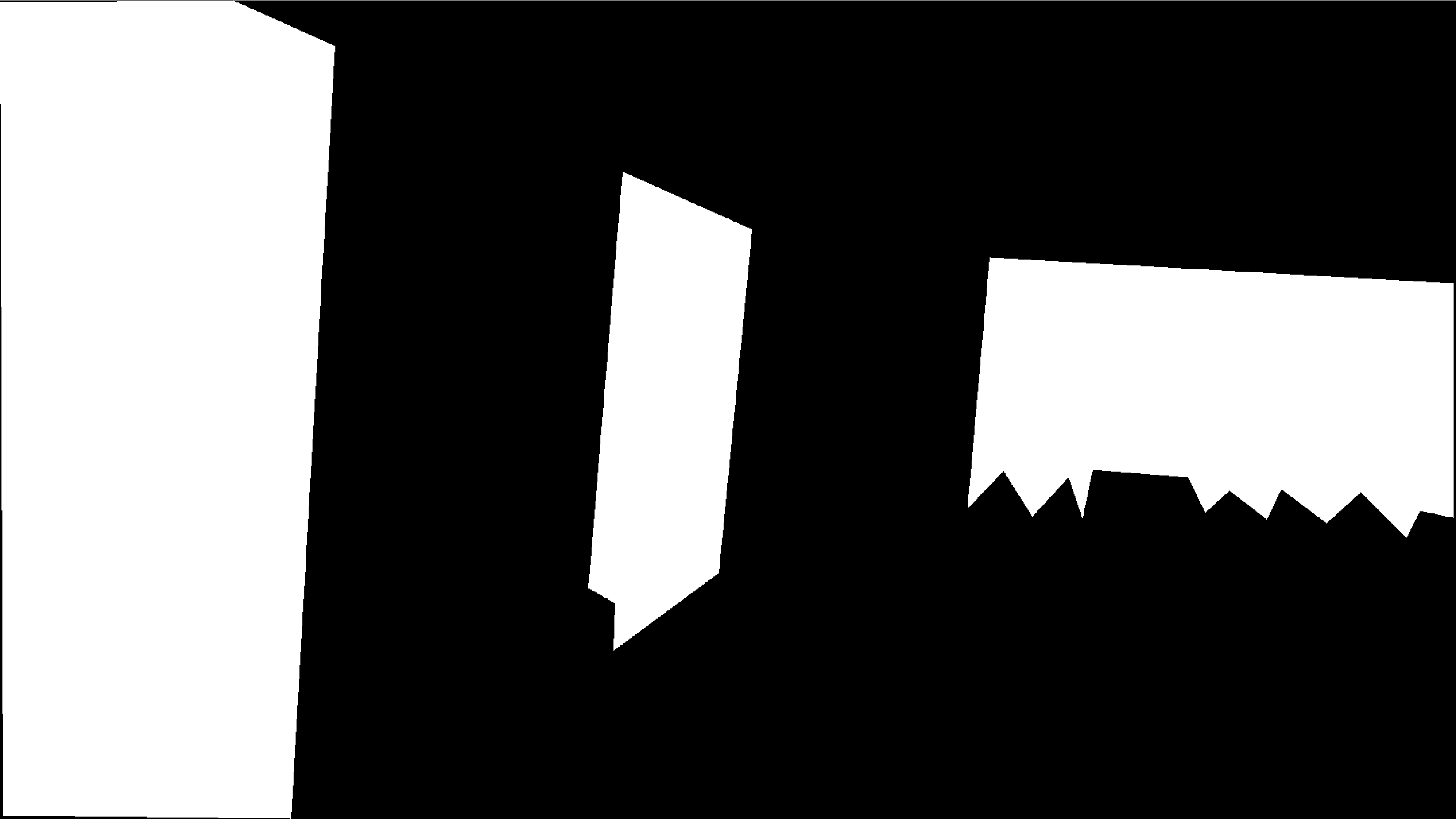}\vspace{2pt}
        \includegraphics[width=0.83in,height=0.63in]{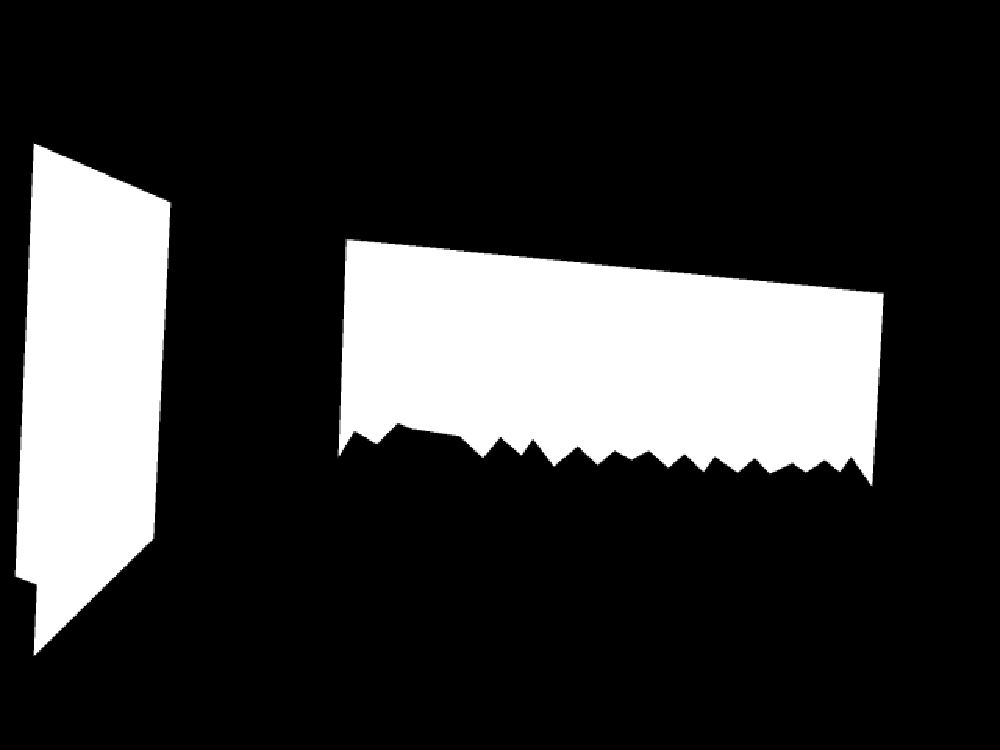}\vspace{2pt}
        \includegraphics[width=0.83in,height=0.63in]{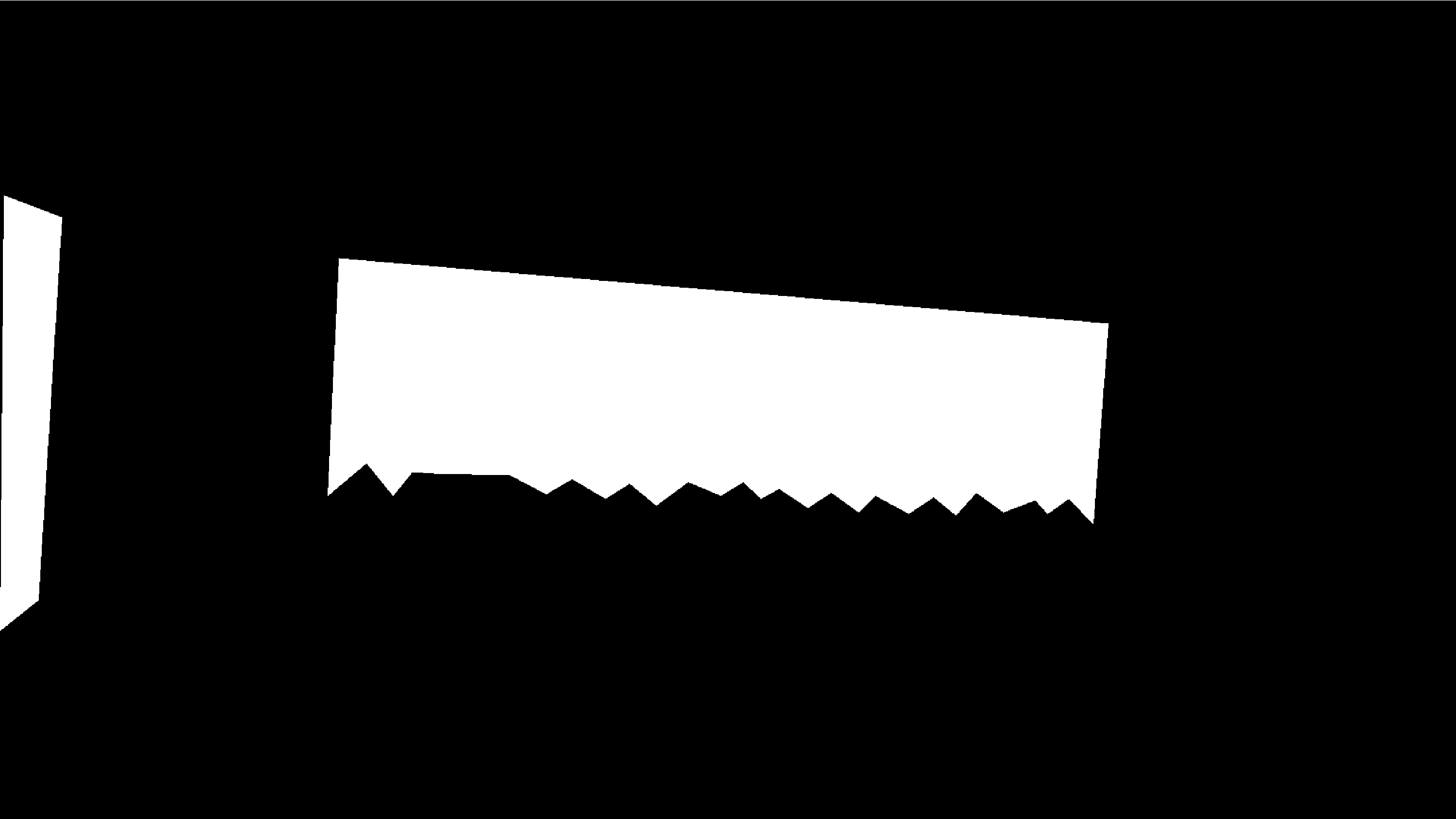}
    \end{minipage}%
    \begin{minipage}[t]{0.121\linewidth}
        \centering
        \includegraphics[width=0.83in,height=0.63in]{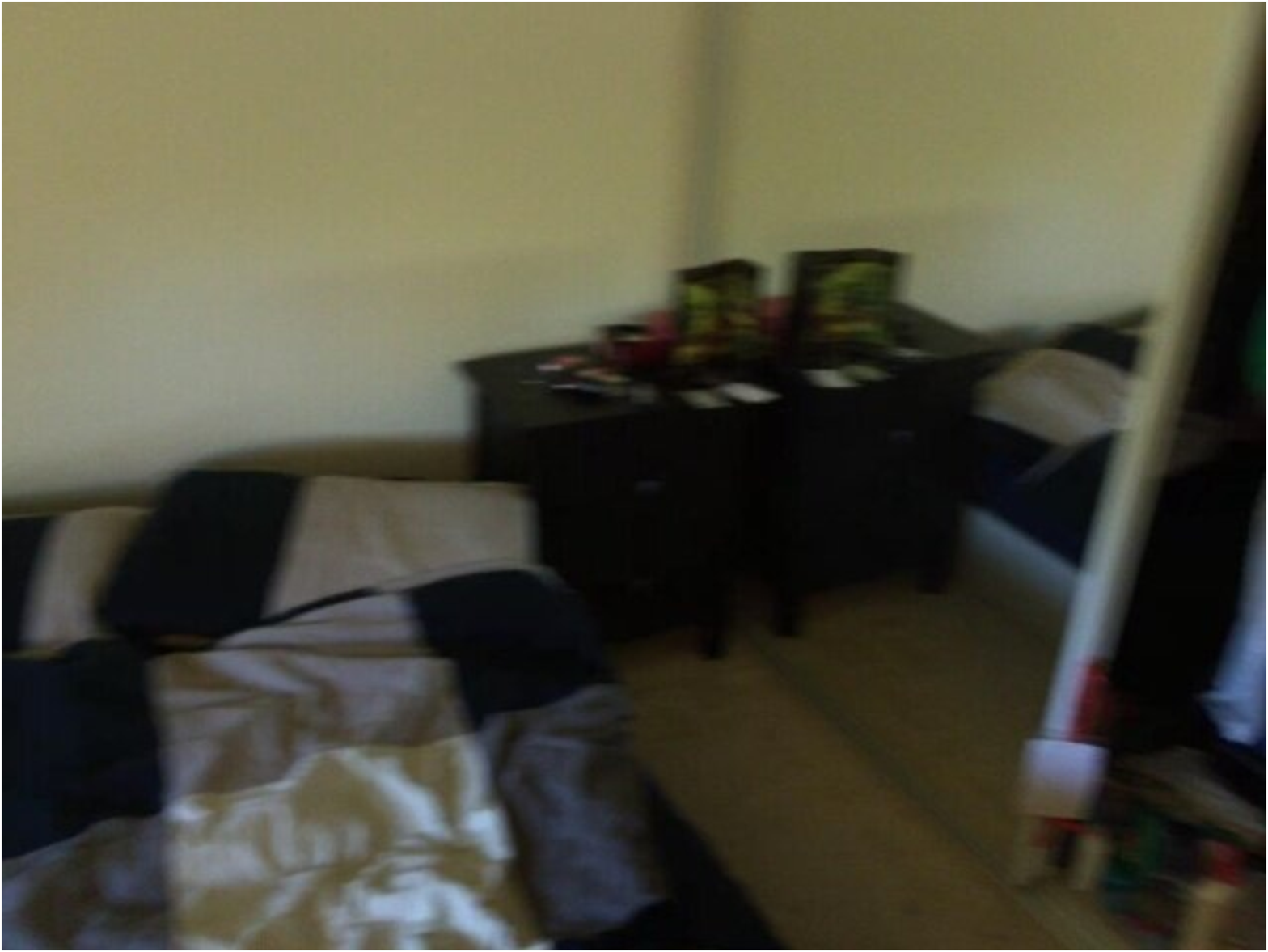}\vspace{2pt}
        \includegraphics[width=0.83in,height=0.63in]{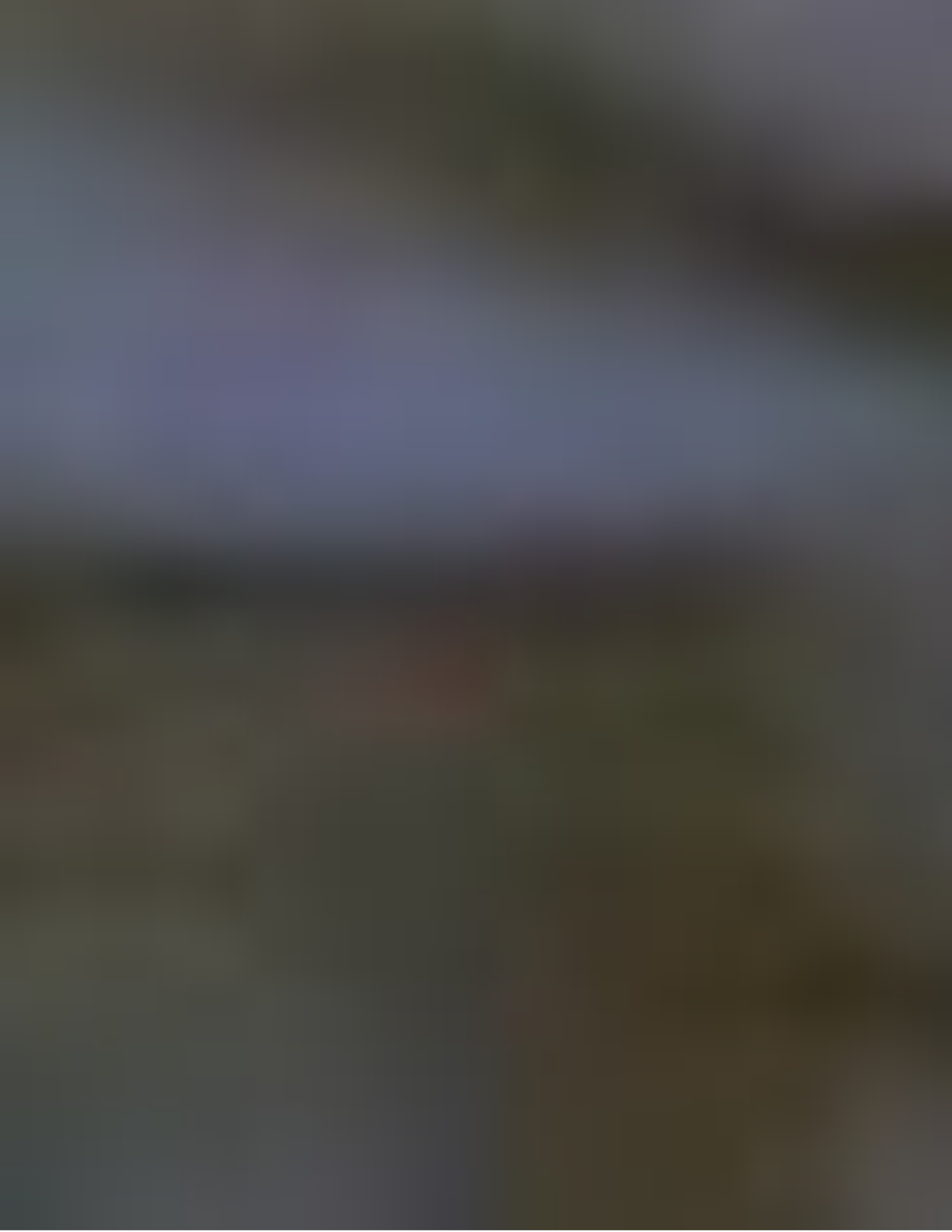}\vspace{2pt}
        \includegraphics[width=0.83in,height=0.63in]{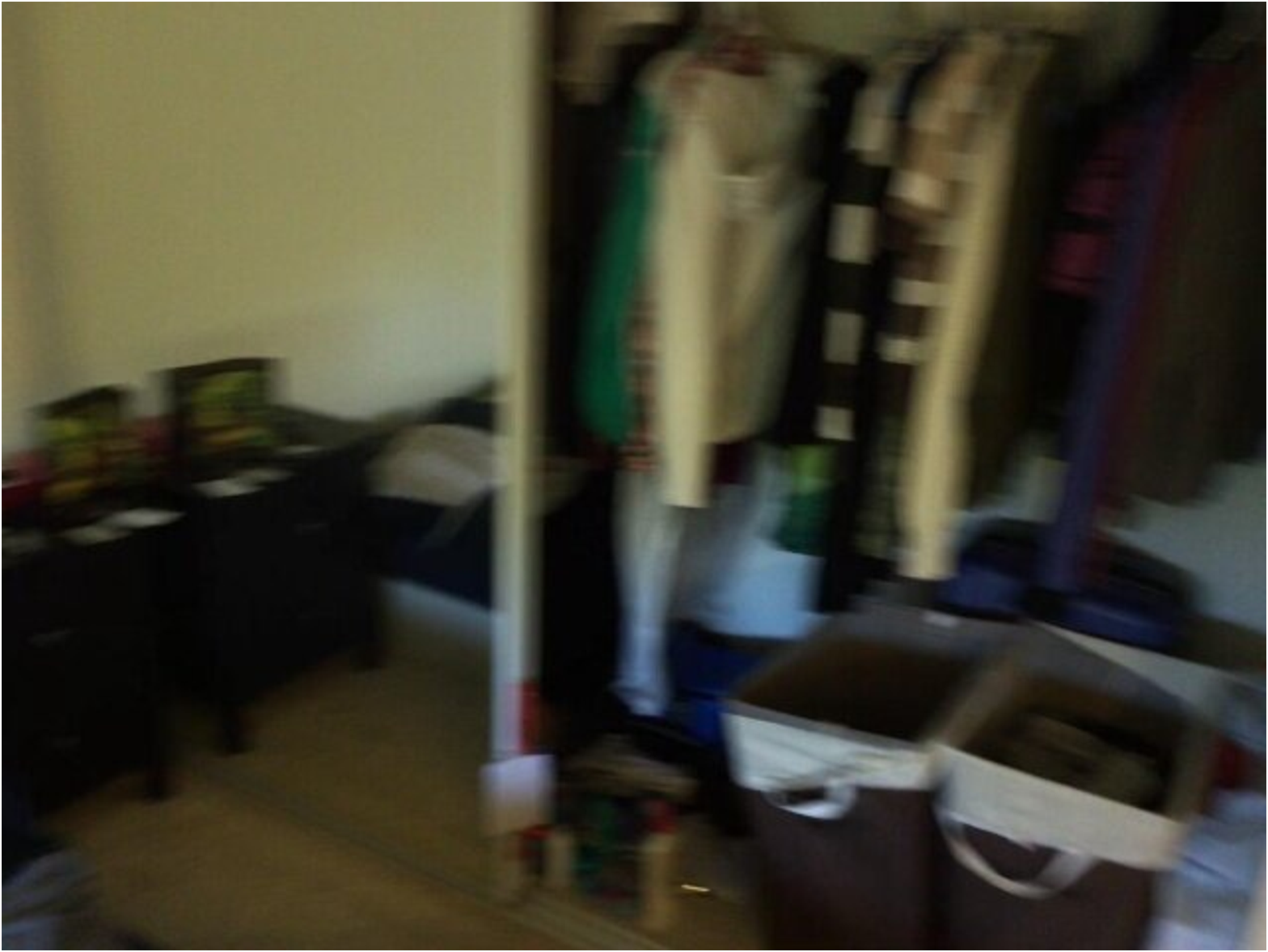}
    \end{minipage}%
    \begin{minipage}[t]{0.121\linewidth}
        \centering
        \includegraphics[width=0.83in,height=0.63in]{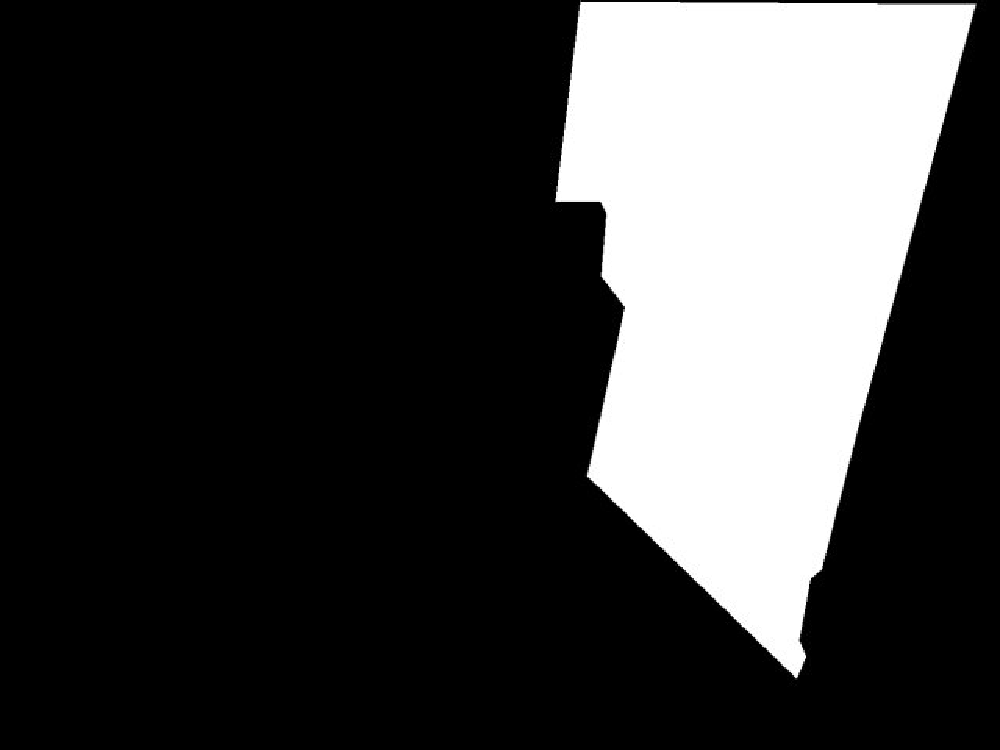}\vspace{2pt}
        \includegraphics[width=0.83in,height=0.63in]{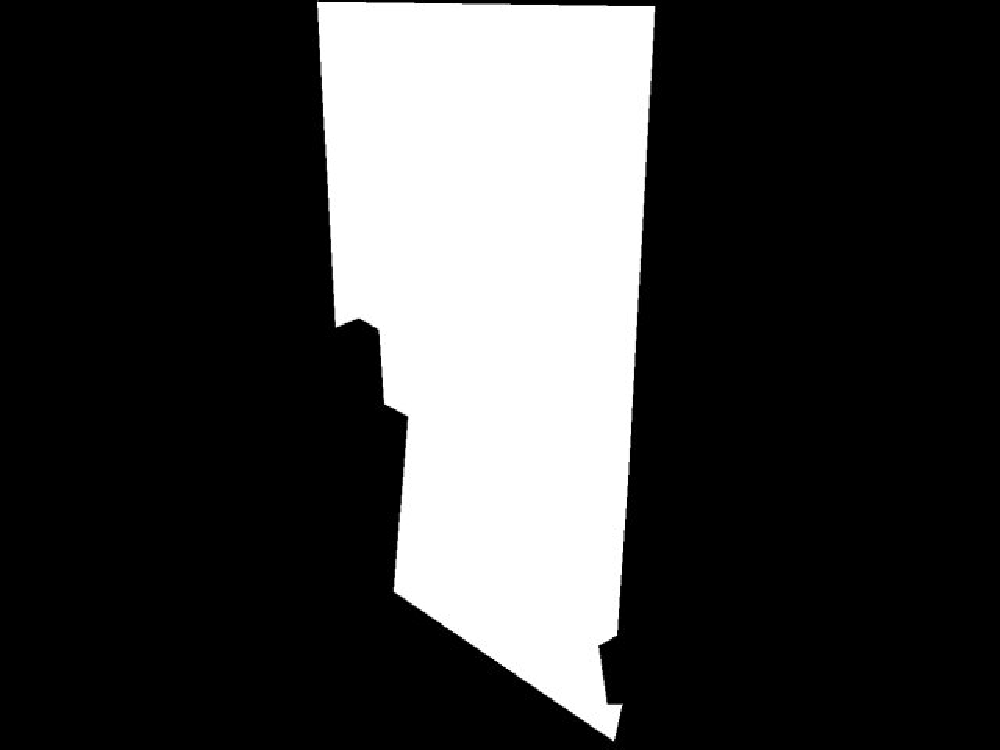}\vspace{2pt}
        \includegraphics[width=0.83in,height=0.63in]{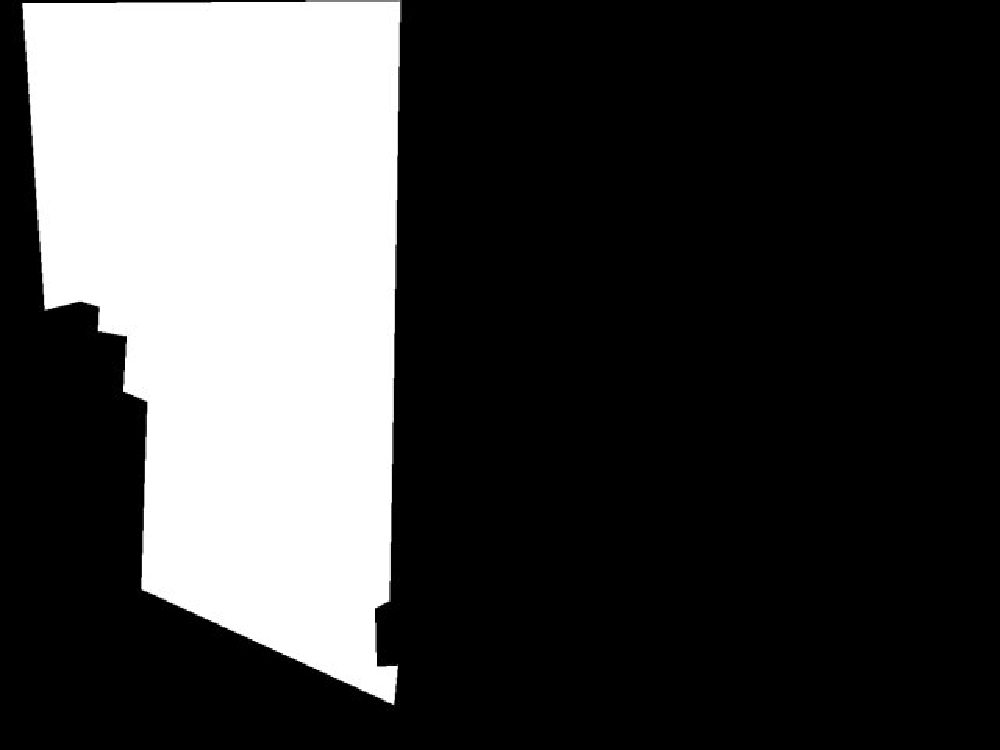}
    \end{minipage}%
    \begin{minipage}[t]{0.121\linewidth}
        \centering
        \includegraphics[width=0.83in,height=0.63in]{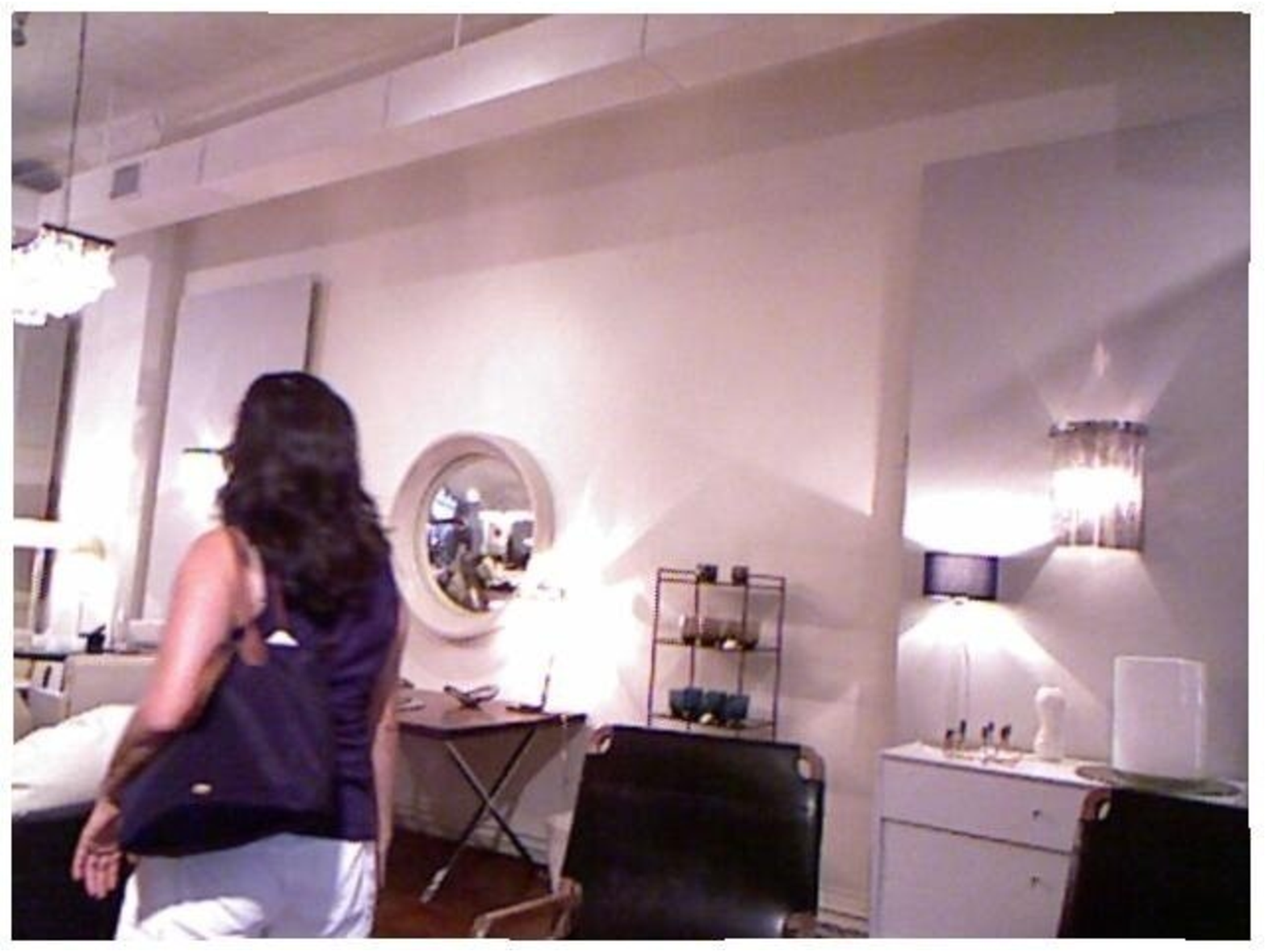}\vspace{2pt}
        \includegraphics[width=0.83in,height=0.63in]{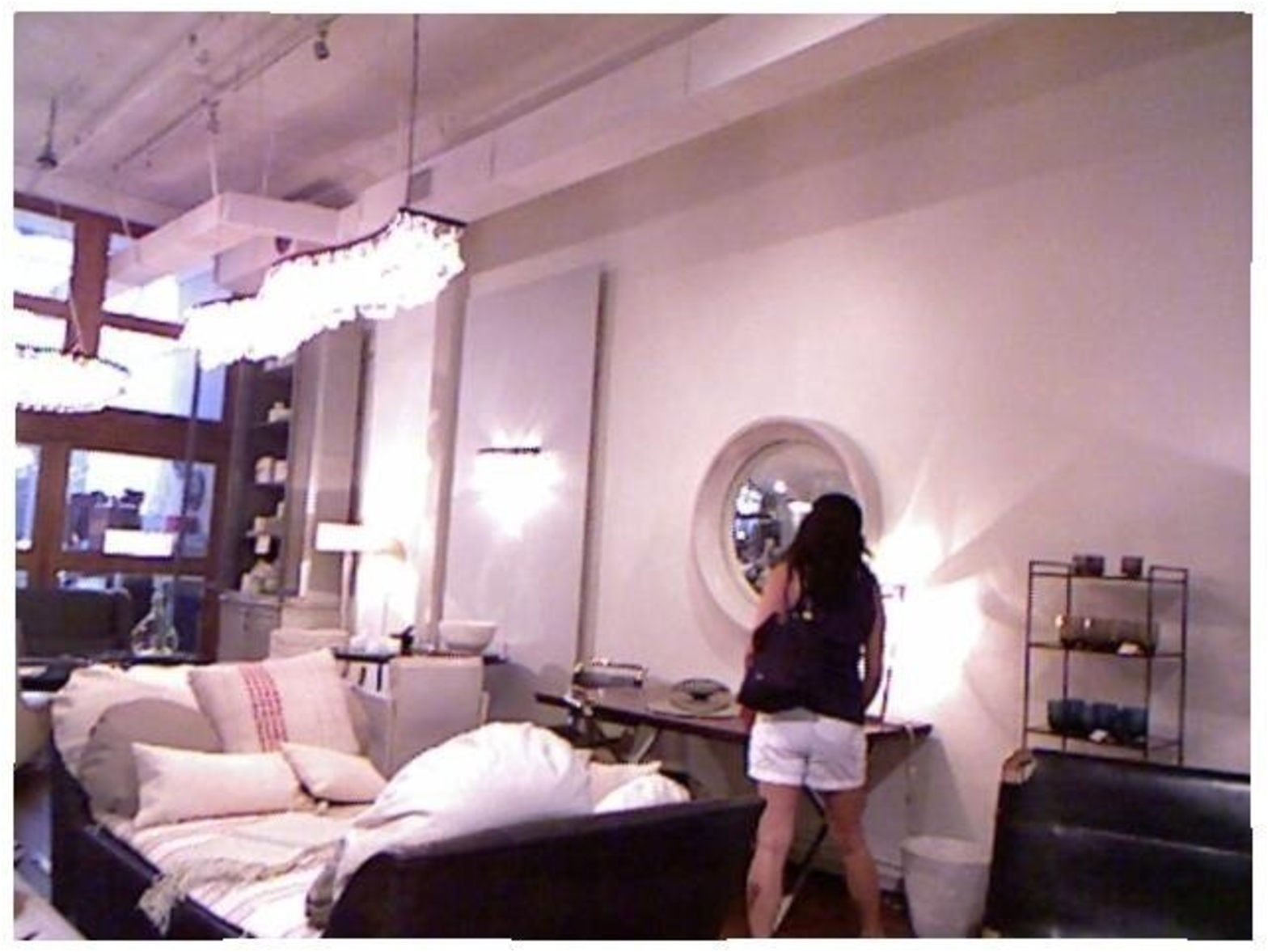}\vspace{2pt}
        \includegraphics[width=0.83in,height=0.63in]{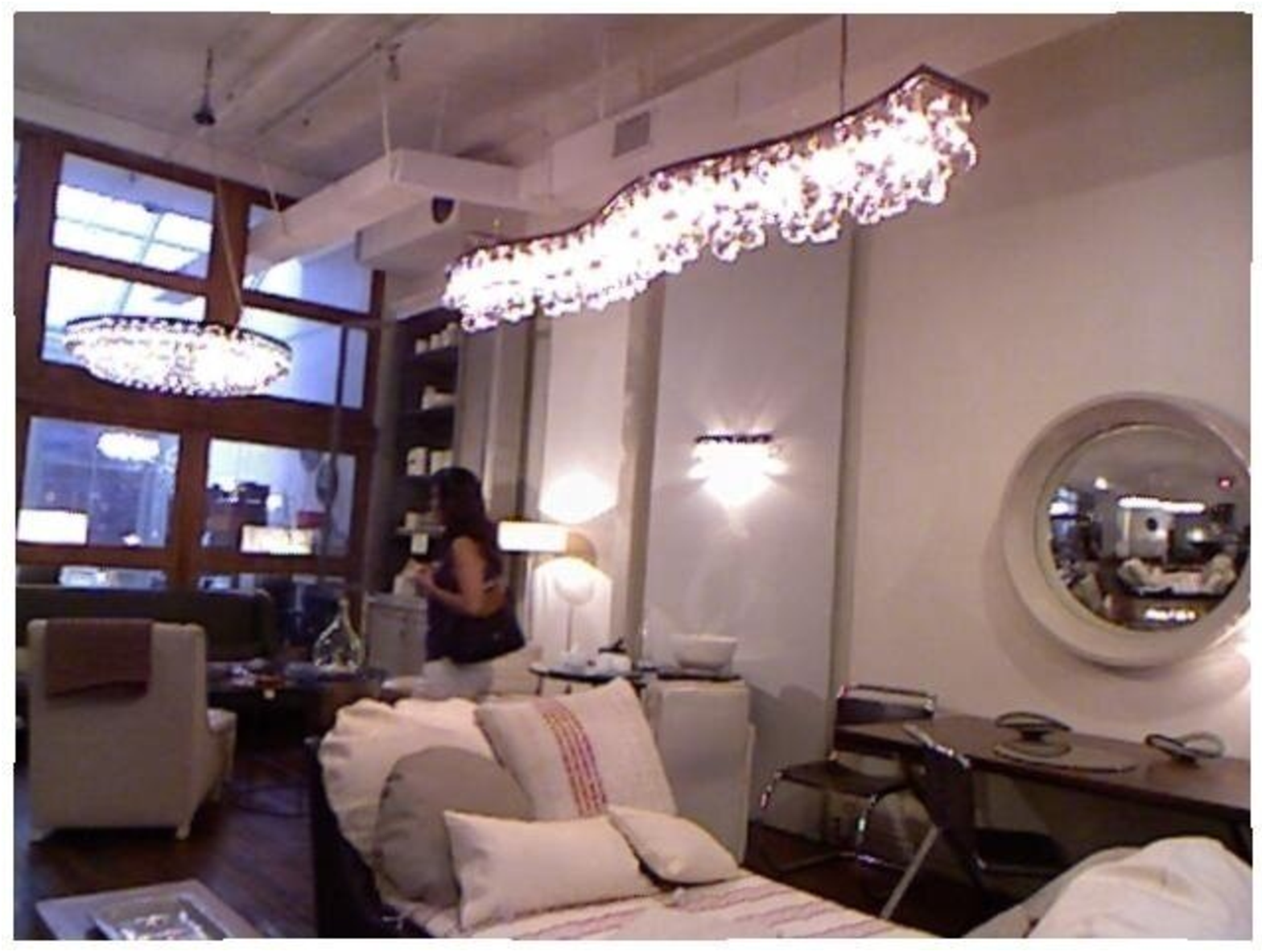}
    \end{minipage}%
    \begin{minipage}[t]{0.121\linewidth}
        \centering
        \includegraphics[width=0.83in,height=0.63in]{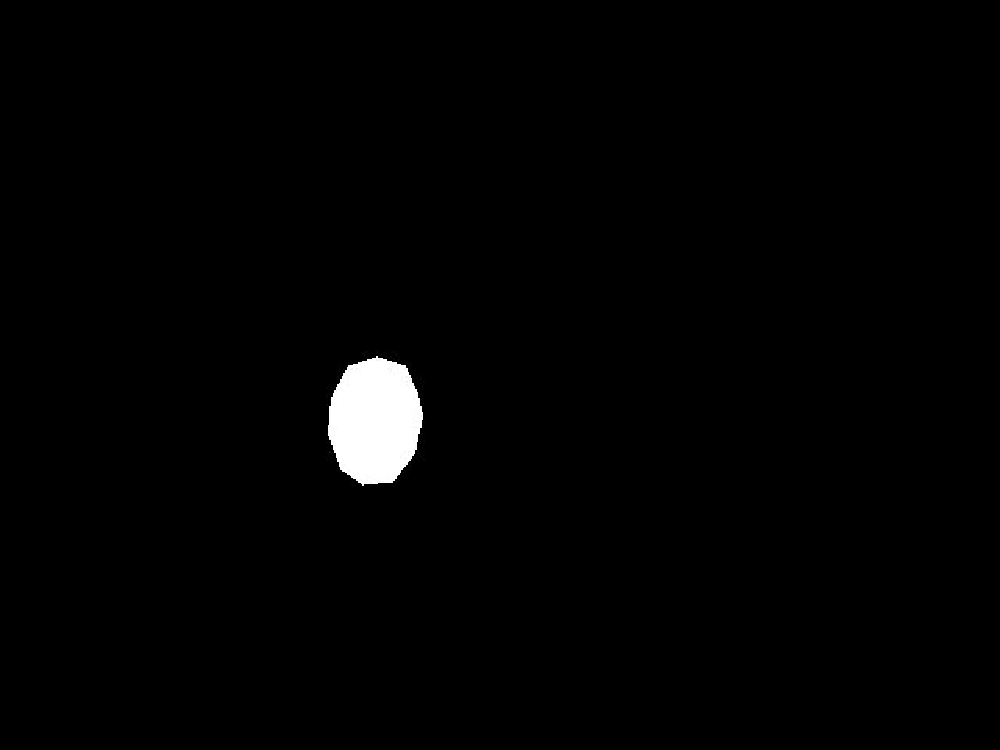}\vspace{2pt}
        \includegraphics[width=0.83in,height=0.63in]{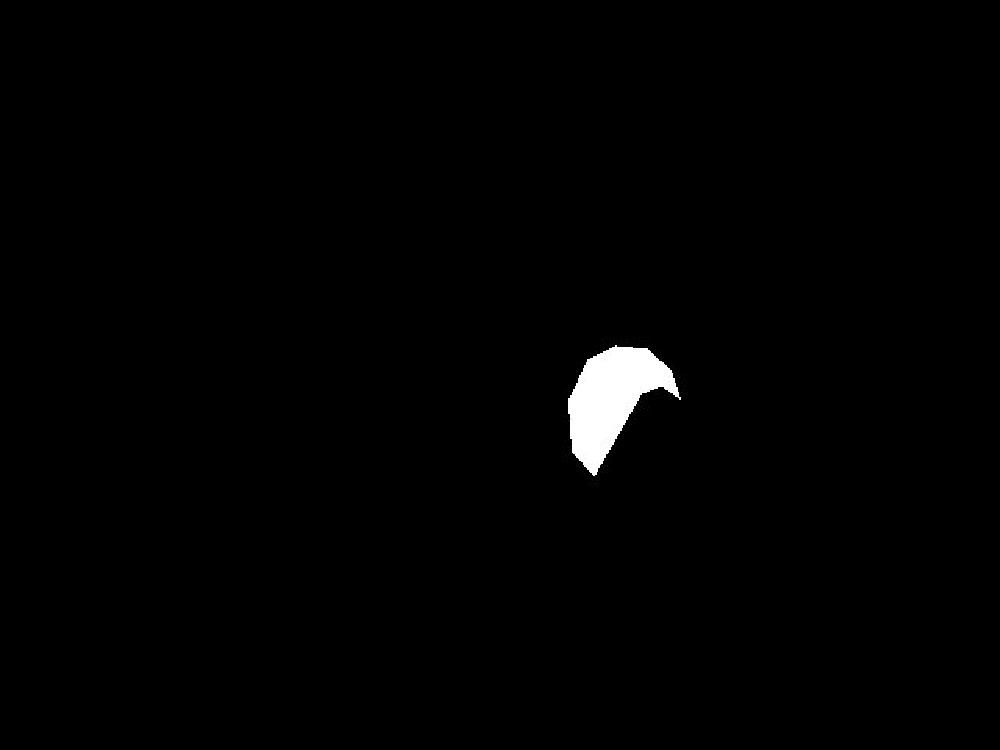}\vspace{2pt}
        \includegraphics[width=0.83in,height=0.63in]{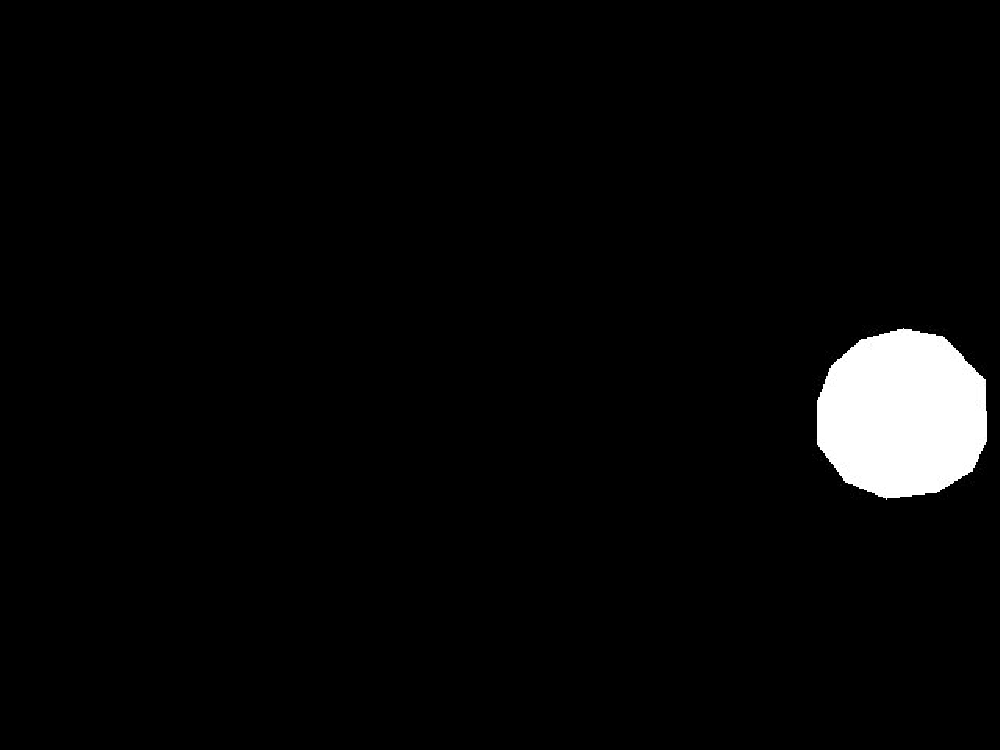}
    \end{minipage}%
  \caption{Videos in our ViMirr dataset show high diversity and low similarity. They cover lots of daily scenes.}
  \label{fig:examples_of_vimirr}
\end{figure*}

In the fusion block, we first use dual gated attention features $D_{t-1}$, $D_{t}$ to refine the original 
features $F_{t-1}$, $F_{t}$, and then fuse the refined features $G_{t-1}$, $G_{t}$ with the 
correspondence features $R_{t-1}$, ${R_t}$ to obtain enhanced dual gated short-term attention features $E_{t-1}$, $E_t$. The fusion block process can be formulated as:
\begin{equation}
    \begin{aligned}
        E_{t-1} = R_{t-1} + Conv_{3\times3}(Cconcat(F_{t-1}, D_{t-1})),
    \end{aligned}
    \label{eq:DG_2}
\end{equation}
\begin{equation}
    \begin{aligned}
        E_{t} = R_{t} + Conv_{3\times3}(Cconcat(F_{t}, D_{t})),
    \end{aligned}
    \label{eq:DG_2}
\end{equation}

\subsection{Short-long Fusion Module}
The SLF module is designed to fuse short-term features with long-term features to further focus on the mirror with a global view. The reason 
to take long-term 
features into account is that we notice the mirror frequently appears throughout the whole video. Here, we utilize the LA module, instead of DGSA, to obtain the long-term relation features 
because we find that the dual gated mechanism may be confused by the mirror appearance changes 
in 
long video clips.  LA module follows the design of the cross-attention module \cite{b5}. The difference is that we are using it to extract the long-term correspondence, not the short-term correspondence.

\begin{figure}[htb]
  \centering
   \includegraphics[width=0.6\linewidth]{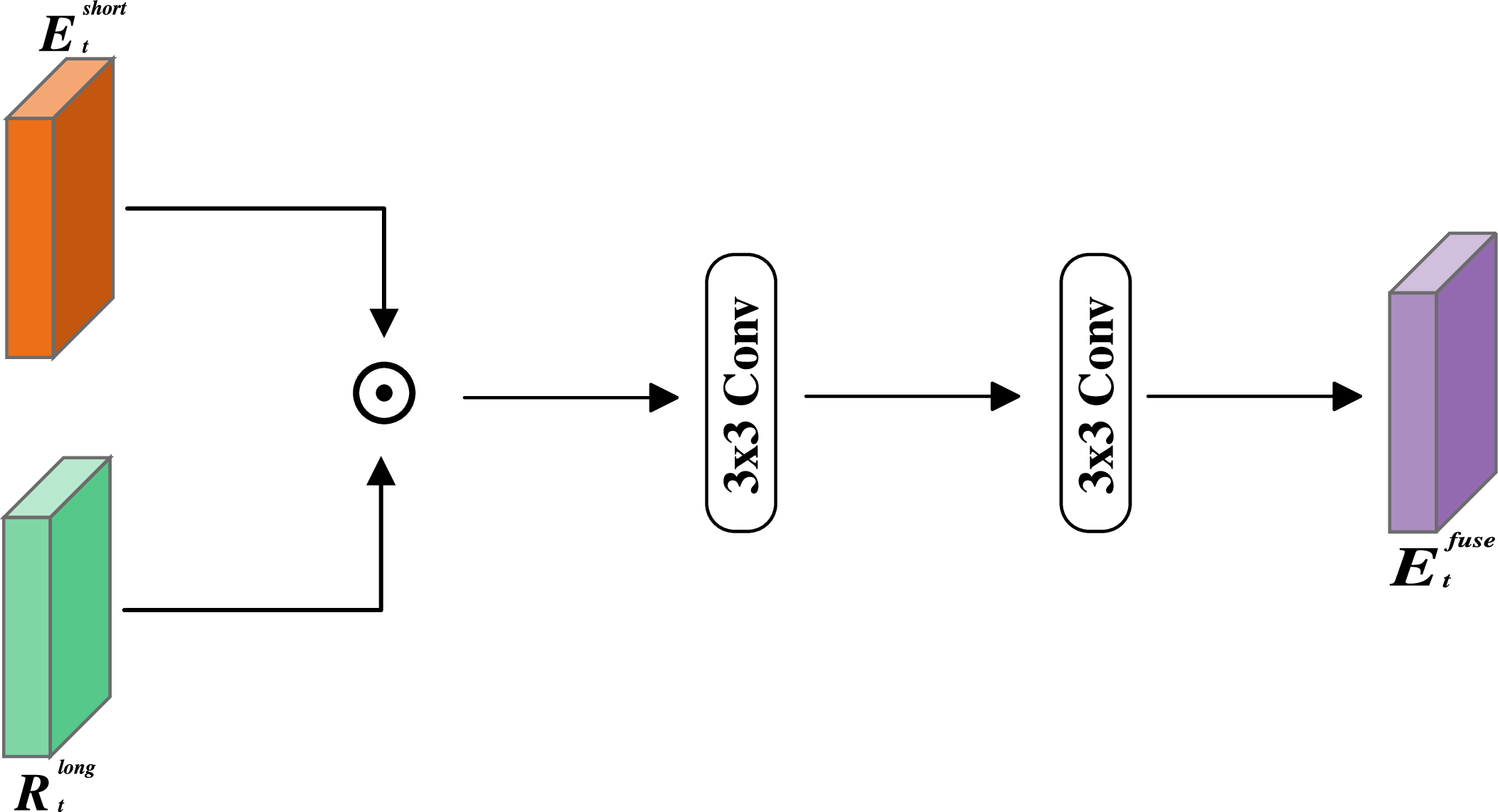}
   \caption{Details of the Short-long Fusion (SLF) module.} 
   \label{fig:SIF}
\end{figure}

Fig.~\ref{fig:SIF} shows the architecture of the SLF module. We weight the enhanced short-term attention features $E_t^{short}$ with the long-term attention features $R_t^{long}$. In this way, the correspondence features $E_t^{fuse}$ are extracted, which encodes both the appearance of the mirror from the short-term features and the position of the mirror from the long-term features. 

\subsection{Loss function}
Following~\cite{b17}, we adopt the binary cross-entropy (BCE) and the Lovasz-hinge loss 
to supervise the training of the mirror maps:
\begin{equation}
    \begin{aligned}
      \mathcal{L} & = \sum_{i}^{i\in\left\{t-1,t,n\right\}}\mathcal{L}_{hinge}(P_i,G_i) + \mathcal{L}_{bce}(P_i, G_i),
    \end{aligned}
    \label{eq:loss}
\end{equation}
where $\mathcal{L}_{hinge}$, $\mathcal{L}_{bce}$ are the lovasz-hinge loss and the binary cross-entropy (BCE) loss. $P_i$ and $G_i$ are the final predicted map and the ground truth of frames. 


\begin{figure*}[htb]
  \centering
    \begin{minipage}[b]{0.121\linewidth}
        \centering
        \includegraphics[width=0.83in,height=0.55in]{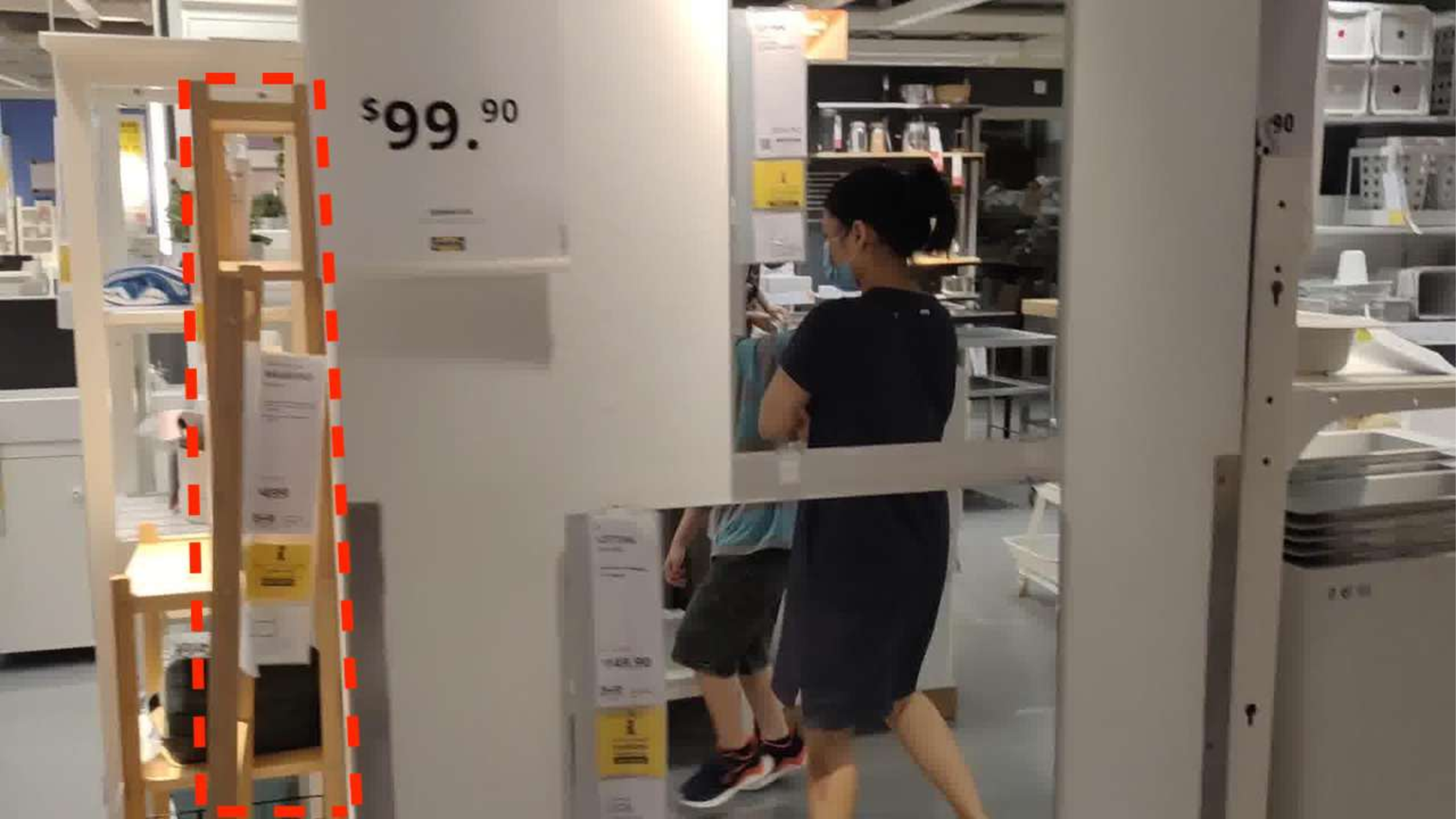}\vspace{2pt}
        \includegraphics[width=0.83in,height=0.55in]{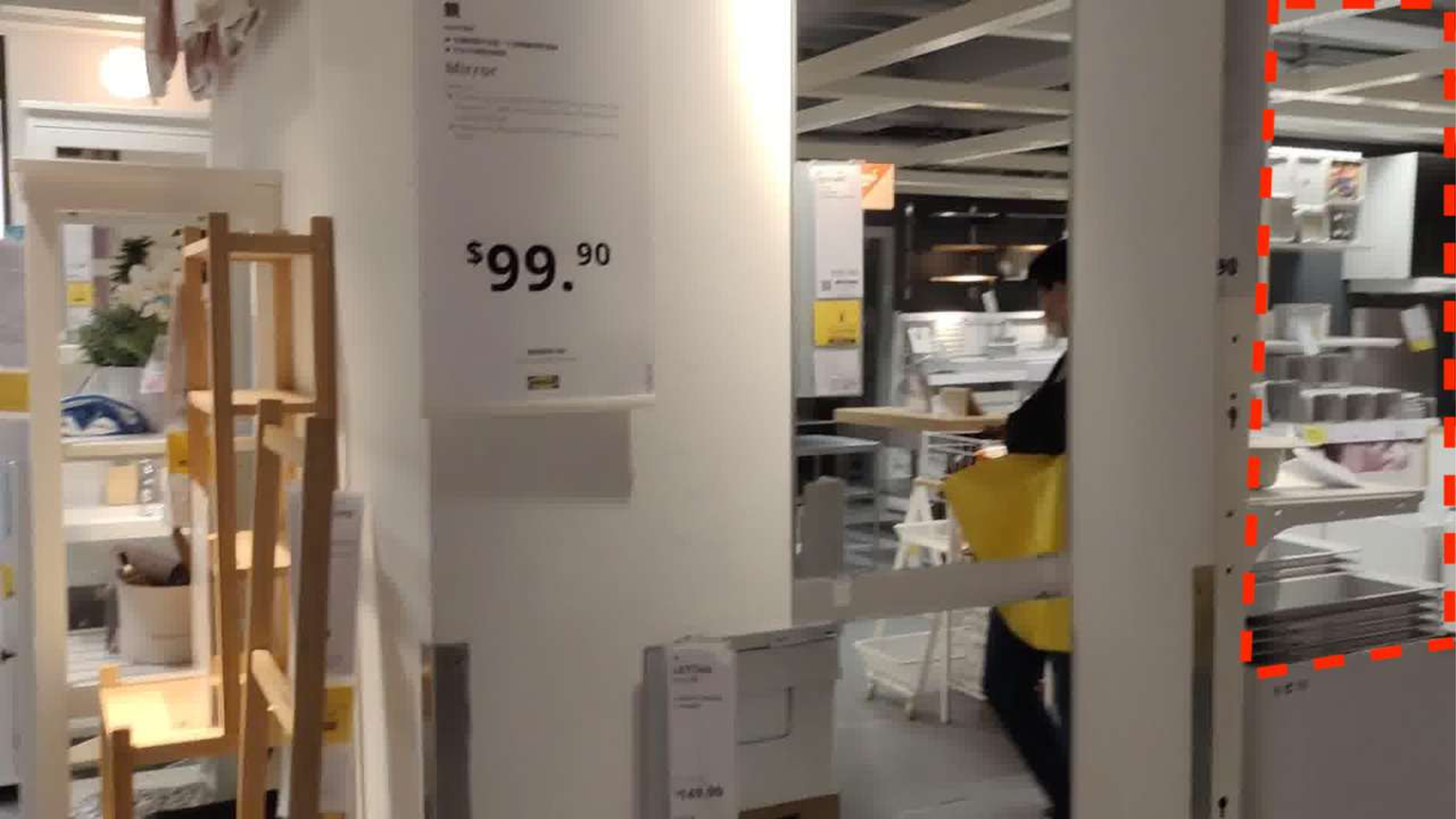}\vspace{2pt}
        \includegraphics[width=0.83in,height=0.55in]{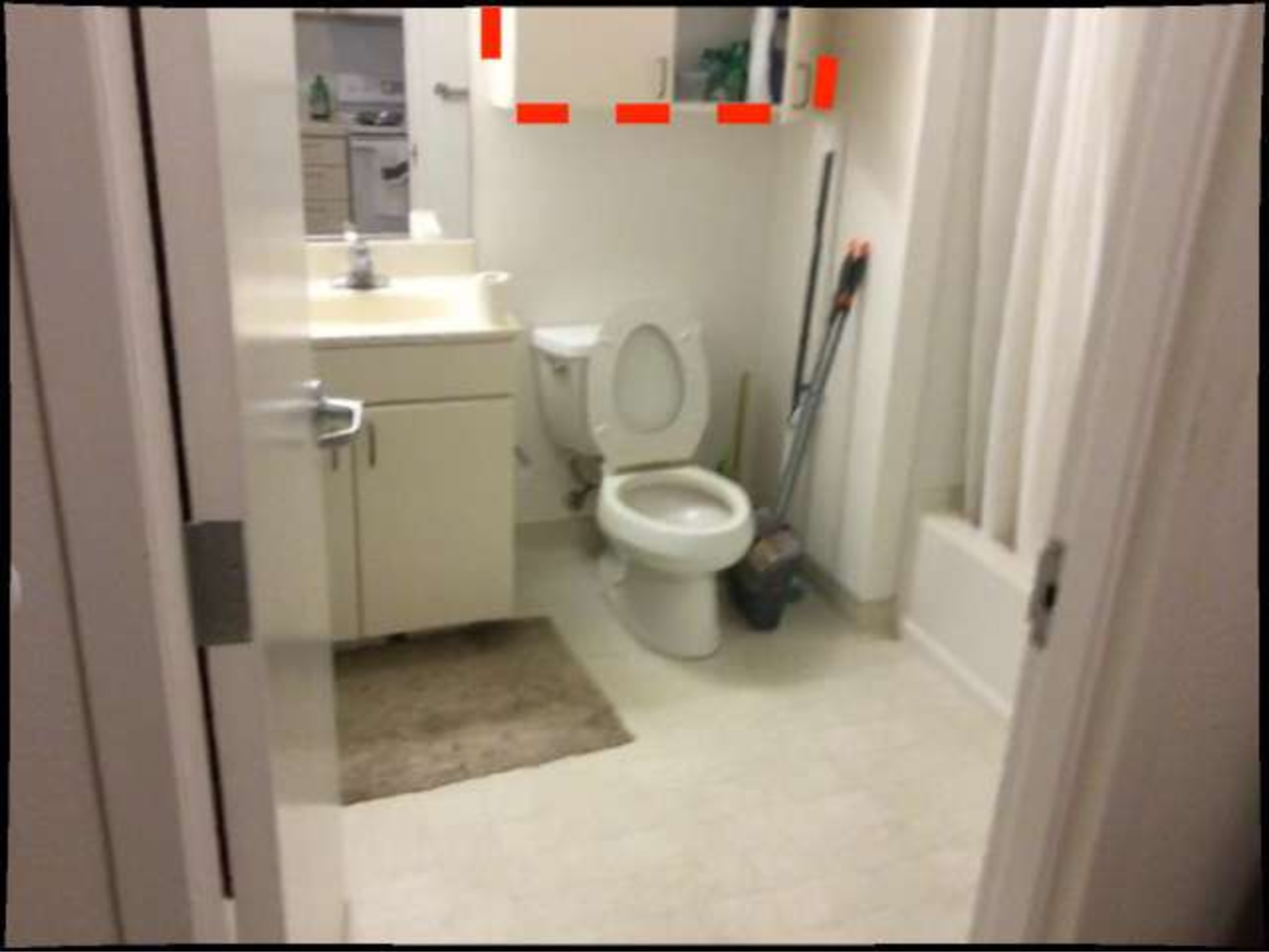}\vspace{2pt}
        \includegraphics[width=0.83in,height=0.55in]{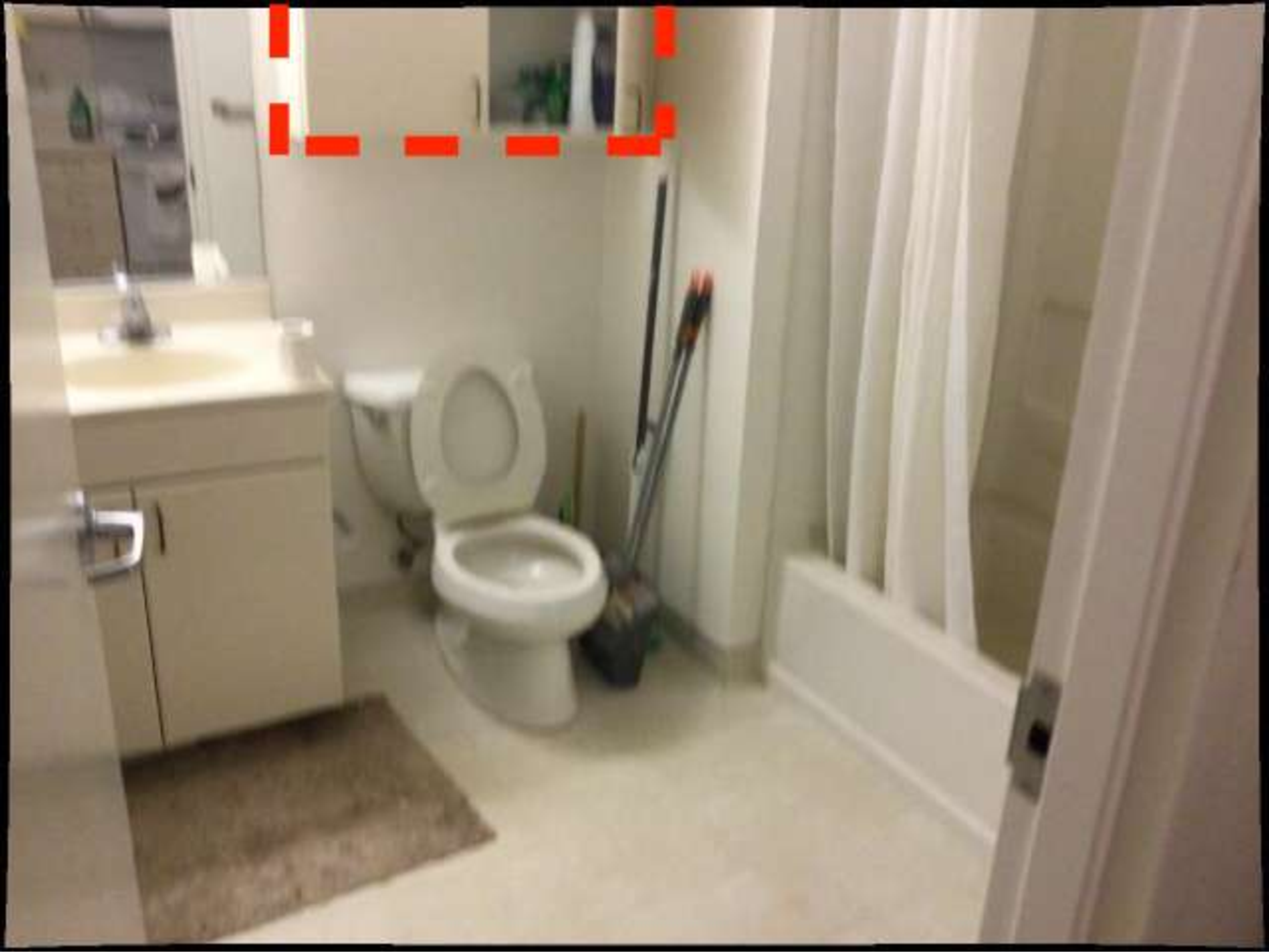}
        Image
    \end{minipage}%
    \begin{minipage}[b]{0.121\linewidth}
        \centering
        \includegraphics[width=0.83in,height=0.55in]{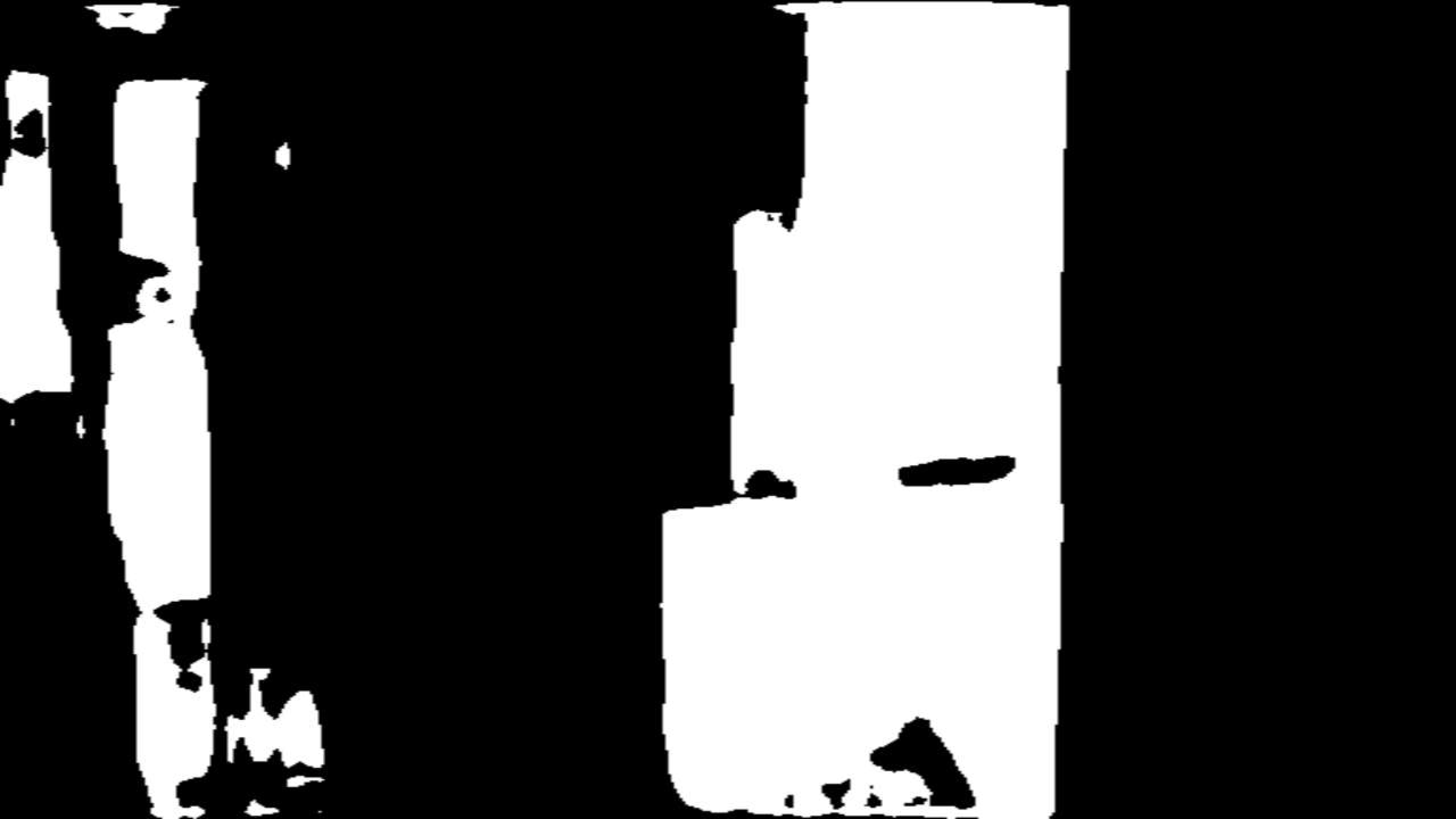}\vspace{2pt}
        \includegraphics[width=0.83in,height=0.55in]{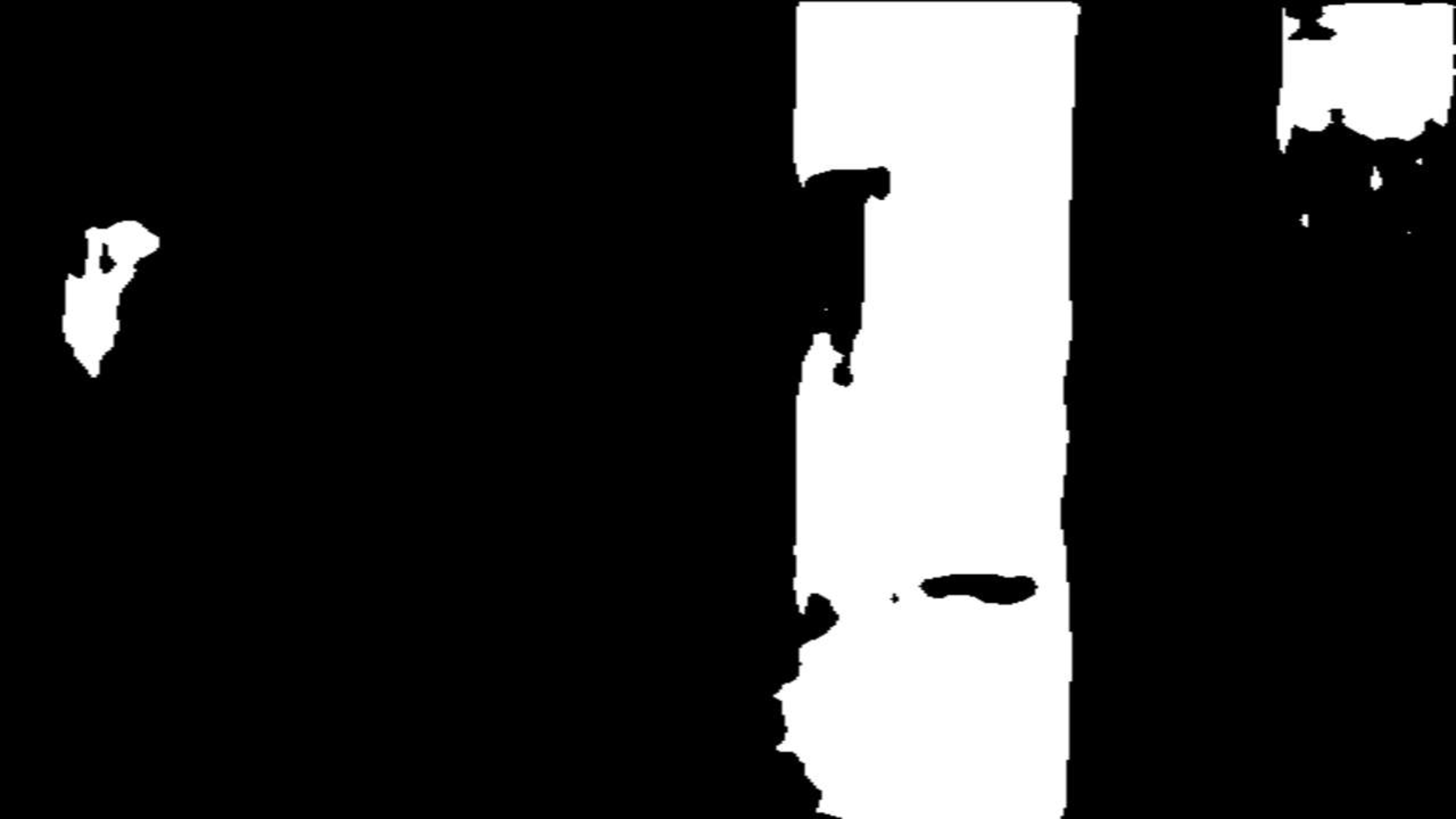}\vspace{2pt}
        \includegraphics[width=0.83in,height=0.55in]{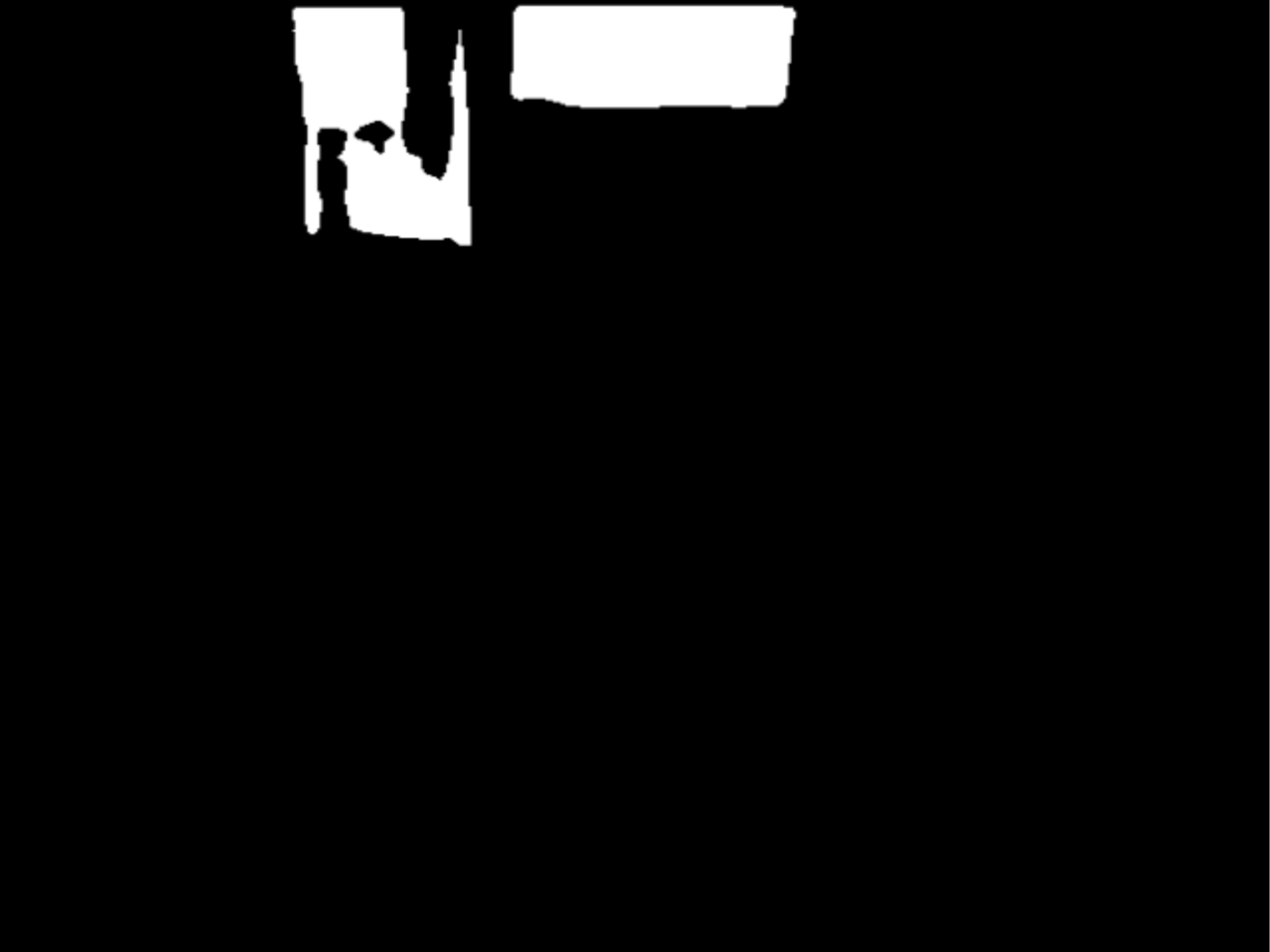}\vspace{2pt}
        \includegraphics[width=0.83in,height=0.55in]{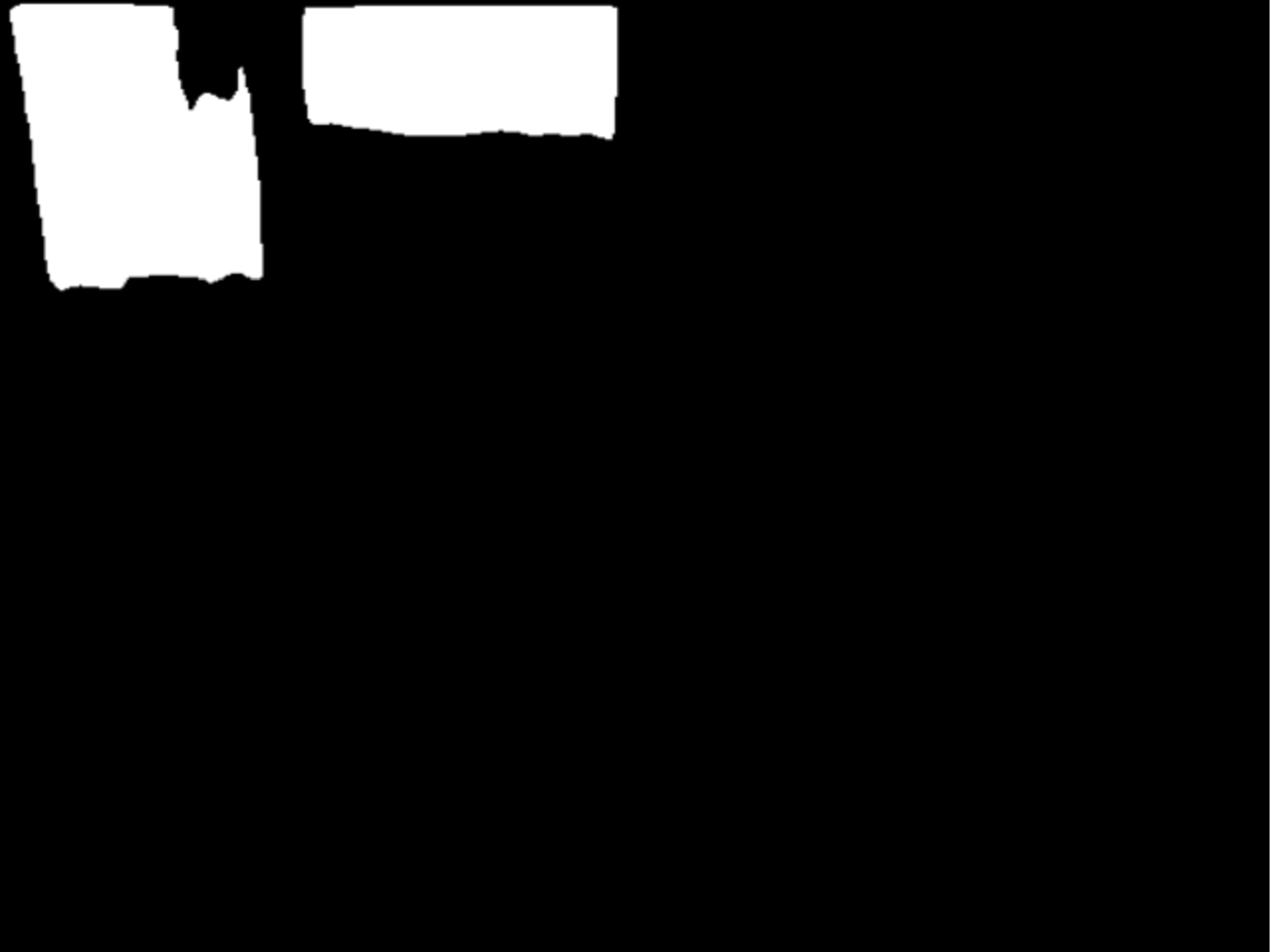}
        GDNet
    \end{minipage}%
    \begin{minipage}[b]{0.121\linewidth}
        \centering
        \includegraphics[width=0.83in,height=0.55in]{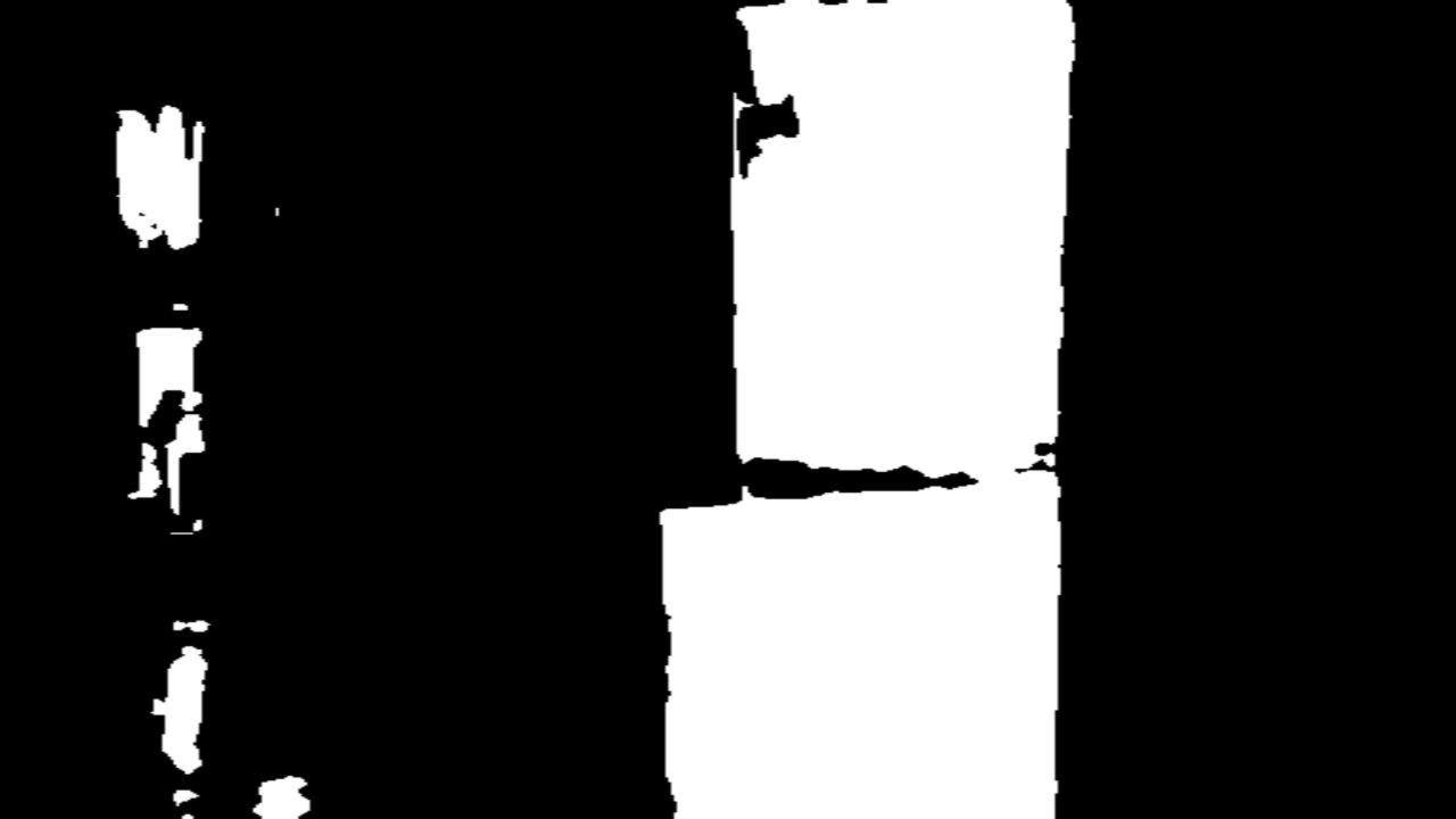}\vspace{2pt}
        \includegraphics[width=0.83in,height=0.55in]{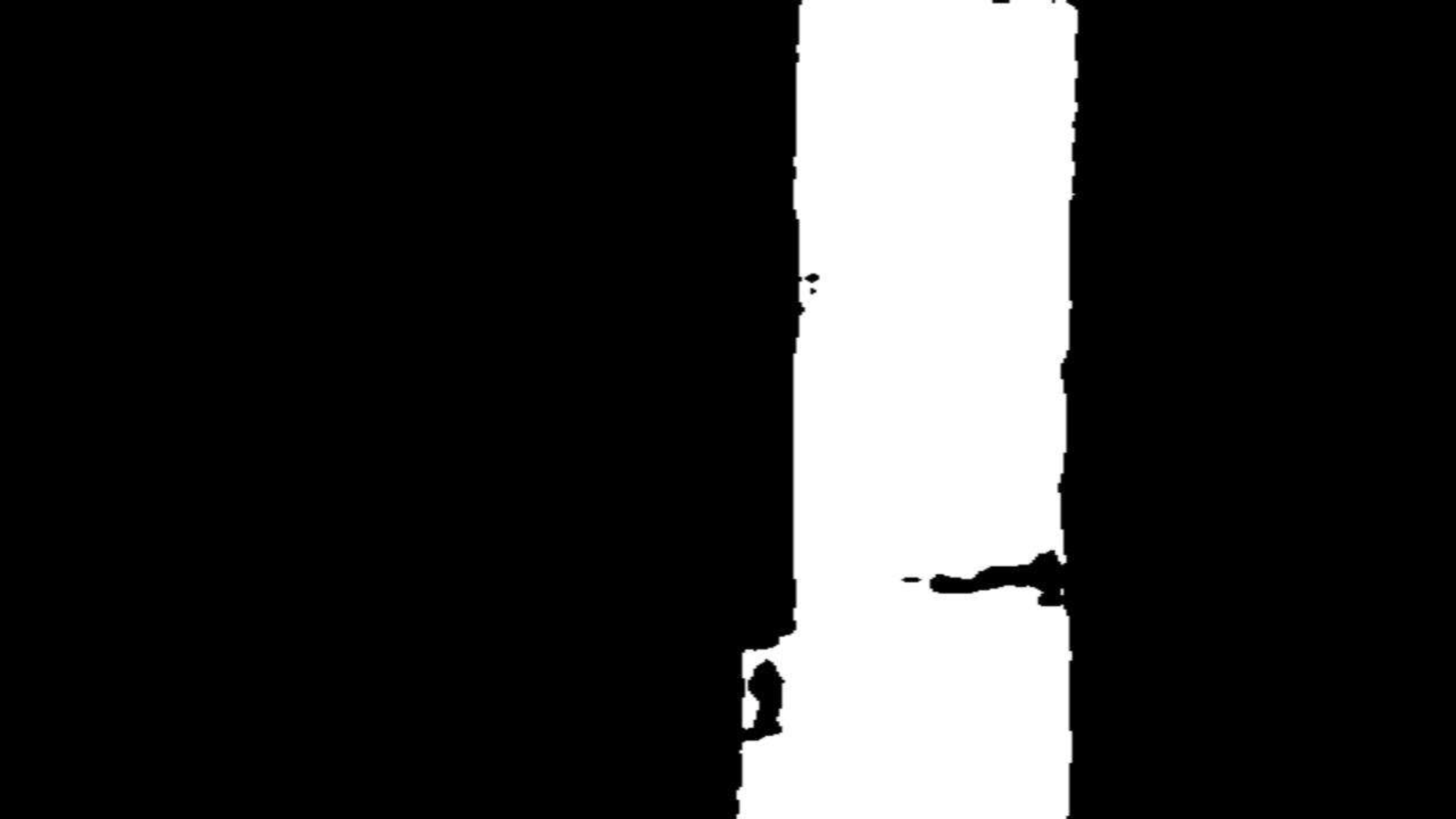}\vspace{2pt}
        \includegraphics[width=0.83in,height=0.55in]{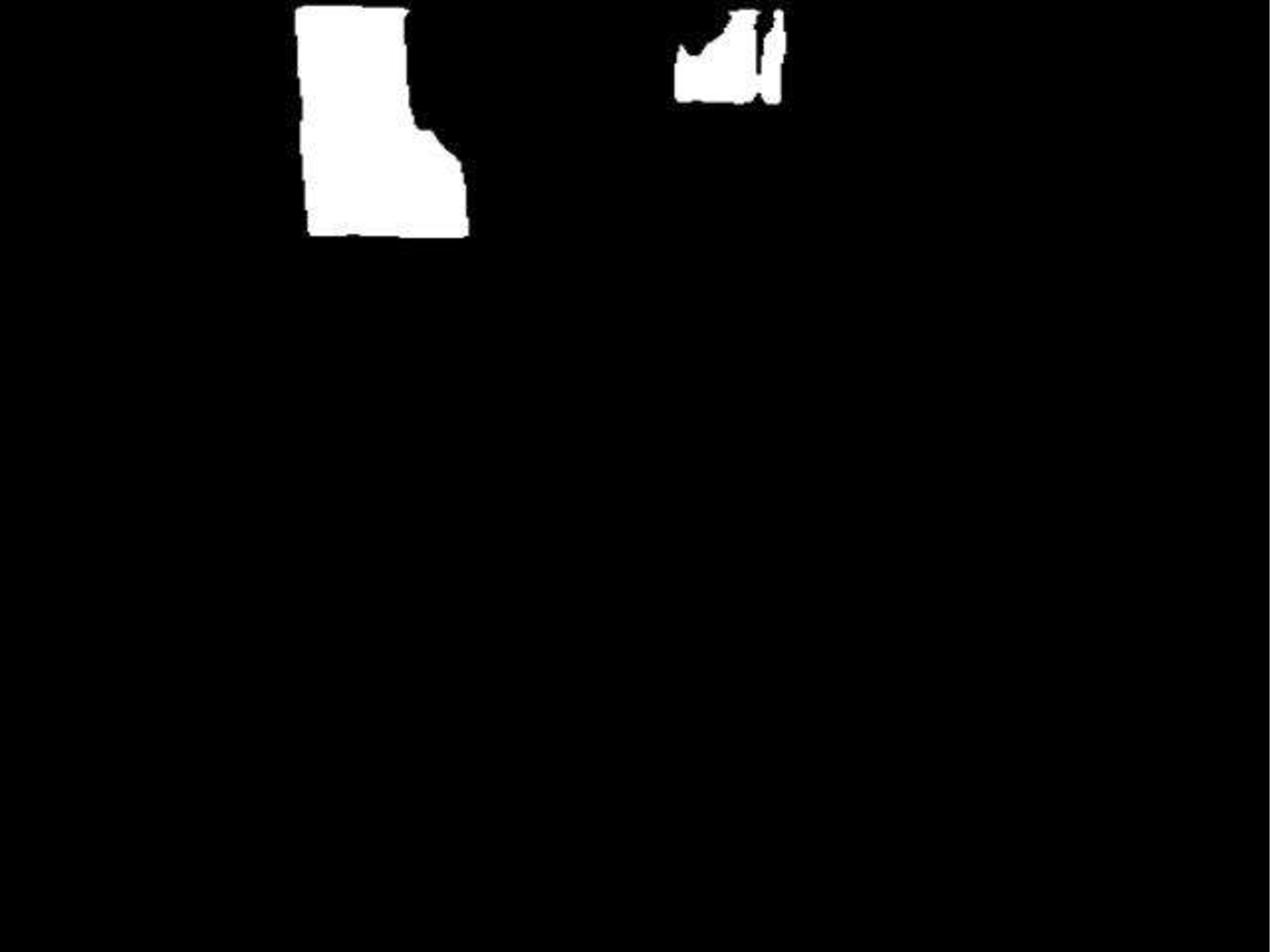}\vspace{2pt}
        \includegraphics[width=0.83in,height=0.55in]{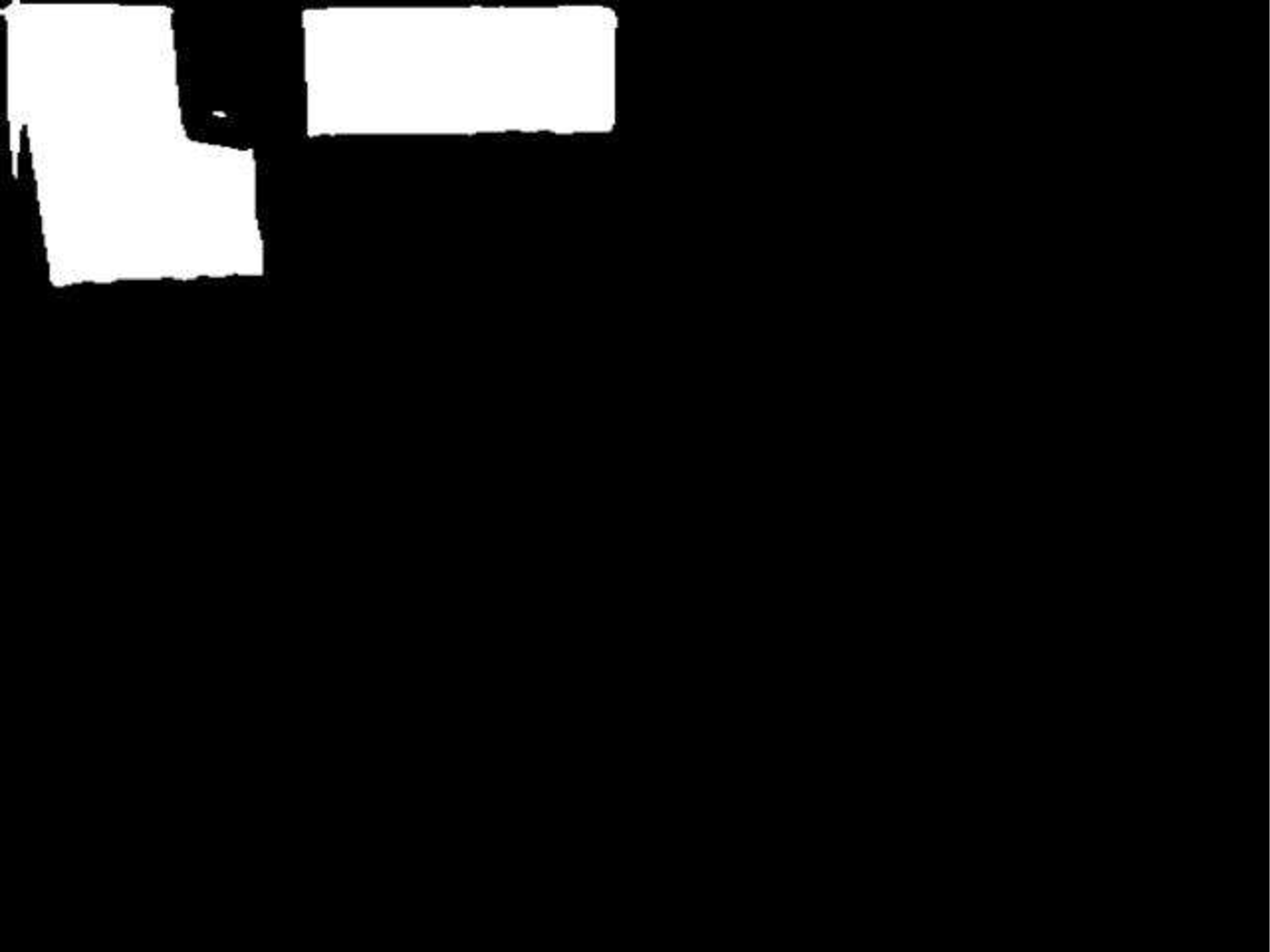}
        SCOTCH
    \end{minipage}%
    \begin{minipage}[b]{0.121\linewidth}
        \centering
        \includegraphics[width=0.83in,height=0.55in]{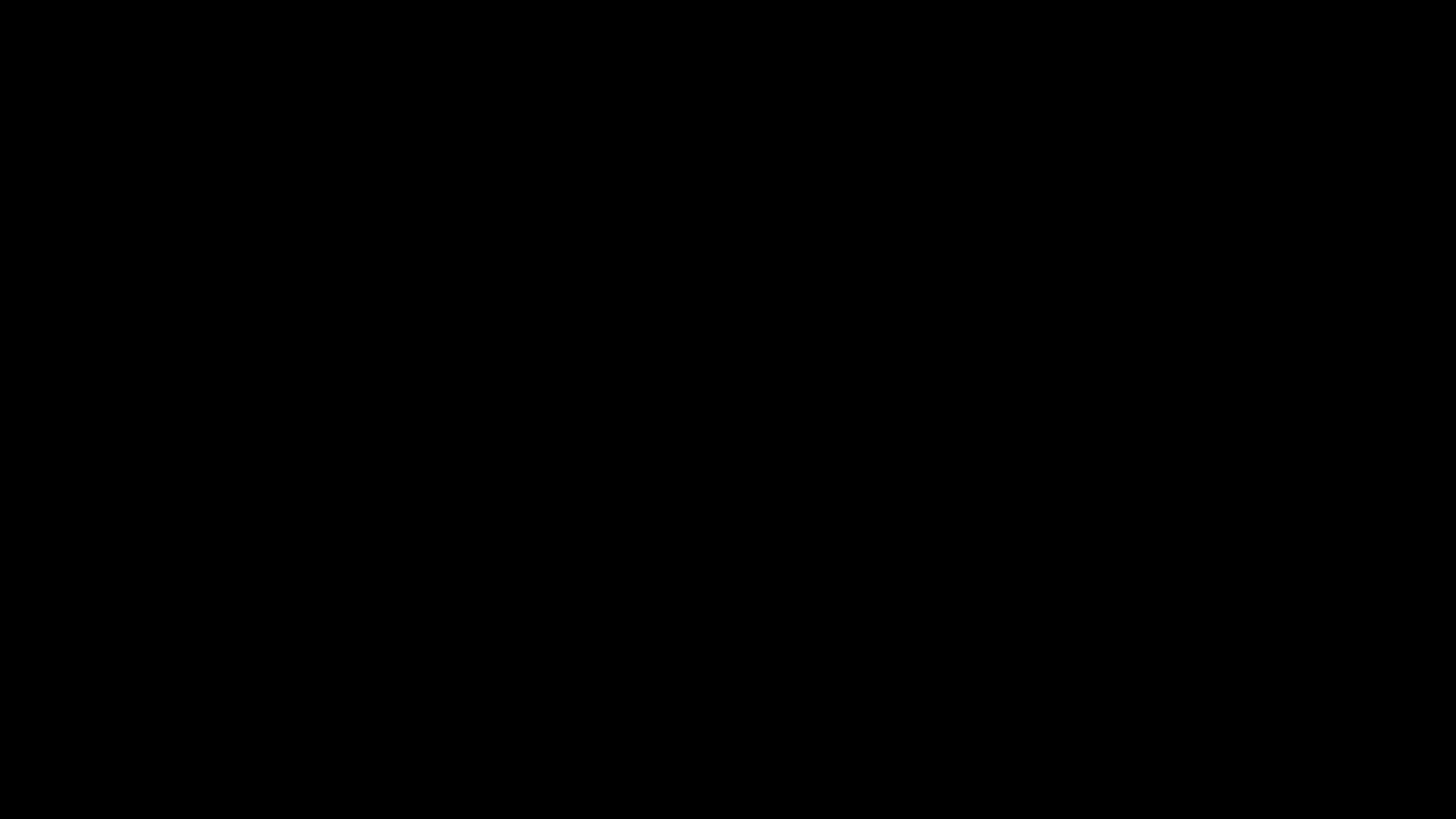}\vspace{2pt}
        \includegraphics[width=0.83in,height=0.55in]{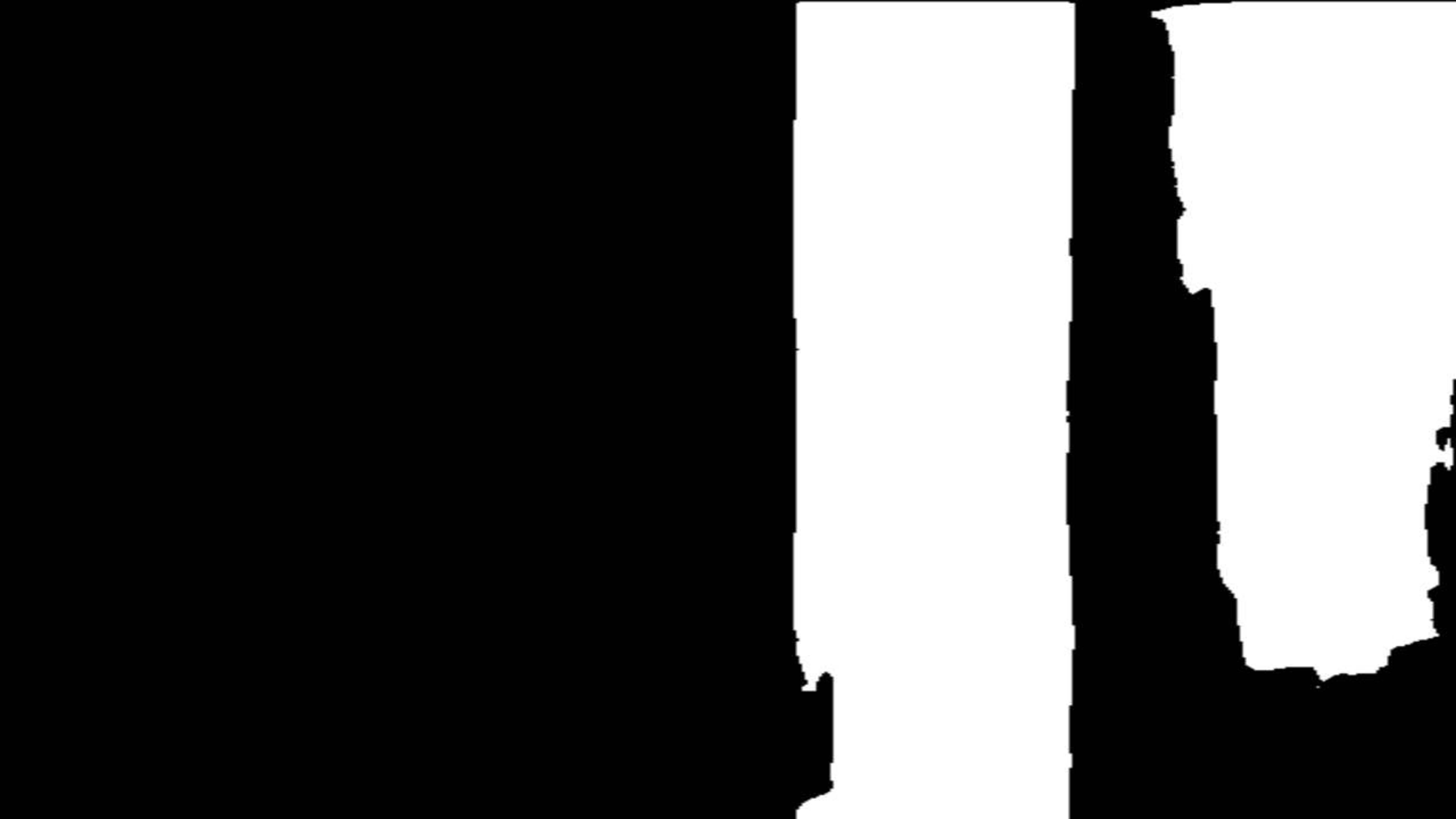}\vspace{2pt}
        \includegraphics[width=0.83in,height=0.55in]{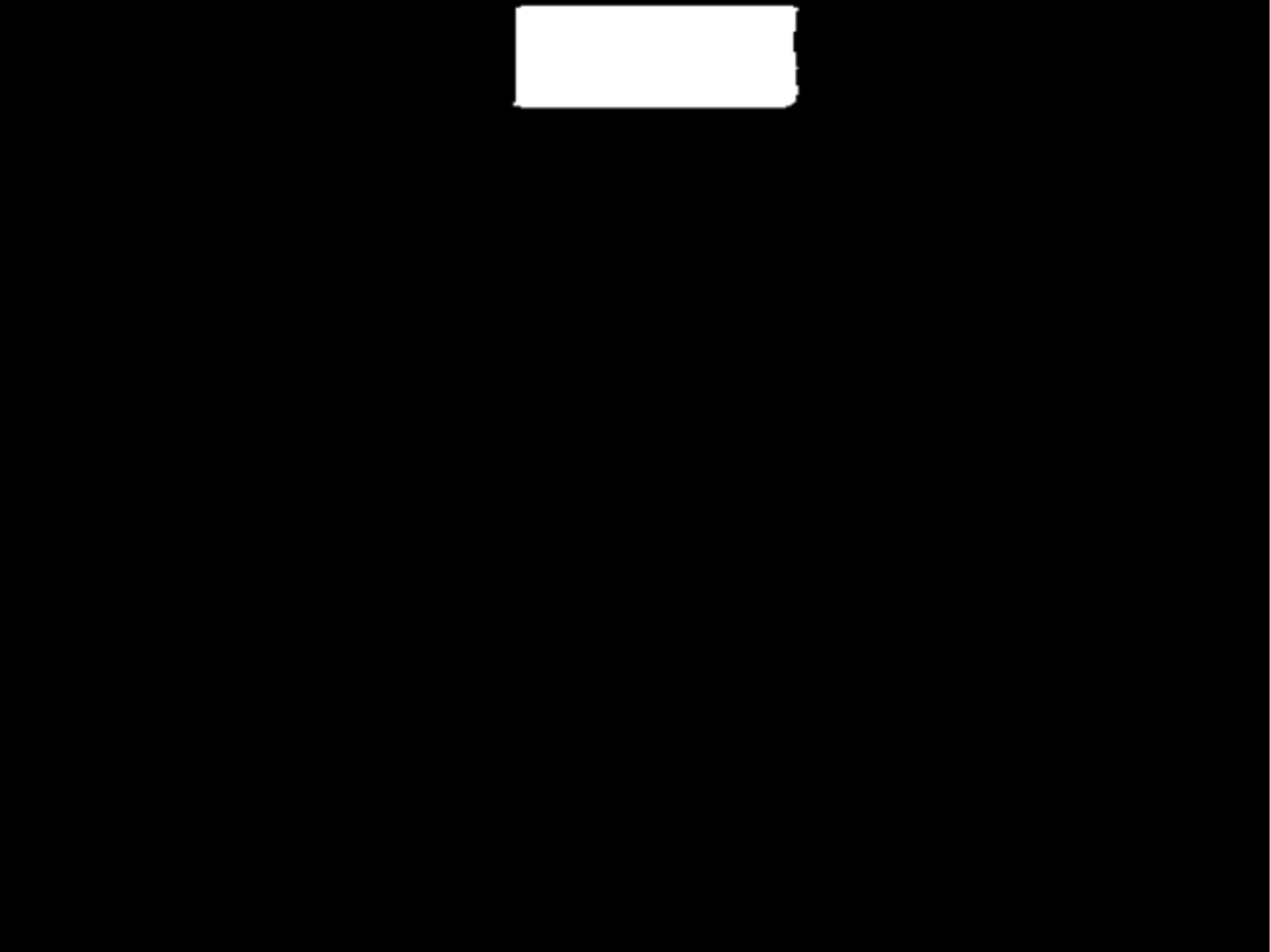}\vspace{2pt}
        \includegraphics[width=0.83in,height=0.55in]{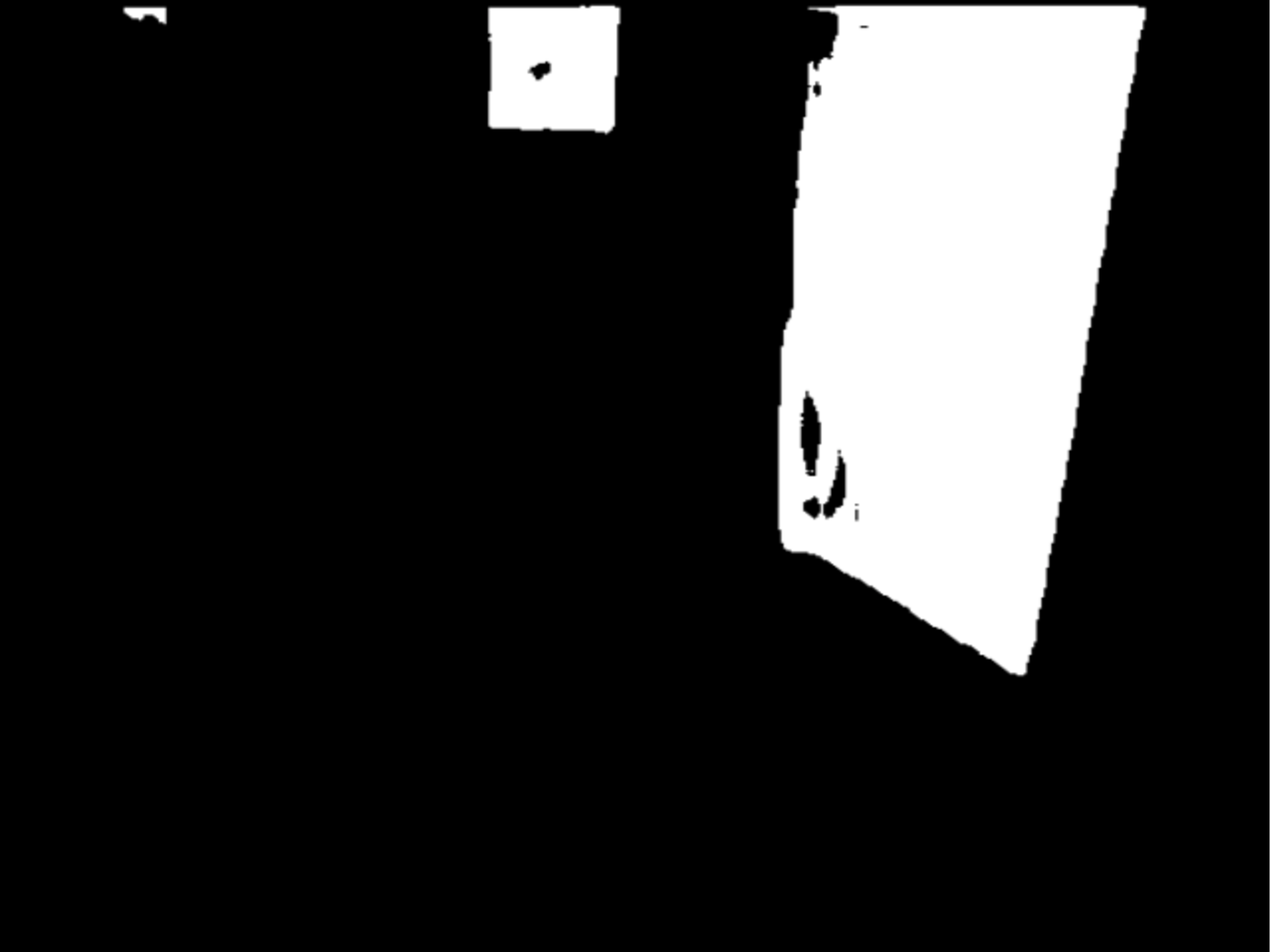}
        PMD
    \end{minipage}%
    \begin{minipage}[b]{0.121\linewidth}
        \centering
        \includegraphics[width=0.83in,height=0.55in]{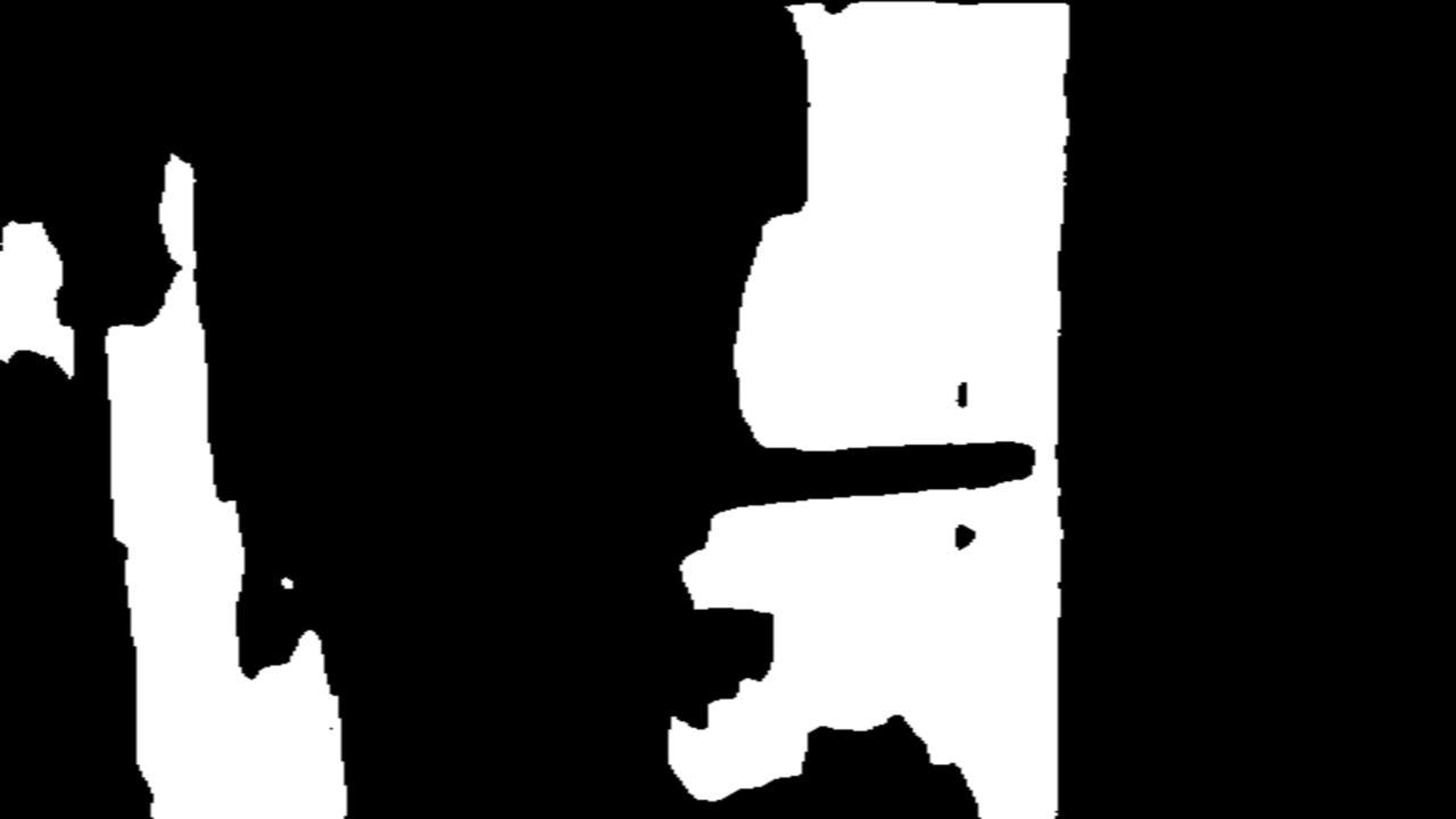}\vspace{2pt}
        \includegraphics[width=0.83in,height=0.55in]{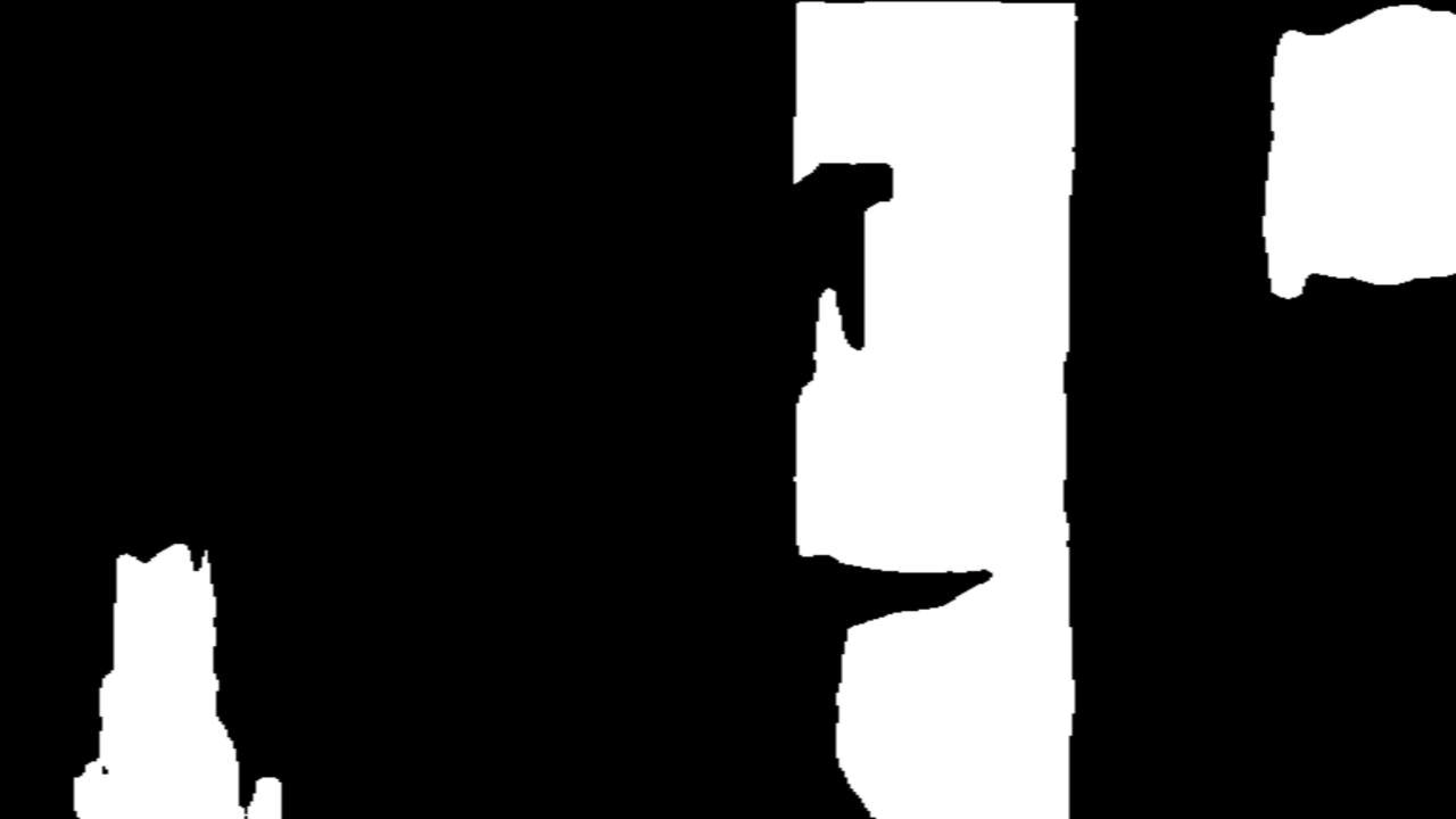}\vspace{2pt}
        \includegraphics[width=0.83in,height=0.55in]{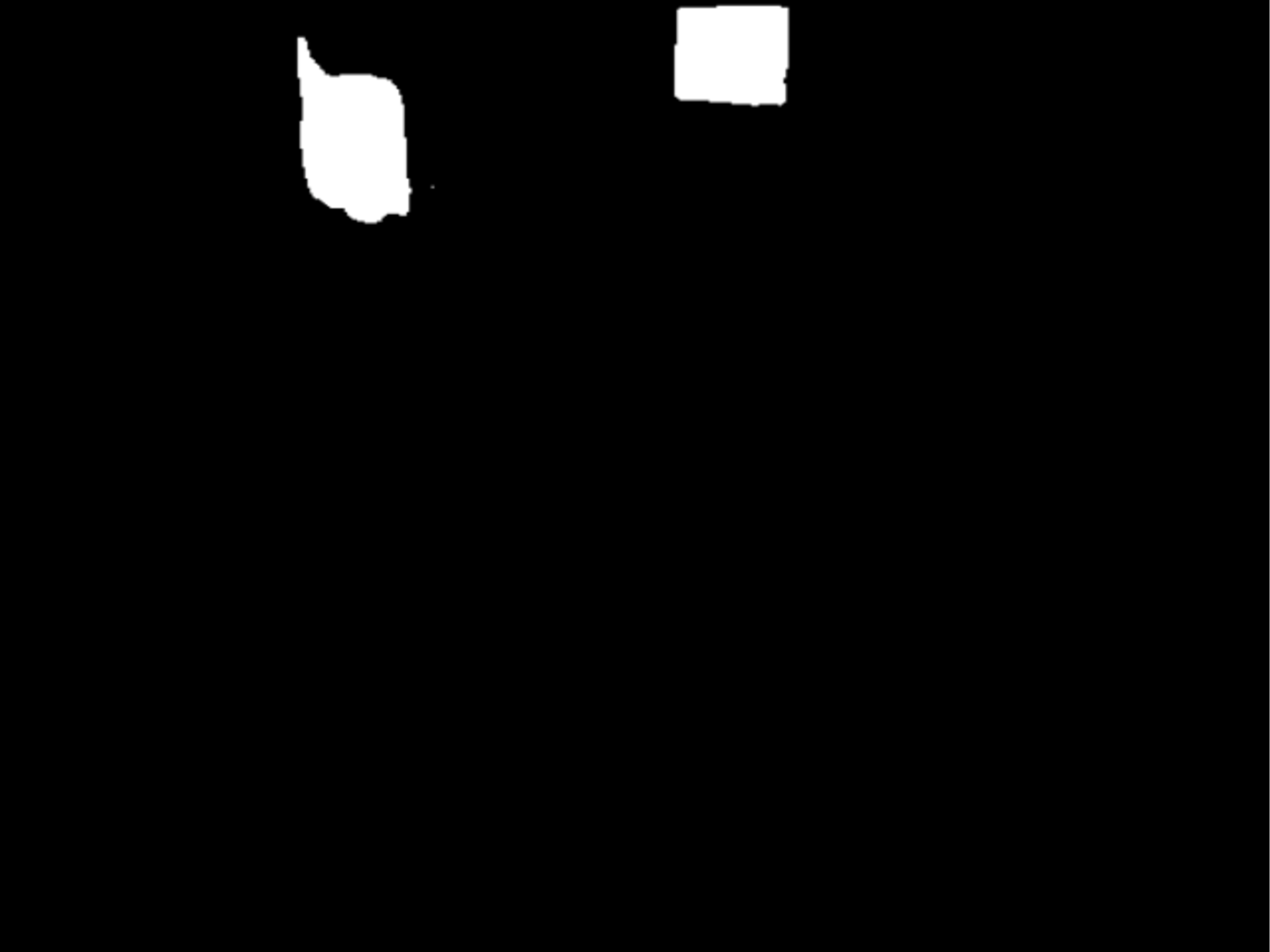}\vspace{2pt}
        \includegraphics[width=0.83in,height=0.55in]{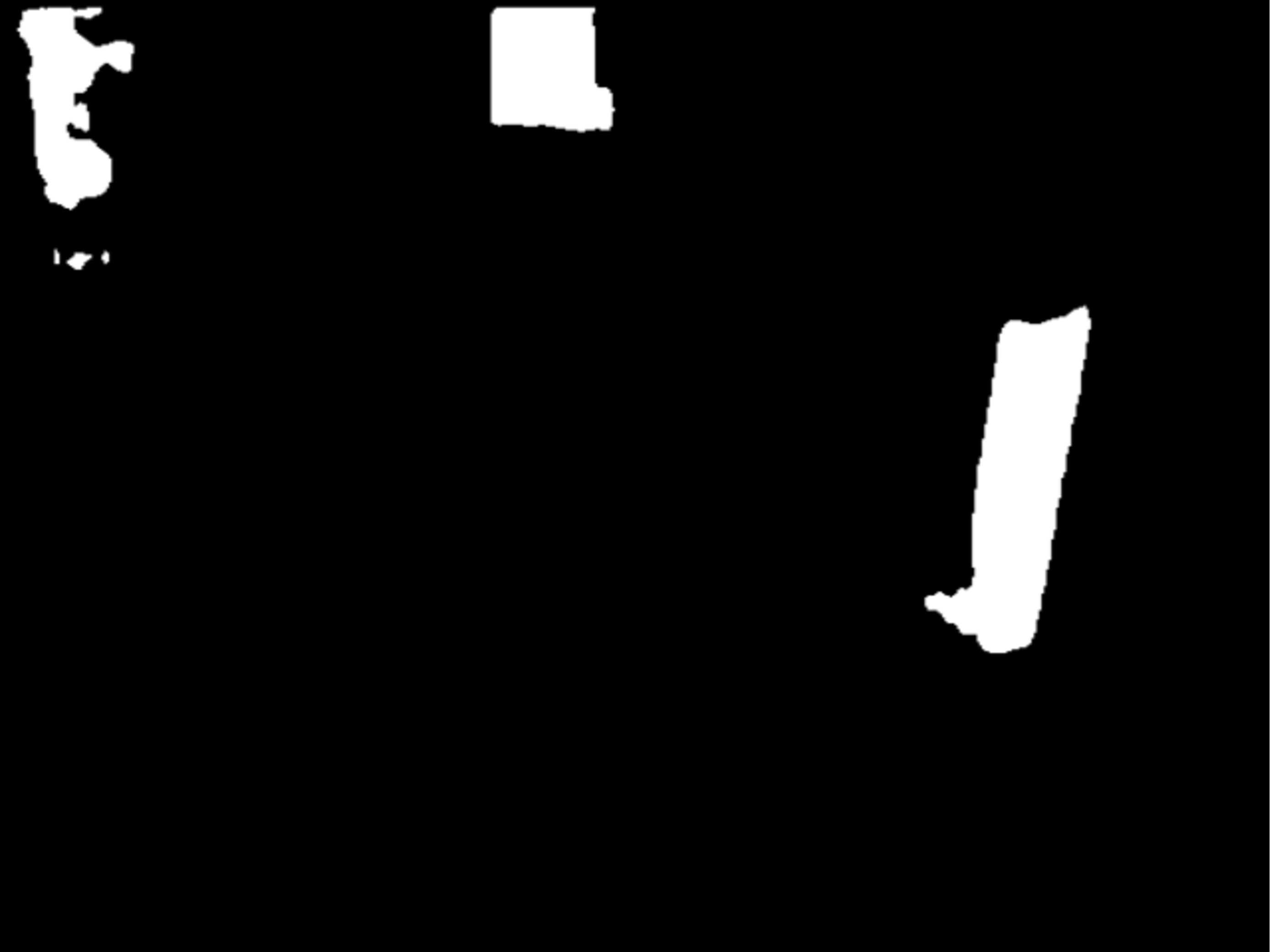}
        HetNet
    \end{minipage}%
    \begin{minipage}[b]{0.121\linewidth}
        \centering
        \includegraphics[width=0.83in,height=0.55in]{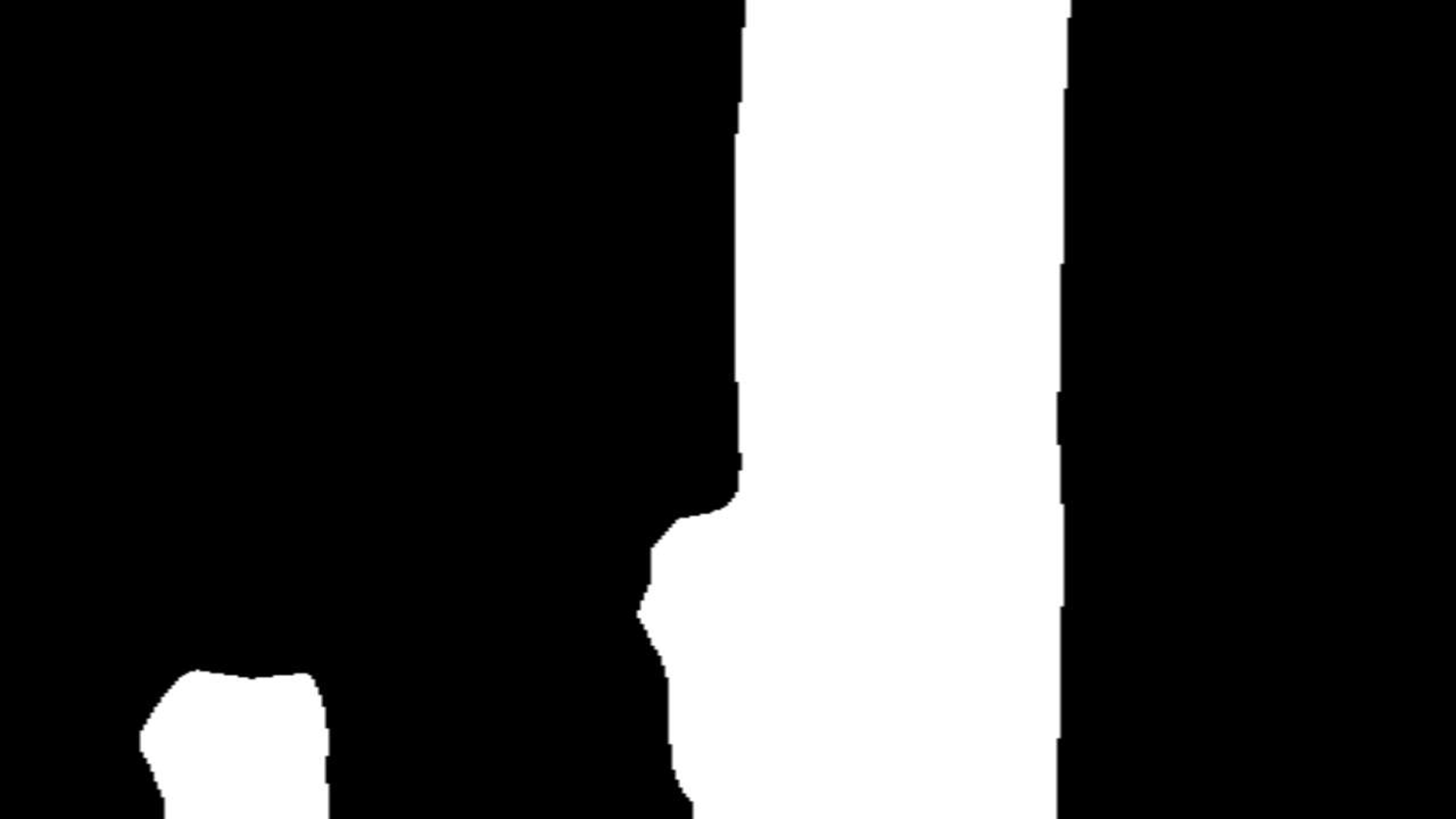}\vspace{2pt}
        \includegraphics[width=0.83in,height=0.55in]{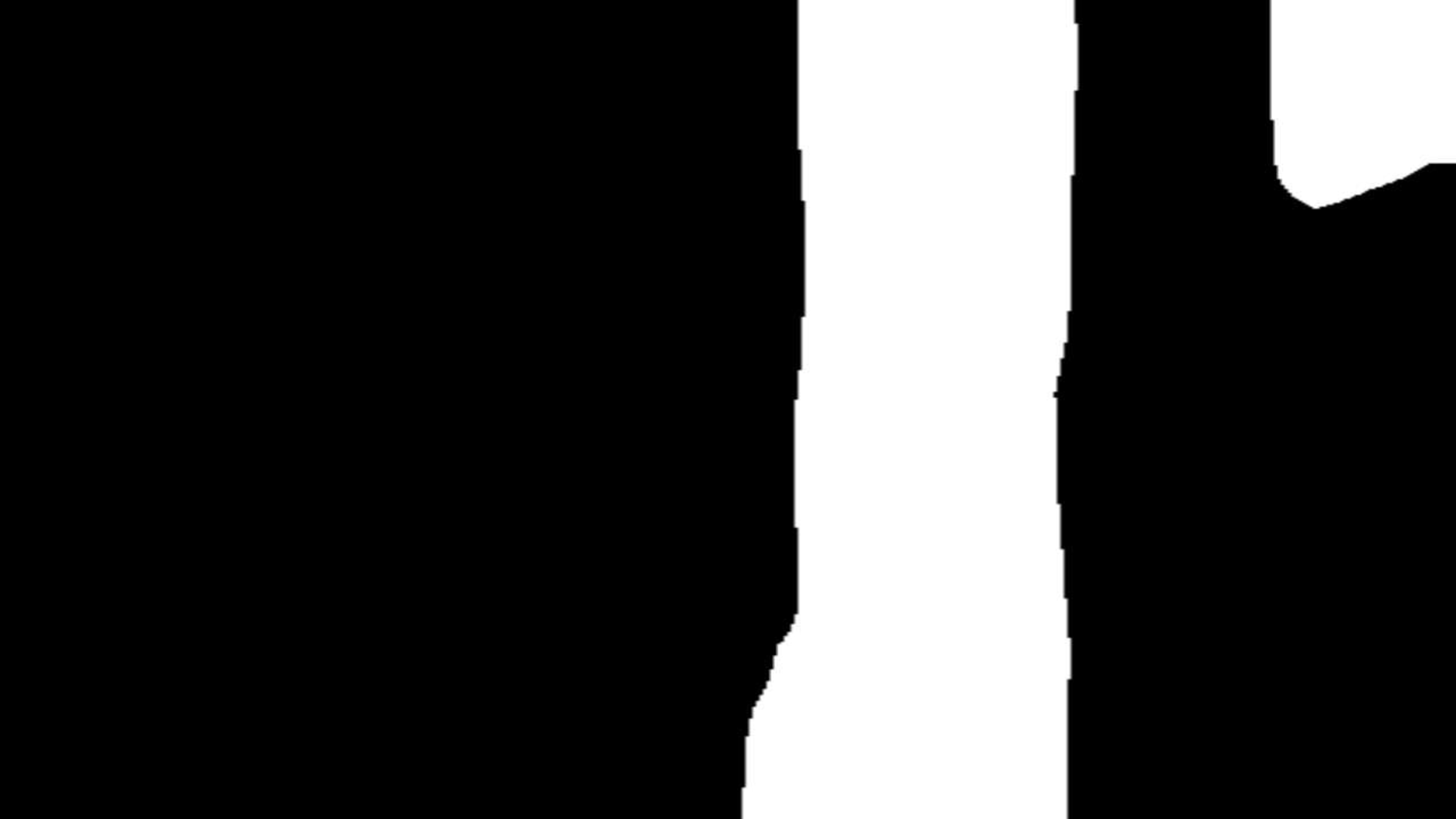}\vspace{2pt}
        \includegraphics[width=0.83in,height=0.55in]{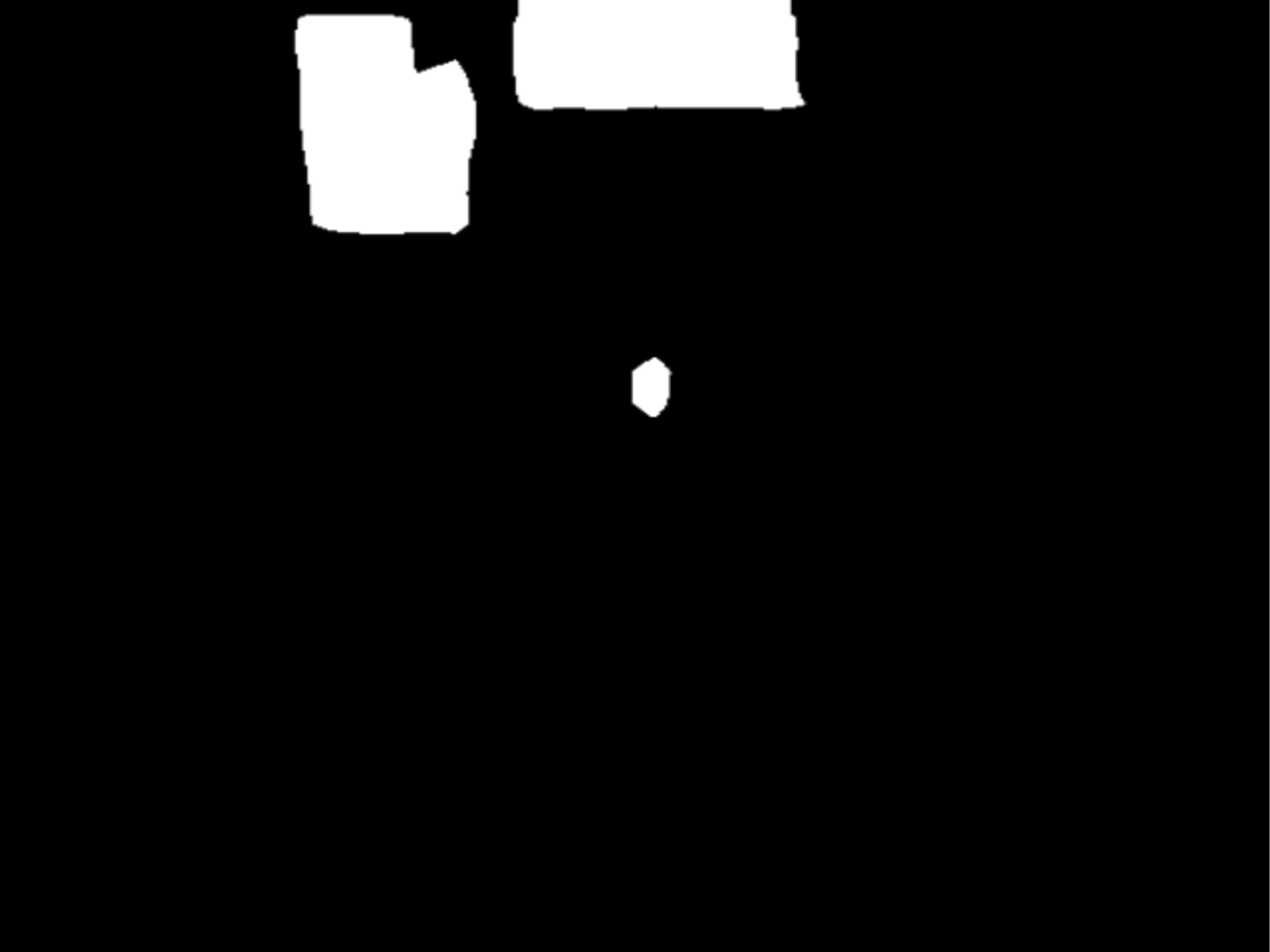}\vspace{2pt}
        \includegraphics[width=0.83in,height=0.55in]{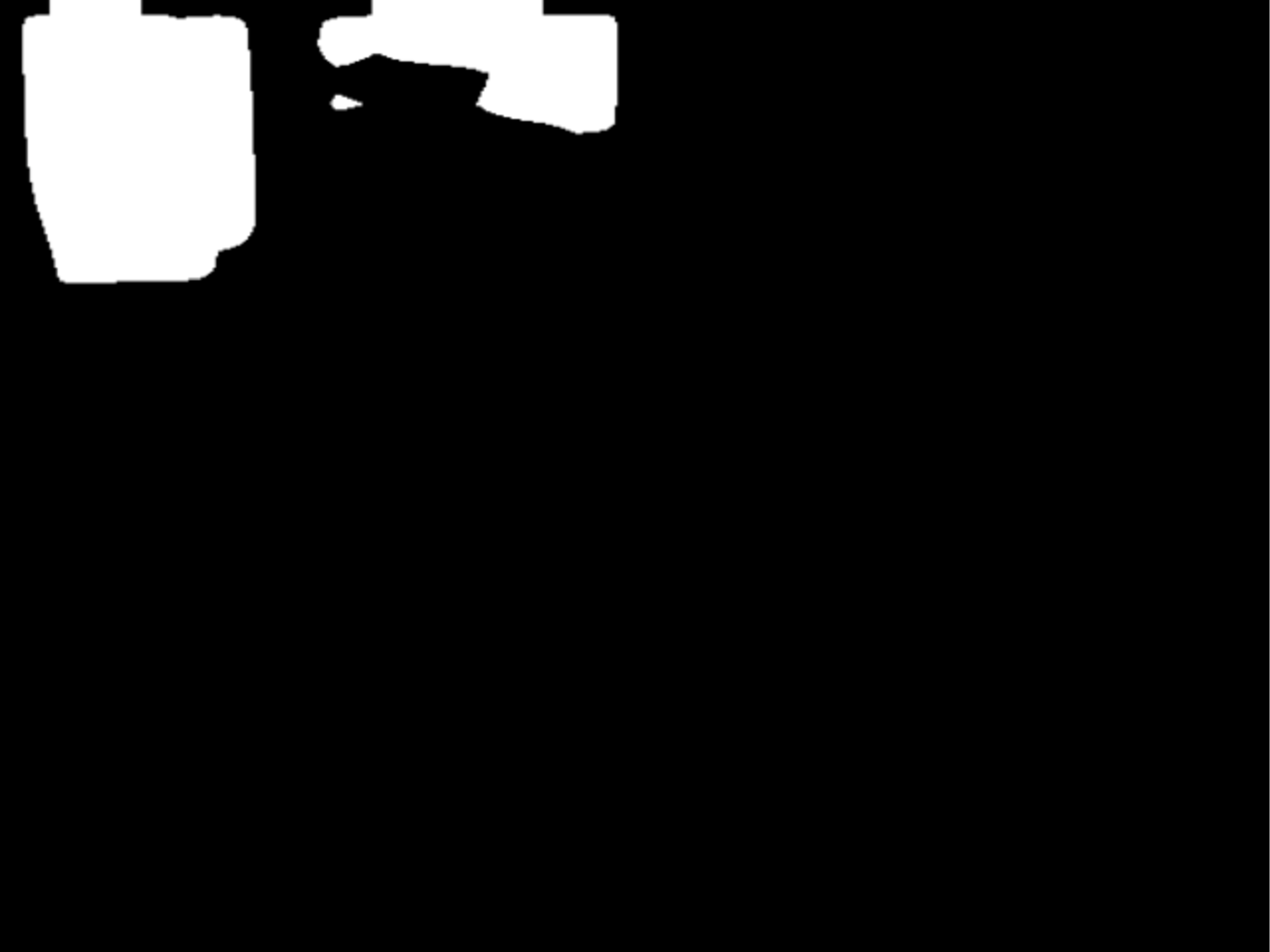}
        VMD
    \end{minipage}%
    \begin{minipage}[b]{0.121\linewidth}
        \centering
        \includegraphics[width=0.83in,height=0.55in]{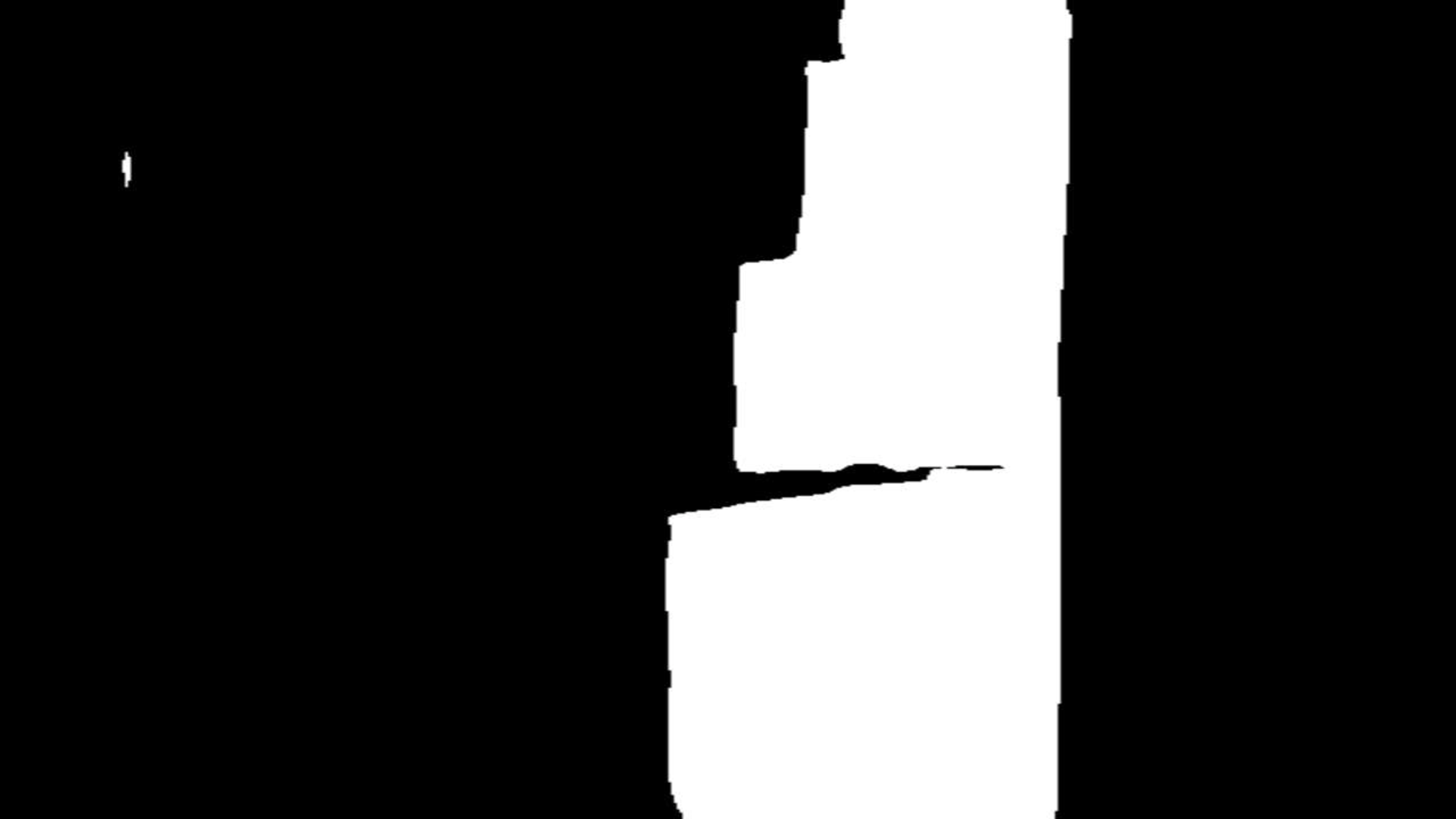}\vspace{2pt}
        \includegraphics[width=0.83in,height=0.55in]{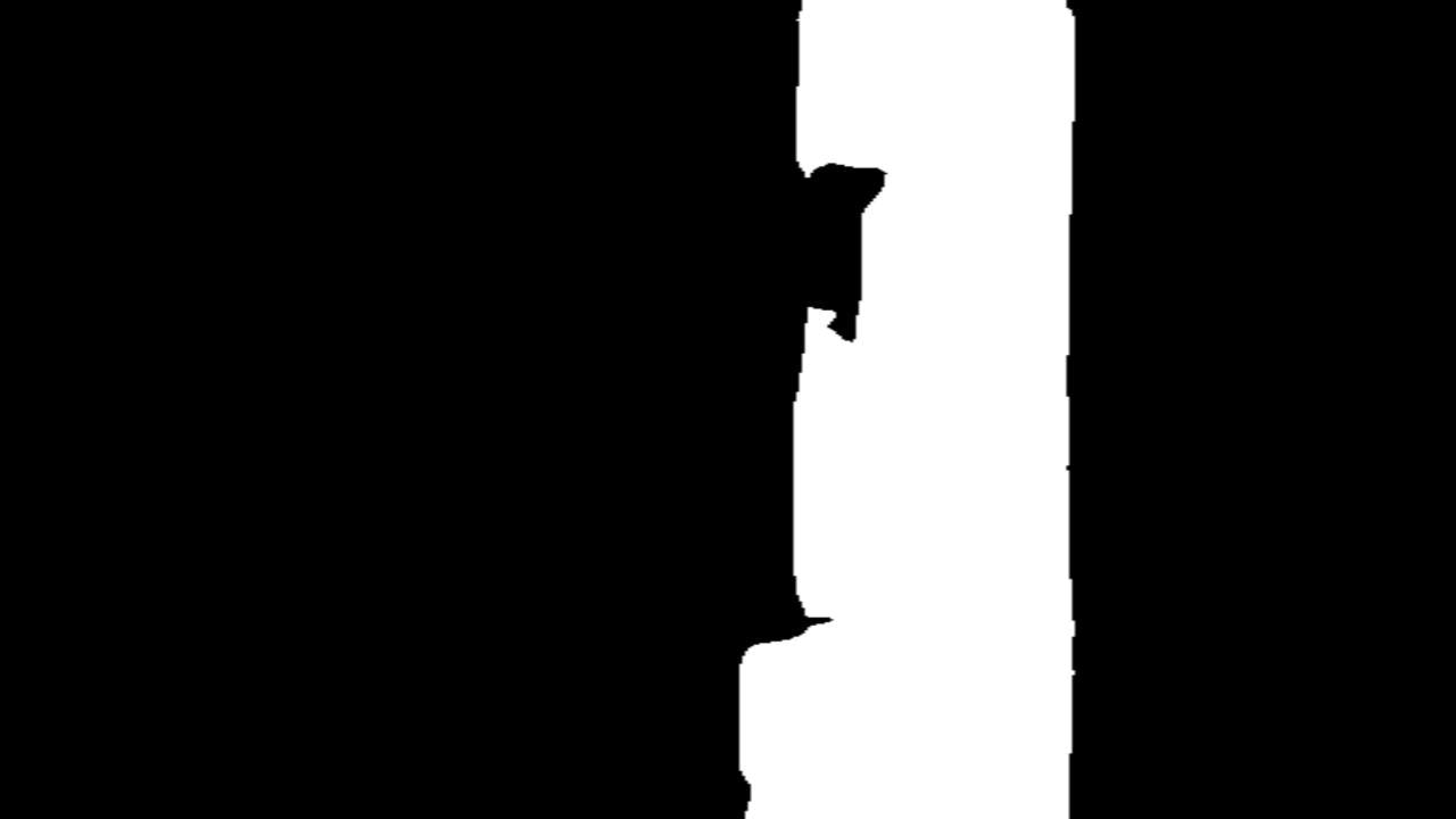}\vspace{2pt}
        \includegraphics[width=0.83in,height=0.55in]{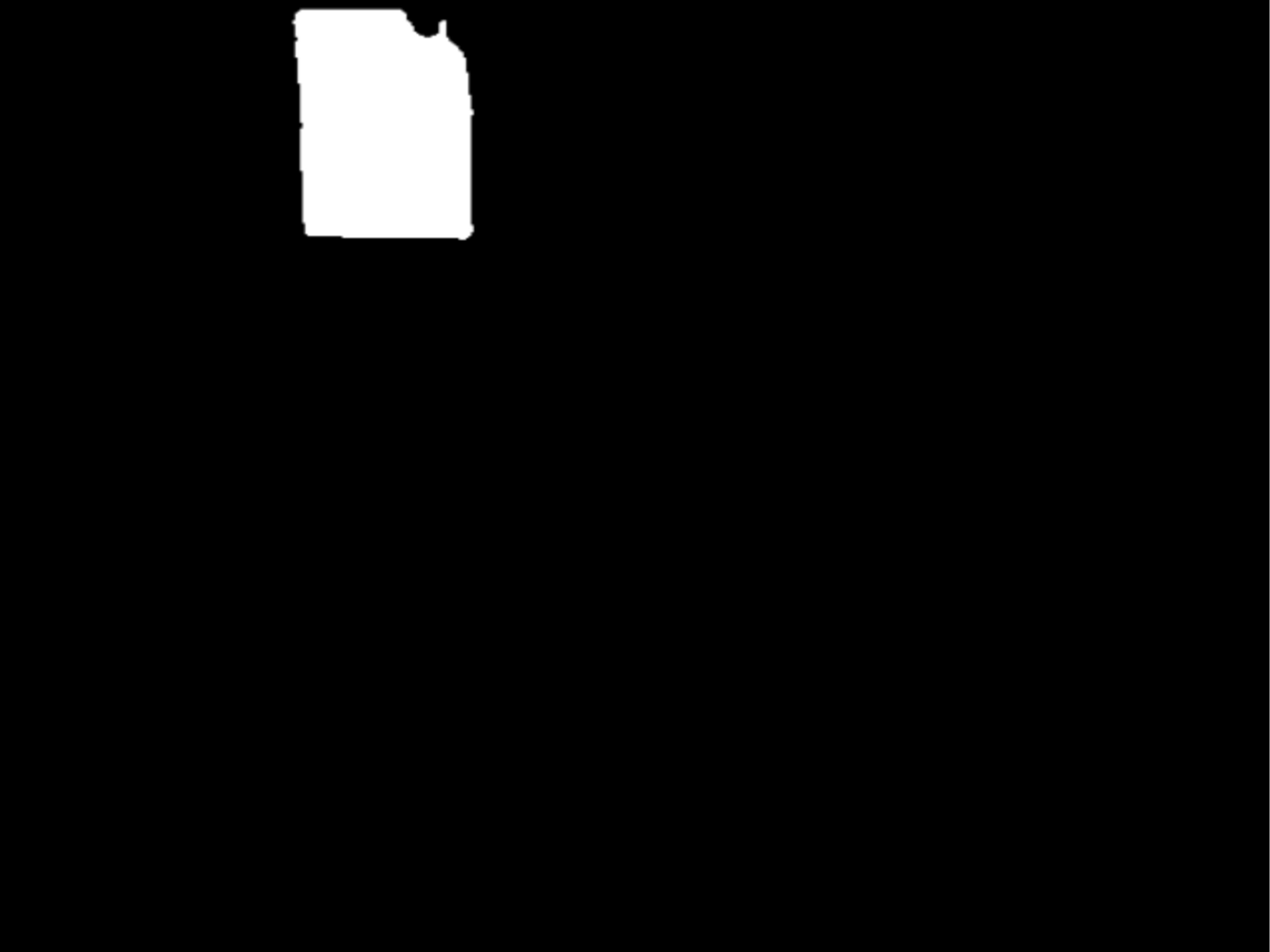}\vspace{2pt}
        \includegraphics[width=0.83in,height=0.55in]{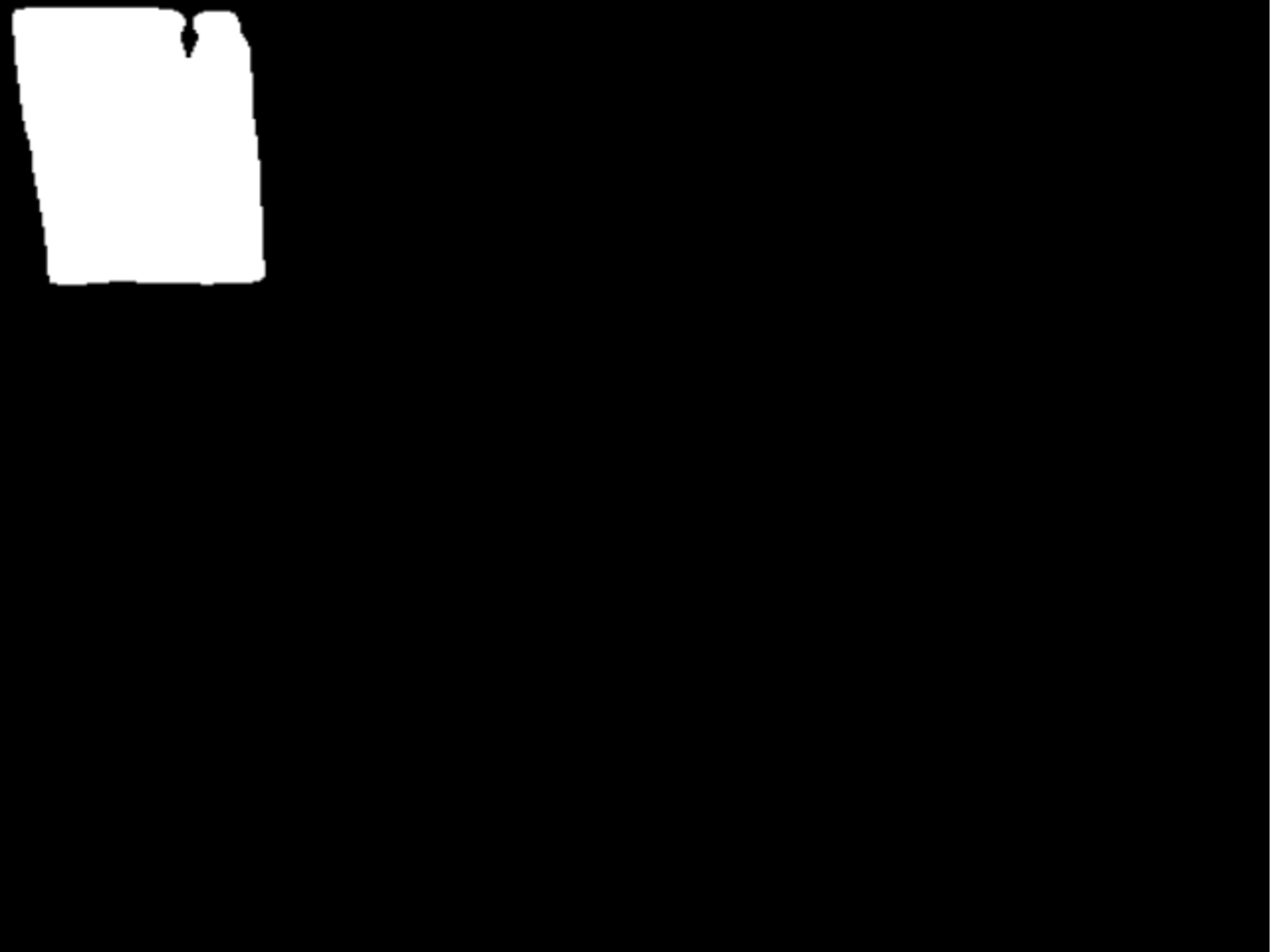}
        Ours
    \end{minipage}%
    \begin{minipage}[b]{0.121\linewidth}
        \centering
        \includegraphics[width=0.83in,height=0.55in]{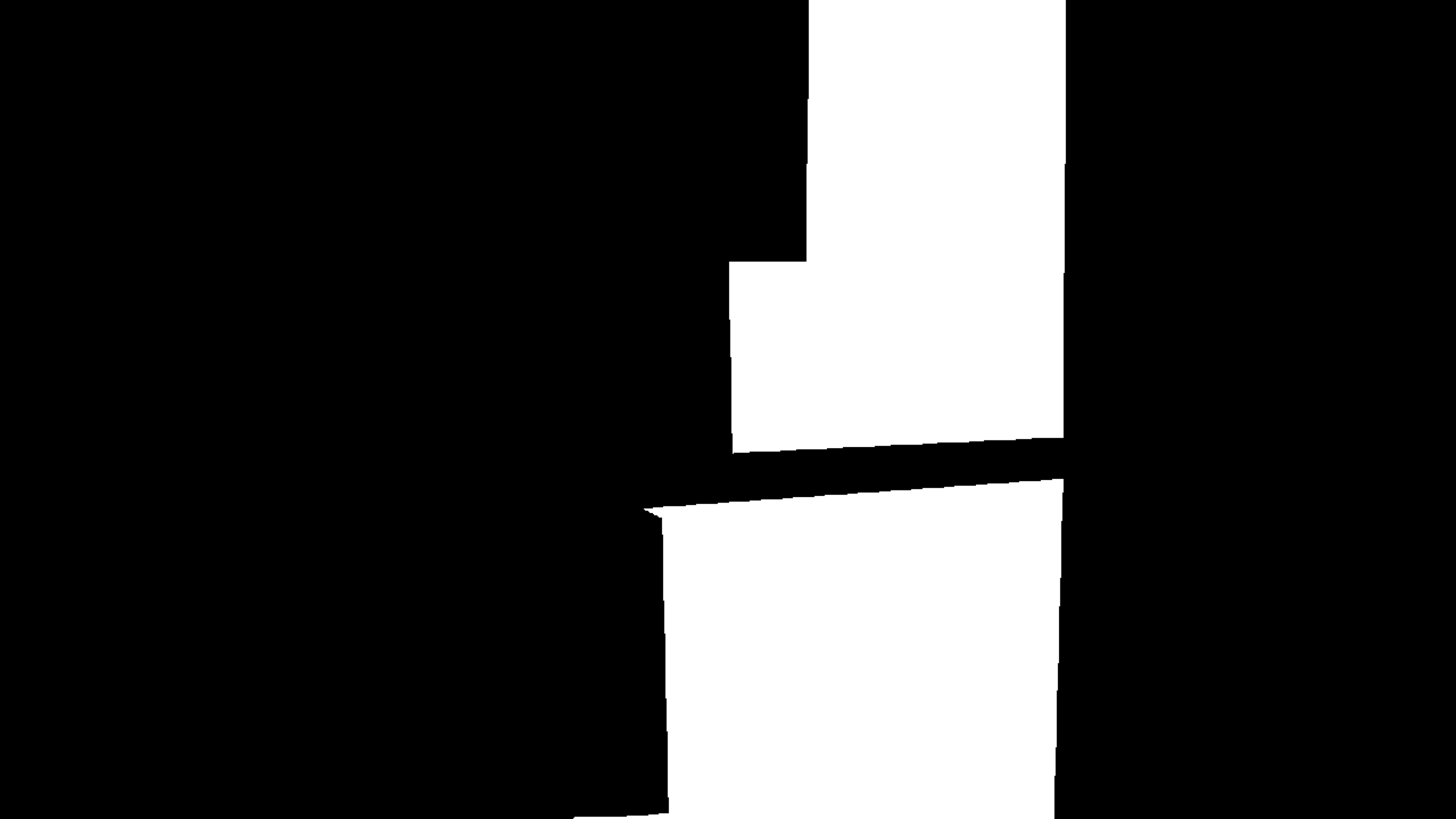}\vspace{2pt}
        \includegraphics[width=0.83in,height=0.55in]{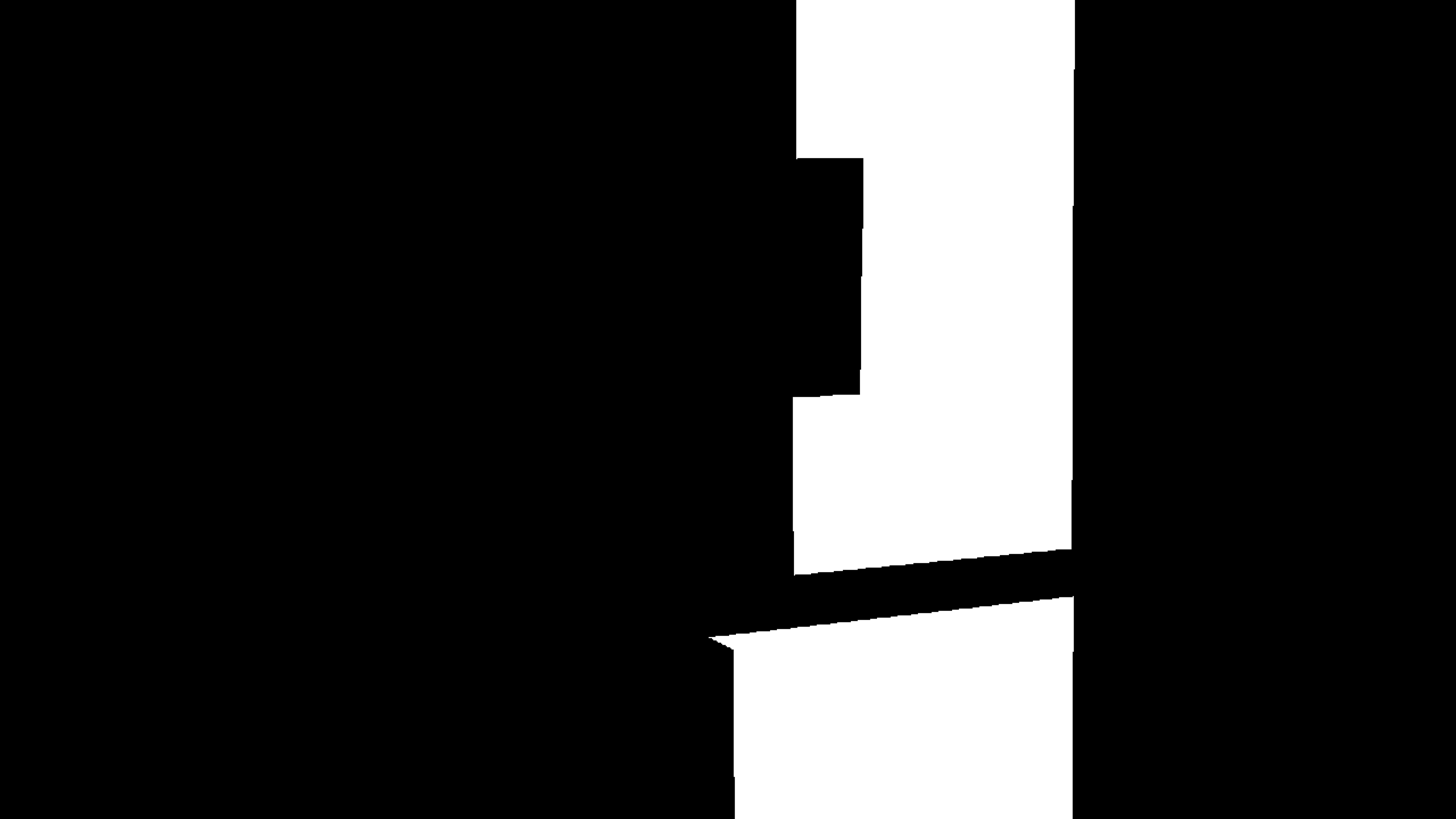}\vspace{2pt}
        \includegraphics[width=0.83in,height=0.55in]{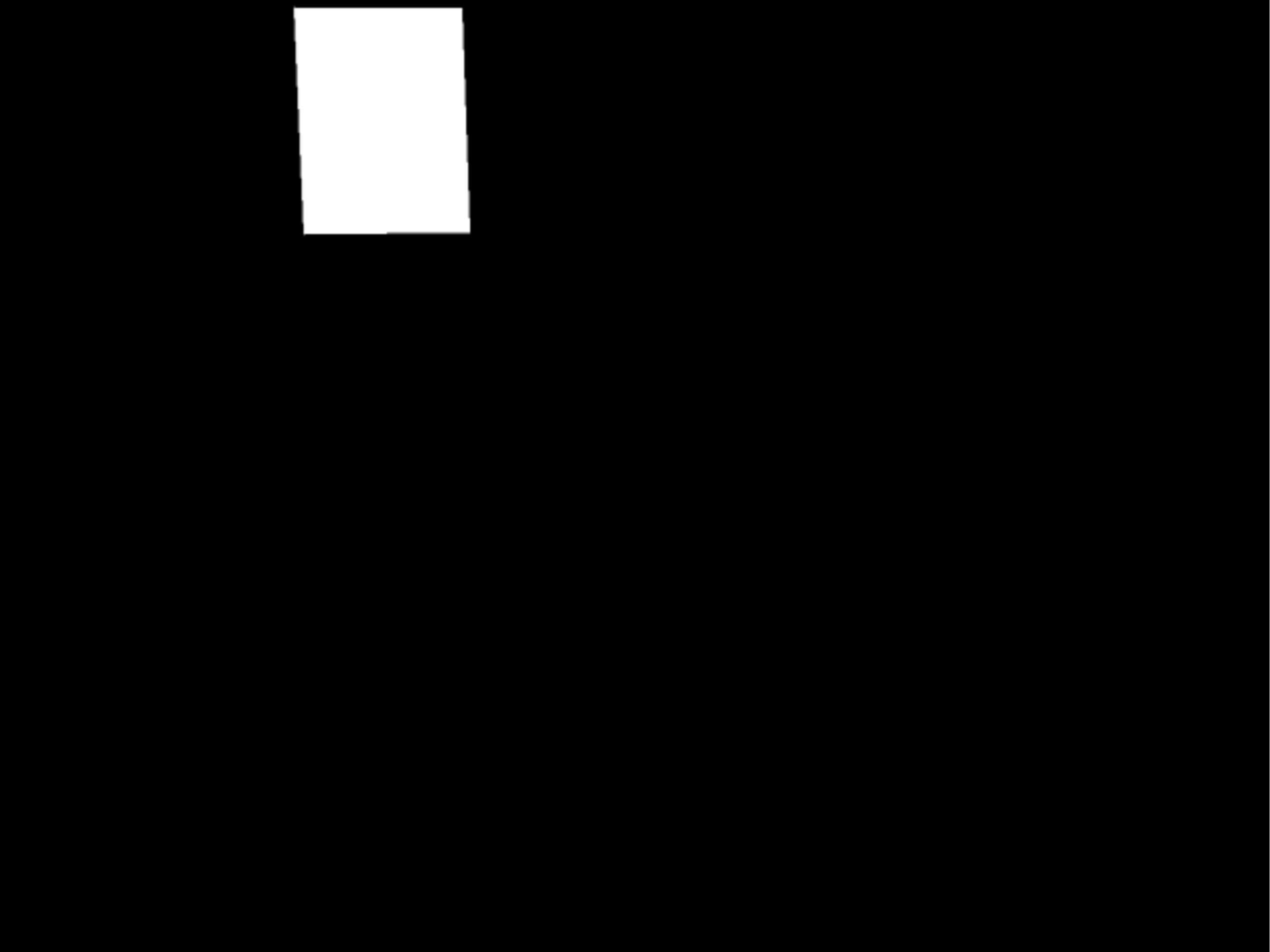}\vspace{2pt}
        \includegraphics[width=0.83in,height=0.55in]{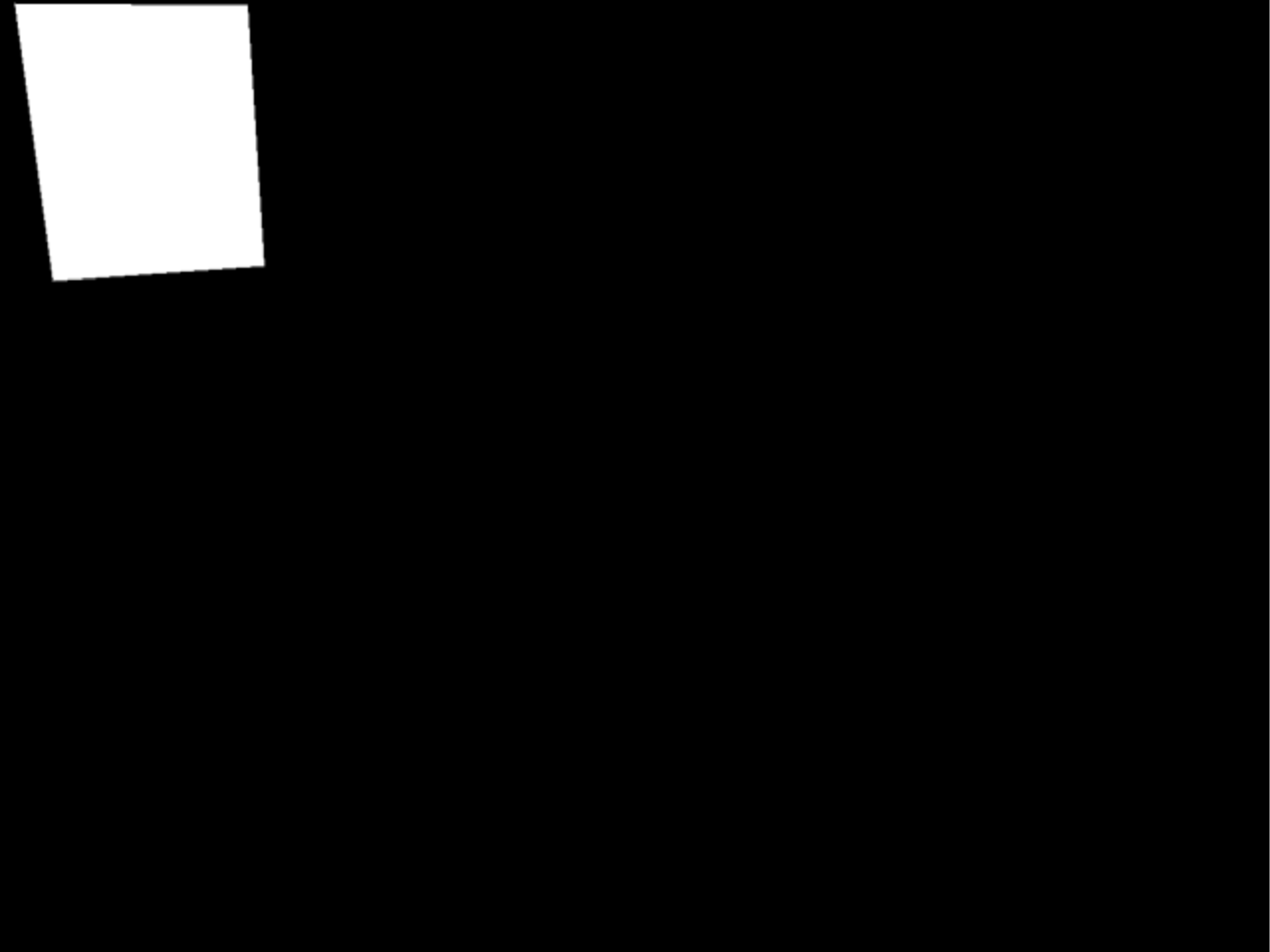}
        Ground Truth
    \end{minipage}%
  \caption{Visual results of our model, compared with relevant state-of-the-art methods.}
  \label{fig:qual_res}
\end{figure*}

\begin{table*}[htb]
  \centering
  \caption{Quantitative results on the VMD dataset (second column) and our benchmark (third column). The best results are shown in bold.}
  \label{tab:quan_res}
  \begin{tabular}{c|cccc|cccc}
    \hline
    \multirow{2}{*}{Method} & \multicolumn{4}{c|}{VMD} & \multicolumn{4}{c}{ViMirr} \\ 
                            \cline{2-9}
                            & IoU$\uparrow$ &$F_\beta\uparrow$ &Accuracy$\uparrow$  &MAE$\downarrow$    
                            & IoU$\uparrow$ &$F_\beta\uparrow$ &Accuracy$\uparrow$  &MAE$\downarrow$ \\
    \hline
    TVSD \cite{b18} &                   0.3060&0.5343&0.8160&0.1839&0.1413&0.3394&0.8605&0.1394\\
    SCOTCH \cite{b19} &                   0.5949&0.7281&0.8766&0.1233&0.6289&0.7596&0.9299&0.0702\\
    \hline
    GDNet \cite{b20} &                   
    0.5576&0.7335&0.8820&0.1179&0.5335&0.7118&0.9200&0.0799\\
    \hline
    MirrorNet \cite{b6} &                   0.5417&0.7506&0.8787&0.1211&0.4671&0.7015&0.9055&0.0944\\
    PMD \cite{b7} &                   0.5309&0.7823&0.8771&0.1229&0.5258&0.7417&0.9233&0.0766\\
    SANet \cite{b8} &                   0.4908&0.7202&0.8755&0.1245&0.4479&0.6286&0.9024&0.0975\\
    HetNet \cite{b4} &                   0.5145&0.7547&0.8726&0.1274&0.4544&0.6825&0.9063&0.0935\\
    \hline
    VMD \cite{b5} &                   0.5673&0.7873&0.8950&0.1052&0.4224&0.7001&0.9096&0.0903\\
    \hline
    Ours               &                   \textbf{0.6343}&\textbf{0.8104}&\textbf{0.9004}&\textbf{0.0995}&\textbf{0.6455}&\textbf{0.8261}&\textbf{0.9515}&\textbf{0.0484}\\
    \hline
  \end{tabular}
\end{table*}

\section{EXPERIMENTS}
\subsection{Datasets}
Recently, Lin \textit{et al.} proposed the first video mirror dataset (VMD) in \cite{b5}, which contains 14,987 frames from 269 videos with corresponding annotated masks. However, we notice that their data were mostly collected from similar scenes. 
In particular, more than 95\% of their data are collected from furniture stores (e.g. IKEA). 
This limits the diversity of the data, and will 
affect the generalization 
of the model to other scenes. 
Following \cite{b7}, we 
use SSIM \cite{b21} to study the similarity of videos in VMD. 
As the frames are similar in the same video, we use the first frame in each video 
to calculate the similarity 
score. Our ViMirr obtains 39.54\% similarity score, which is much lower than the 51.21\% of the VMD dataset. The details of the similarity score calculation are given in Section 1.2 of the supplemental material.



To address the limitations of VMD dataset, 
we construct the 
ViMirr dataset, 
which has 19,255 frames from 276 videos. Fig.~\ref{fig:examples_of_vimirr} shows some example video frames in 
ViMirr. 
To cover diverse realistic scenes, we studied five existing widely used datasets (\textit{i.e.} Matterport3D \cite{b22}, NYUv2 \cite{b14}, ScanNet \cite{b15}, DAVIS \cite{b23} and YouTube-VOS \cite{b24}), 
and manually selected 78 videos from NYUv2 and 126 videos from ScanNet, which contain mirrors in the videos. The indices of the videos selected are provided in Section 1.1 of the supplemental material. 
Moreover, we captured 13 videos to provide more popular scenarios (e.g., gym, lift) in daily life. Some examples we captured are given in Section 1.3 of the supplemental material. 
For both the collected and captured videos, we then manually annotated the mirrors in each frame. Example annotations can be seen in Fig.~\ref{fig:examples_of_vimirr}.

\subsection{Implementation Details}
The model was implemented in PyTorch \cite{b25} and trained 
on a PC with an NVIDIA RTX 4090 GPU card. During training, the images are resized 
to $512\times512$. We use Mix Transformer (MiT) \cite{b16} pre-trained on ADE20K \cite{b26,b27} dataset as 
the backbone network to extract image features. We adopt AdamW \cite{b28} with a weight decay of $5\times 10\textsuperscript{-4}$ as 
the optimizer. We have tried different values for learning rate, batch size, and the number of training epochs and empirically set the values for them to be 0.00001, 5, and 15, respectively.

\subsection{Comparison with the State-of-the-arts Methods}
We compare our method with state-of-the-art methods from four relevant fields: TVSD {\cite{b18} and SCOTCH \cite{b19} for video shadow detection, GDNet \cite{b20} for single-image glass detection,  
 MirrorNet \cite{b6}, PMD \cite{b7}, SANet \cite{b8} and HeNet \cite{b4} for single-image mirror detection and VMDNet \cite{b5} for video mirror detection, and the metrics we use are: intersection over union (IoU), F-measure ($F_\beta$), pixel-accuracy, and mean absolute error (MAE). These methods are chosen for comparison because their code is available. Although~\cite{b12} reported the state-of-the-art performance in image mirror detection, its code is not available and we cannot evaluate its performance on ViMirr or VMD data. 

Table \ref{tab:quan_res} shows the quantitive results on the VMD dataset and the 
proposed ViMirr dataset. 
Our method 
achieves the best performance on all four metrics. 
The different performances on the two datasets are mainly due to the different data distributions. VMD is a dataset built specifically for mirrors, while most of our data is selected from two public datasets. As a result, the mirror positions in VMD are toward the center, while the mirror positions in ViMirr are more scattered. 
Fig. \ref{fig:qual_res} shows the visual comparisons. We can see that the image sequences contain some regions (e.g. wood shelf or the door-like area of the first two rows and cabin in the third and fourth rows where red dotted lines circles) that look like 
mirrors. VMDNet tends to detect 
these regions as mirrors, while our method can differentiate them well. 

 \begin{table}[t]
 \begin{center}
 \caption{Ablation study results, trained and tested on the VMD dataset. "Baseline" denotes our network without all proposed modules. "CA" is the cross attention module proposed in \cite{b5}. "DGSA" is our dual-gated short-term attention module. "SLF" is our short-long fusion module.} \label{tab:ablation_study}
    \begin{tabular}{c|cccc}
        \hline
        Method & IoU$\uparrow$ & $F_\beta\uparrow$ & MAE$\downarrow$ & Accuracy$\uparrow$\\
        \hline
        Baseline & 0.6075 & 0.7676 & 0.1056 & 0.8943 \\
        \hline
        +CA & 0.6126 & 0.7919 & 0.1054 & 0.8946 \\
        +DGSA & 0.6147 & 0.8017 & 0.1045 & 0.8954 \\
        \hline
        +DGSA+SLF & 0.6343 & 0.8104 & 0.0995 & 0.9004\\
        \hline
\end{tabular}
\end{center}
\end{table}

\subsection{Ablation study}
We carried out ablation studies to demonstrate the effectiveness of our model. 
The last row in Table \ref{tab:ablation_study} shows that our final model with DGSA module and SLF module outperforms other baselines on all four metrics. The CA module brings improvements of baseline which shows 
the effectiveness of the spatial and temporal correspondence features. Compared with it, 
the DGSA module further improves the baseline, 
especially on $F_{\beta}$ by filtering the dual correspondence features. By fusing long-term correspondence features, the SLF module significantly benefits the mirror video detection tasks from a global view. A visual example of the ablation study is provided in Section 2 of the supplemental material.

\section{CONCLUSION}
In this paper, we have proposed a transformer network for Video Mirror Detection. It detects mirrors by fusing appearance features extracted from a short-term module and context information extracted from a long-term attention module. In addition, we construct a challenging benchmark that includes 19,255 frames from 281 videos covering a variety of daily scenes. Our experimental results demonstrate that the proposed model achieves state-of-the-art performance on both the VMD dataset and the benchmark. Our future work will consider improving the efficiency of our model to handle real-time video mirror detection and large-scale videos.

\end{document}